\documentclass[acmlarge]{acmart}
\pdfoutput=1
\usepackage{booktabs} 
\usepackage{enumerate}
\usepackage{booktabs}
\usepackage{makecell}
\usepackage{multirow}
\usepackage{flushend}
\usepackage{latexsym}
\usepackage{multirow}
\usepackage{amsmath}
\usepackage{amsfonts}
\usepackage{subfigure}
\usepackage[noend]{algpseudocode}
\usepackage{graphicx}
\usepackage{color}
\usepackage{algpseudocode}  
\usepackage{amsmath}
\usepackage{rotating} 
\usepackage{bm}
\usepackage{lipsum}
\usepackage{verbatim}
\usepackage{adjustbox}
\usepackage[graphicx]{realboxes}
\usepackage{rotating}
\usepackage{array}
\usepackage{threeparttable}
\usepackage{soul}
\usepackage{tabularx}
\usepackage{subcaption}
\usepackage{subfigure}
\usepackage{xcolor}
\usepackage[most]{tcolorbox}
\tcbuselibrary{breakable}
\usepackage{caption}
\usepackage{enumitem}
\definecolor{sgreen}{HTML}{75A99C}

\acmDOI{0000001.0000001}
    
\received[Received]{April 5, 2024}
\received[Revised]{June 29, 2024}
\received[Accepted]{September 7, 2024}

\newcommand{\revised}[1]{\textcolor{blue}{}\textcolor{black}{#1}}

\usepackage[linesnumbered,vlined,ruled,commentsnumbered]{algorithm2e}
\newlength\mylen

\makeatletter
\newcommand{\algorithmfootnote}[2][\footnotesize]{%
  \let\old@algocf@finish\@algocf@finish
  \def\@algocf@finish{\old@algocf@finish
    \leavevmode\rlap{\begin{minipage}{\linewidth}
    #1#2
    \end{minipage}}%
  }%
}

\theoremstyle{definition}
\newtheorem{definition}{Definition}[section]

\usepackage{xspace}

\usepackage{filecontents}

\begin{document}
\setlength{\intextsep}{5pt plus 1pt minus 0pt}
\setlength{\textfloatsep}{5pt plus 1pt minus 0pt}
\captionsetup[figure]{skip=1pt}
\captionsetup[table]{skip=3pt}

\title{Relational Prompt-based Pre-trained Language Models for Social Event Detection}

\author{Pu Li}
\authornote{Equal Contributions.}
\affiliation{%
  \institution{Kunming University of Science and Technology}
  \city{Kunming}
  \country{China}
  }
\email{lip@stu.kust.edu.cn}

\author{Xiaoyan Yu}
\authornotemark[1]
\affiliation{%
  \institution{Beijing Institute of Technology}
  \city{Beijing}
  \country{China}
  }
\email{xiaoyan.yu@bit.edu.cn}

\author{Hao Peng}
\affiliation{%
  \institution{Beihang University}
  \city{Beijing}
  \country{China}
  }
\email{penghao@buaa.edu.cn}

\author{Yantuan Xian}
\authornote{Corresponding author.}
\affiliation{%
  \institution{Kunming University of Science and Technology}
  \city{Kunming}
  \country{China}
  }
\email{xianyt@kust.edu.cn}

\author{Linqin Wang}
\affiliation{%
  \institution{Kunming University of Science and Technology}
  \city{Kunming}
  \country{China}
  }
\email{linqinwang7767@163.com}

\author{Li Sun}
\affiliation{%
  \institution{North China Electric Power University}
  \city{Beijing}
  \country{China}
  }
\email{ccesunli@ncepu.edu.cn}

\author{Jingyun Zhang}
\affiliation{%
  \institution{Beihang University}
  \city{Beijing}
  \country{China}
  }
\email{zhangjingyun@buaa.edu.cn}

\author{Philip S. Yu}
\affiliation{%
  \institution{University of Illinois at Chicago}
  \city{Chicago}
  \country{USA}
 }
\email{psyu@uic.edu}

\begin{abstract}
Social Event Detection (SED) aims to identify significant events from social streams, and has a wide application ranging from public opinion analysis to risk management. 
In recent years, Graph Neural Network (GNN) based solutions have achieved state-of-the-art performance. 
However, GNN-based methods often struggle with \revised{missing and noisy edges} between messages, affecting the quality of learned message embedding. 
Moreover, these methods statically initialize node embedding before training, which, in turn, limits the ability to learn from message texts and relations simultaneously.
In this paper, we approach social event detection from a new perspective based on Pre-trained Language Models (PLMs), and present \textit{$\mathrm{RPLM}_{SED}$} (\textbf{R}elational prompt-based \textbf{P}re-trained \textbf{L}anguage \textbf{M}odels for \textit{S}ocial \textit{E}vent \textit{D}etection).
We first propose a new pairwise message modeling strategy to construct social messages into message pairs with multi-relational sequences.
Secondly, a new multi-relational prompt-based pairwise message learning mechanism is proposed to learn more comprehensive message representation from message pairs with multi-relational prompts using PLMs.
Thirdly, we design a new clustering constraint to optimize the encoding process by enhancing intra-cluster compactness and inter-cluster dispersion, making the message representation more distinguishable. 
We evaluate the \textit{$\mathrm{RPLM}_{SED}$} on three real-world datasets, demonstrating that the \textit{$\mathrm{RPLM}_{SED}$} model achieves state-of-the-art performance in offline, online, low-resource, and long-tail distribution scenarios for social event detection tasks.
\end{abstract}


\authorsaddresses{
Authors' addresses: 
P. Li, Faculty of Information Engineering and Automation, and Yunnan Key Laboratory of Artificial Intelligence, Kunming University of Science and Technology, Kunming 650500, China; email: \path{lip@stu.kust.edu.cn};
X. Yu, School of Computer Science and Technology, Beijing Institute of Technology, Beijing 100081, China; email: \path{xiaoyan.yu@bit.edu.cn};
H. Peng, School of Cyber Science and Technology, Beihang University, No. 37 Xue Yuan Road, Haidian District, Beijing, 100191, China; email: \path{penghao@buaa.edu.cn}; 
Y. Xian, Faculty of Information Engineering and Automation, and Yunnan Key Laboratory of Artificial Intelligence, Kunming University of Science and Technology, Kunming 650500, China; email: \path{xianyt@kust.edu.cn};
L. Wang, Faculty of Information Engineering and Automation, and Yunnan Key Laboratory of Artificial Intelligence, Kunming University of Science and Technology, Kunming 650500, China; email: \path{linqinwang7767@163.com};
L. Sun, School of Control and Computer Engineering, North China Electric Power University, Beijing 102206, China; email: \path{ccesunli@ncepu.edu.cn};
J. Zhang, School of Cyber Science and Technology, Beihang University, No. 37 Xue Yuan Road, Haidian District, Beijing, 100191, China; email: \path{zhangjingyun@buaa.edu.cn};
P. S. Yu, Department of Computer Science, University of Illinois at Chicago, Chicago 60607, IL; email: \path{psyu@uic.edu}.
}


\maketitle

\renewcommand{\shortauthors}{P. Li et al.}
\section{Introduction}
\label{sec: introduction} 
Social events refer to the various incidents or activities occurring in reality that elicit widespread public interest and discourse across platforms such as Twitter, Weibo, Facebook, Tumblr, Telegram, etc.
These events encompass a broad spectrum of topics, including news reports, sudden public incidents, social trends, market dynamics, and more~\cite{cordeiro2016online}.
Social Event Detection (SED) aims to group messages reporting on the same incident to identify significant events within them from social streams~\cite{atefeh2015survey,li2021deep}. 
Owing to its capability to provide deep insights into public issues and trends, it is extensively applied in fields such as public opinion analysis~\cite{peng2021streaming}, financial market analysis~\cite{nisar2018twitter}, risk management~\cite{liu2017Event}, and political analysis~\cite{marozzo2018analyzing}. 

Social data's incompleteness, ambiguity, and streaming nature present more significant challenges for social event detection than traditional text mining tasks. 
Early social event detection methods typically focused on utilizing text contents~\cite{amiri2016short, morabia2019sedtwik, sahnoun2020event, wang2017neural, yan2015probabilistic, zhao2011comparing} or attributes extracted from texts~\cite{ feng2015streamcube, xie2016topicsketch, xing2016hashtag} to identify and categorize events. 
These methods focus on categorizing messages into different events but fail to detect events from social streams.
In recent years, Graph Neural Network (GNN) based social event detection methods have demonstrated tremendous potential ~\cite{peng2019fine, peng2021streaming, cao2021knowledge, cui2021mvgan, ren2022known, peng2022reinforced, ren2023uncertainty}. 
These methods typically model social messages into graphs to integrate semantic and structural information.
By leveraging GNNs to learn message representations from graphs and employing clustering techniques for event detection, these approaches significantly enhance the performance of social event detection.
For instance, \citet{peng2019fine} construct social messages as a Heterogeneous Information Network (HIN) and, upon obtaining representations through Graph Convolutional Networks (GCN), employ a pairwise sampling method to address the challenges presented by the numerous event categories and the sparse number of instances per category.
\citet{cui2021mvgan} respectively construct semantic and temporal views by leveraging hashtags and time distribution of messages, then integrate representations from semantic and temporal views using a hashtag-based multi-view graph attention mechanism.
\citet{cao2021knowledge} construct messages as a HIN graph and then transform it into a homogeneous graph, utilizing the learning and inductive capabilities of GNNs to achieve incremental event detection.
Moreover,~\citet{peng2022reinforced}, recognizing that different relations between messages may have varying degrees of value, \revised{models} social messages into a weighted multi-relational graph, employing reinforcement learning to determine the optimal thresholds for various relations.

\begin{figure}[h]
    \centering
    \includegraphics[width=1\linewidth]{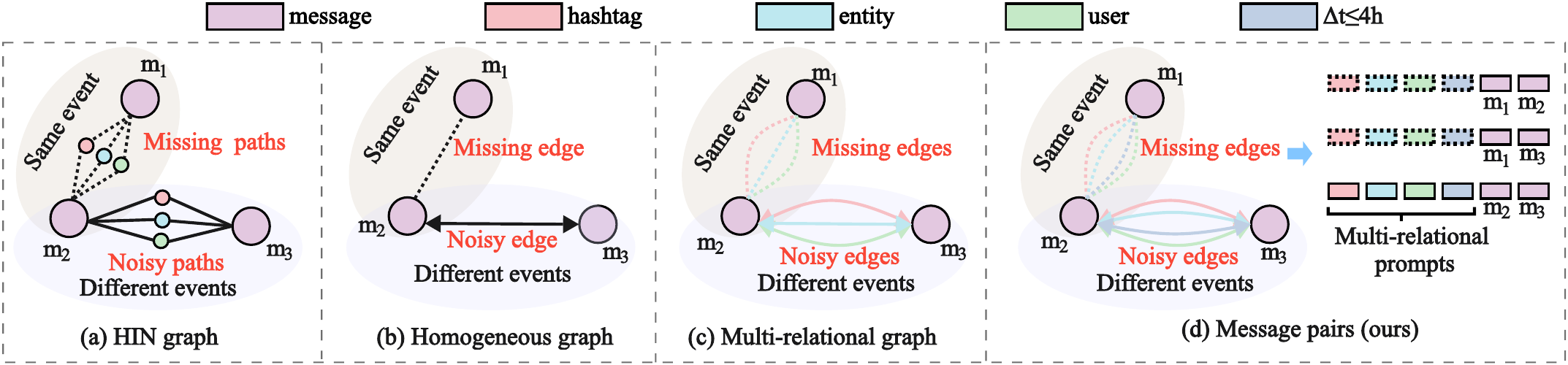}
    \caption{Social message modeling methods. 
    \textbf{(a) HIN graph} models social messages into the HIN graph by extracting attributes from the messages (e.g., entities, hashtags, users, etc.) as relational nodes connecting to the related message nodes.
     \textbf{(b) Homogeneous graph} is obtained by mapping a HIN with message nodes retained and adding edges between message nodes that share common adjacent relation nodes.
    \textbf{(c) Multi-relational graph}: Similar to (b), but transforms relation nodes into different types of edges between message nodes.
    \textbf{(d) Message pairs}: Converting multi-relational edges in the multi-relational graph into multi-relational prompts. 
    Regardless of the actual co-occurrence information between messages, corresponding multi-relational prompts will exist.}
    \label{fig:message modeling}
\end{figure}

\revised{However, GNN-based social event detection methods still exhibit the following limitations.}
\begin{enumerate}
    \item 
    \revised{Graphs constructed from common attributes (e.g., entities, hashtags, users, etc.) between messages face issues of missing and noisy edges~\cite{liu2022towards}.
    These issues are primarily manifested in the lack of connections between message nodes within the same event and the existence of connections between nodes from different events. 
    Whether modeling social messages as HIN graph (Figure~\ref{fig:message modeling}(a)), homogeneous graph (Figure~\ref{fig:message modeling}(b)), or multi-relational graph (Figure~\ref{fig:message modeling}(c)), none of these approaches can handle well above issues.
    This challenge is multifaceted. 
        On the one hand, intra-event messages may have no connections. 
        This impedes GNNs from capturing associations between these nodes, resulting in an incomplete representation of related information. 
        On the other hand, attribute co-occurrence is probable for inter-event messages, potentially introducing noise and impacting the quality of node representation.
        Besides, GNN-based methods fail to transmit information adequately in sparse graph scenarios, leading to suboptimal message representations.}
    \item 
     \revised{GNN-based methods initialize node embeddings statically via pre-trained word embedding models~\cite{mikolov2013efficient,pennington2014glove,bojanowski2017enriching} before training. 
    This practice prevents the acquisition of better semantic embeddings from the message content in subsequent learning processes~\cite{NEURIPS2021_f18a6d1c}.
    Meanwhile, GNNs inherently separate the utilization of relations/edges and message content when learning message embeddings from explicit message graphs.
    The role of edges primarily select nodes for aggregation rather than directly participating in the processing of message representations.
    This approach may not fully consider the complexity of relations between messages and the potential interactions between these relations and message content~\cite{ding2023eliciting}.}
    \item 
    \revised{GNN-based methods exhibit insufficient constraints on the training process, treating messages merely as targets to be pulled closer or pushed apart~\cite{hermans2017defense}.}
\end{enumerate}

Herein, to circumvent the aforementioned challenges of GNN-based methods, we put forward a perspective of learning expressive message representations with Pre-trained Language Models (PLMs), instead of GNNs.
Thus, we present a novel \textbf{R}elational prompt-based \textbf{P}re-trained \textbf{L}anguage \textbf{M}odel for \textit{S}ocial \textit{E}vent \textit{D}etection, namely \textit{$\mathrm{RPLM}_{SED}$}.
Concretely, we \textbf{first} propose a new pairwise message modeling strategy to model social messages.
As illustrated in Figure~\ref{fig:message modeling}(d), we sample other messages for each message to form pairs and convert the multiple relations between the two into a multi-relational sequence to preserve structural information.
\revised{For any two messages, each type of relation between them and the existence or non-existence of these relations can be comprehensively represented as a multi-relational sequence.
This strategy addresses the issues of missing and noisy edges in the message graph and achieves the joint modeling of the messages' structure and semantics.}
\textbf{Secondly}, a new multi-relational prompt-based pairwise message learning mechanism is proposed to learn more detailed and comprehensive message representations from semantic and structural information between messages.
\revised{It breaks away from traditional methods that rely on explicit relational graph learning and aggregation of message representations, and dividing message representation learning and aggregation into two independent parts.}
We transform the multi-relational sequences into multi-relational prompt embeddings and integrate them with the content embeddings of messages.
By consistently encoding the embeddings of message pairs along with their corresponding multi-relational prompt embeddings through PLMs~\cite{devlin-etal-2019-bert,liu2019roberta,zhang2022twhin,loureiro2023tweet}, we achieve the simultaneous utilization of message content and structural information, as well as the dynamic embedding and encoding of message content.
To obtain more robust and stable message representations from their corresponding message pairs, a similarity-based message representation aggregation approach is introduced to aggregate the representations extracted from message pairs with high confidence levels, serving as the final representation of the message. 
This approach also avoids the over-smoothing of representations and enhances the similarity between representations within the same event. 
\textbf{Finally}, we design a new clustering constraint encompassing inter-cluster and intra-cluster losses. 
\revised{Unlike previous optimization methods that solely consider messages as targets, the objective of the clustering constraint optimization is focused on event clusters.
The inter-cluster loss aims to distance the cluster centers of different events, whereas the intra-cluster loss ensures that messages of the same event are gathered close to their respective cluster center.
This clustering constraint strengthens the cohesiveness within clusters and the separation between them, thereby effectively improving the distinguishability of the message representations.}
By leveraging the inherent abilities of PLMs to learn, retain, and expand knowledge, we achieve incremental social event detection.

\revised{We conduct comprehensive experiments on three datasets, Events2012~\cite{mcminn2013building}, Events2018~\cite{mazoyer2020french}, and Arabic-Twitter~\cite{alharbi2021kawarith}.
Analyses and comparisons are conducted with one community detection method~\cite{liu2020story}, eight GNN-based methods~\cite{peng2019fine,cao2021knowledge,ren2022known,peng2022reinforced,ren2021transferring,cui2021mvgan,ren2022evidential,ren2023uncertainty}, and seven pre-training-based methods~\cite{mikolov2013efficient,devlin-etal-2019-bert,liu2019roberta} across online, offline, low-resource, and long-tail distribution scenarios.
In all scenarios, $\mathrm{RPLM}_{SED}$ demonstrates SOTA performance compared to all benchmark methods.
Additionally, we compare and analyze the performance of $\mathrm{RPLM}_{SED}$ based on different PLMs and report the time consumption for each part of $\mathrm{RPLM}_{SED}$.
Finally, we analyze the performance of three density-based clustering algorithms~\cite{campello2013density,ester1996density,ankerst1999optics} on the representations obtained by  $\mathrm{RPLM}_{SED}$.
The codes for all baseline models and $\mathrm{RPLM}_{SED}$, along with all datasets, are publicly accessible on GitHub\footnote{\url{https://github.com/RingBDStack/RPLM_SED}}.}

In summary, the contributions of this paper are as follows:
\begin{itemize}
\item We present a novel relational prompt-based pre-trained language model for social event detection ($\mathrm{RPLM}_{SED}$). 
To the best of our knowledge, this is the first SED method based on PLM that integrates both semantic information from social messages and structural relations between them.
It breaks away from traditional GNN-based methods, no longer relying on explicit structural relations between messages to learn representations.
\item  A new pairwise message modeling strategy is proposed to solve the issues of missing and noisy edges in the message graph, while simultaneously achieving joint modeling of message content and structural information.
\item  A new multi-relational prompt-based pairwise message learning mechanism is proposed to learn more detailed and comprehensive message representations from message pairs with multi-relational prompts.
It leverages the powerful semantic understanding and context encoding capabilities of PLMs, consistently encoding both message text pairs and their corresponding multiple relations, thus enabling the simultaneous utilization of textual content and structural information.
\item A new clustering constraint is designed to enhance the discriminability of message representations. 
This clustering constraint loss pushes the cluster centers of different events apart and constrains messages from the same event to be closer to their corresponding cluster center.
\item Comprehensive experiments and analyses have been conducted on three datasets, illustrating that the proposed approach surpasses the current SOTA methods for social event detection regarding effectiveness. 
\end{itemize}

The structure of this paper is as follows: 
Section~\ref{sec: background} outlines the background and preliminaries of our work.
In Section~\ref{sec: methodology}, we describe the technical details of the proposed model, named $\mathrm{RPLM}_{SED}$. 
Section~\ref{sec: Experimental Setup} presents the experimental setup, and Section~\ref{sec: Results And Discussion} discusses the experiment's results. 
Section~\ref{sec: Related Work} provides an overview of related works. 
Finally, we conclude the paper in Section~\ref{sec: conclusion}.

\section{Background and Overview}
\label{sec: background}
This section first provides an overview of the problems and challenges encountered in social event detection. 
Subsequently, we detail the core definitions related to social event detection. 
Additionally, Table~\ref{tab:notations} presents a comprehensive list of the primary symbols used throughout this paper.

\begin{table}[htp]
    \centering
    \caption{Glossary of notations.}
    \label{tab:notations}
    \begin{tabularx}{\linewidth}{l|X}
	\hline
	\textbf{Notation} & \textbf{Description}  \\ 
	\hline
	  $S$       &A social stream \\
	  $M$; $m$   & A message block; A message in a message block or event\\
	  $E_i$     & Set of events from message block $M_i$ \\
  	$e_k$     & $k\text{-th}$ event in \(E_i\) \\
        $\mathcal{G}$   & A multi-relational message graph \\
        $\mathcal{M}$   & The set of all nodes in $\mathcal{G}$ \\
        $w$         & The window size for maintaining the model \\
        $\theta$; $\theta^{plm}$ & The parameters of model; The parameters with the PLM section of model\\
        $\zeta$ & The parameter update weight during model maintenance\\
        $\mathcal{R}_t$ & Edge set with relational type $t$ in $\mathcal{G}$ \\
        $r_{i,s}^t$   &  An edge between \( m_i\) and sampled node \(m_s \) with relational type \(t\) \\
        $r_{i,s}$   & The multi-relational sequence between \( m_i\) and sampled node \(m_s \)  \\
	  $D$; $d$ & Preprocessed message blocks; A message pair\\
        $o$ & The number of heads in structured attention mechanism\\
        $\hat{x}_i$; $\hat{x}_s$ &  The representations of \(m_i\) and sampled node \(m_s\) obtained through structured attention mechanism and averaging\\
        $q_i$ & The similarity between \(m_i\) and sampled node \(m_s\)\\
        $X_i$; $Q_i$ & A set of candidate representations of \(m_i\); A set of similarities corresponding to \(X_i\)\\
        $I_i$ & A set of indices satisfying the similarity score\\
        $\alpha$ &The similarity threshold\\
        $\overline{x}_i$  & The final message representation of \(m_i\)\\
        $B$ &  The training batch size\\
        $H$; $h$ & The center feature representation matrix; The central feature representation\\
        $L_{pair}$ & The pairwise cross-entropy loss\\
        $L_{intra}$ & The intra-cluster loss  \\
        $L_{inter}$ & The inter-cluster loss  \\
        $L$ & Total loss  \\
        $\kappa$; $\lambda$; $\mu$ & Similarity pairwise cross-Entropy loss weight; Intra-cluster loss weight; Inter-cluster loss weight\\
		\hline
    \end{tabularx}
\end{table}

\subsection{Problem and Challenges}
\label{sec: challenges}
When addressing social event detection tasks, the most straightforward approach is to learn representations of social messages and subsequently cluster these representations.
To acquire higher-quality message representations for clustering, social event detection faces the following three main challenges:

\noindent
\textbf{Challenge 1: How to effectively address the issues of missing and noisy edges in the social message graph for social event detection? } 

The issues of missing and noisy edges in the social message graph are multifaceted. 
They include the lack of direct connections between nodes of the same event, connections between nodes of different events, and the presence of isolated nodes. 
The emergence of these issues can be attributed to various factors, such as describing the same event using different vocabulary or sentences, an individual user expressing opinions on other events, the event having just occurred, or being rarely discussed. 
However, GNN-based methods~\cite{amiri2016short, morabia2019sedtwik, sahnoun2020event, wang2017neural, yan2015probabilistic, zhao2011comparing,feng2015streamcube, xie2016topicsketch, xing2016hashtag} learn message representations from explicit message graph. 
Noisy edges in the graph can adversely affect the learned message representations, while the absence of effective edges may result in incomplete message representations. 
Additionally, isolated nodes within the graph cannot be effectively processed.
Hence, effectively addressing the issues of missing and noisy edges in the social message graph remains one of the challenges in social event detection.

\noindent
\textbf{Challenge 2: How to concurrently leverage the content and structural information of messages to learn their representations?} 

GNN-based methods typically initialize message embeddings statically before training, which precludes the acquisition of better semantic embeddings from message content during the training process.
In addition, the representation aggregation process of GNNs can be comprehended as being guided by the edges/relations to determine which message nodes are selected for subsequent aggregation. 
However, the edges themselves do not directly participate in the processing of message representations.
These approaches inherently separate the utilization of message content from structural information, failing to adequately account for the complexity of inter-message relations and the potential interactions between these relations and message content.
Therefore, exploring a new learning mechanism to address the complexities of social streams is imperative, enabling a more effective utilization of both message content and structural information.

\noindent
\textbf{Challenge 3: How to obtain more distinguishable message representations for social event detection?}

In social event detection, events are detected through clustering algorithms after obtaining message representations.
Both distance-based~\cite{macqueen1967some} and density-based~\cite{campello2013density,ester1996density} clustering algorithms fundamentally rely on the Euclidean distance between vectors.
Ensuring sufficient closeness among messages belonging to the same event while maintaining adequate distance from messages of different events can further improve the accuracy of event detection.
Therefore, enhancing the discriminability of message representations is essential.
However, some common optimization methods, such as the triplet loss adopted in KPGNN~\cite{cao2021knowledge} and FinEvent~\cite{peng2022reinforced}, are designed to select one positive and one negative sample for each message. 
The goal is then to ensure that the distance between each message and its positive sample is significantly smaller than its distance to the negative sample.
Alternatively, the pairwise loss proposed in QSGNN~\cite{ren2022known}, which constrains the distance between all negative sample pairs to be greater than that of positive samples, offers more stringent constraints compared to triplet loss. 
Both triplet loss and pairwise loss approaches treat messages as objects to be pushed away or drawn closer, which may not effectively differentiate all event clusters nor guarantee compactness within event clusters.
Thus, it is necessary to design a more effective optimization method to enhance the distinguishability of message representations and thereby improve event detection accuracy.

\subsection{Problem Definition}
\label{subsec: structural information theory}
In this section, we formalize the definitions of social stream, social event, multi-relational message graph, social event detection, and incremental social event detection as follows:

\begin{definition}
\(S=M_0, M_1, \cdots, M_{i-1}, M_i, \cdots\) represents a \textbf{social stream} consisting of continuous time-ordered social message blocks.
\textbf{Message block} \(M_i = \{ m_j \mid 1 \leq j \leq |M_i| \}\) consists all of messages in a time block \([t_i, t_{i+1})\), where \(|M_i|\) is the total number of messages in \(M_i\), and \(m_j\) is a message within \(M_i\).
\end{definition}

\begin{definition}
A \textbf{social event} denoted as \(e = \{m_j \,|\, 1 \leq j \leq |e|\}\) constitutes a collection of interrelated social messages addressing a common real-world occurrence. 
We assume that each social message is affiliated with at most one event.
\end{definition}

\begin{definition}
We define a \textbf{multi-relational message graph} as \( \mathcal{G} = \{ \mathcal{M}, {\{\mathcal{R}_t\} \mid}_{t=1}^T \}\), where \(\mathcal{M}\) is a set of nodes representing social messages, \({\{\mathcal{R}_t\} \mid}_{t=1}^T \) are sets of multi-relational edges with type \(t\). 
\(r_{i,j}^t \in \mathcal{R}_t\) denotes an edge/relation between message \( m_i\) and \(m_j \) with relational type \(t\). 
Relational types may include occurrences of entities, hashtags, user mentions, or message pairs posted within a predefined period. 
\end{definition}

\begin{definition}
Given a message block \(M_i\), an algorithm for \textbf{social event detection} learns a model \(f(M_i; \boldsymbol{\theta}) = E_i\), where \(E_i = \{e_k \,|\, 1 \leq k \leq |E_i|\}\) represents a set of events within \(M_i\).
\(\boldsymbol{\theta}\) denotes the parameters associated with the model \(f\).
\end{definition}

\begin{definition}
Given a social stream \(S\), an \textbf{incremental social event detection} algorithm learns a series of event detection models \(f_0, \cdots, f_{t-w},  f_t,  f_{t + w}, \cdots \), such that \(f_t( M_i; \boldsymbol{\theta}_t, \boldsymbol{\theta}_{t-w}) = E_i\) is the model for all message blocks in \(\{M_i \mid t + 1 \leq i \leq t+w \}\). 
Here, \(E_i =\{e_k \mid 1 \leq k \leq | E_i |\}\) is a set of events contained in the message block \(M_i\), \(w\) is the window size for continuously updating the model, \(\boldsymbol{\theta}_t\) and \(\boldsymbol{\theta}_{t-w}\) are the parameters of \(f_t\) and \(f_{t-w}\), respectively. 
Note that \(f_t\) inherits part of the parameters \(\boldsymbol{\theta}_{t-w}\) from its predecessor \(f_{t-w}\), and it extends and updates the knowledge learned. 
Specifically, \(f_0\), which does not extend any previous model, is referred to as the initial model.
\end{definition}

\begin{figure}[htp]
    \centering
    \includegraphics[width=1\linewidth]{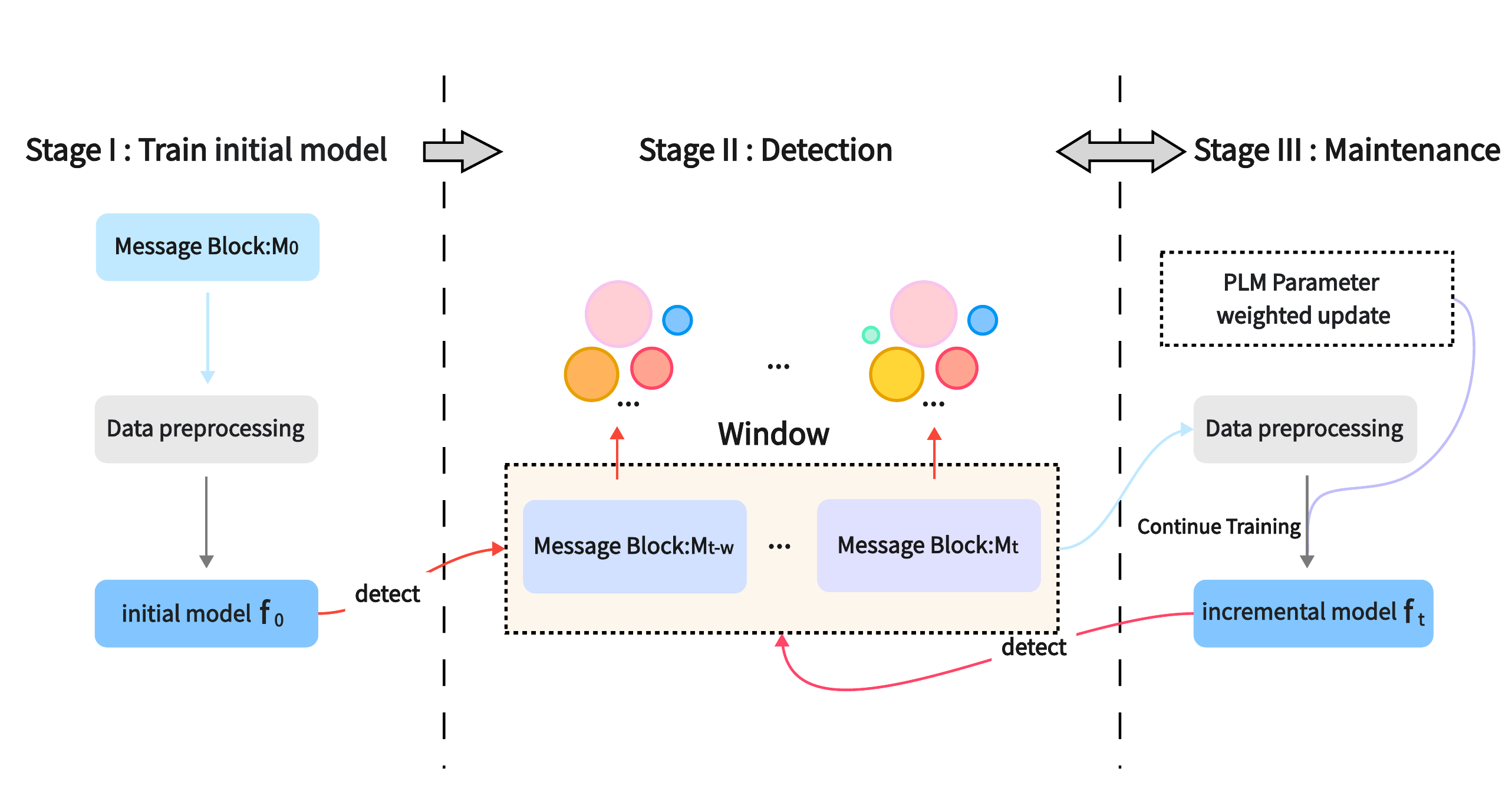}
    \caption{Incremental life-cycle in HP-Event architecture.
    \revised{\textbf{Stage I.} The initial data block \( M_0 \) is utilized to train the initial model.
             \textbf{Stage II.} The model detects events within each message block of the window separately.
             \textbf{Stage III.} The model is maintained through the current window's message blocks. 
             Then, return to the detection phase to detect events of each message block in the next window.
        }
    }
    \label{fig:life-cycle}
\end{figure}

\section{Methodology}
\label{sec: methodology}
This section provides a detailed exposition of the proposed method $\mathrm{RPLM}_{SED}$.
We elucidate the lifecycle of the $\mathrm{RPLM}_{SED}$ model in Section~\ref{Incremental event detection life-cycle}. 
To leverage messages' structure and semantics, we propose a new Pairwise Message Modeling Strategy in Section~\ref{subsec:Pairwise Message Sampling}. 
Section~\ref{sec:proposed_model} and Section~\ref{sec:Optimization Objective} introduce the Multi-relational Prompt-based Pairwise Message Learning Mechanism, and loss function, respectively. 
Additionally, we provide and explain the overall algorithmic process in Section~\ref{subsec:Proposed RPLMSED} and analyze the time complexity of $\mathrm{RPLM}_{SED}$ in Section~\ref{subsec:Time Complexity}.

\begin{figure}[h]
    \centering
    \includegraphics[width=1\linewidth, trim={0.5cm 0 0.4cm 0}, clip]{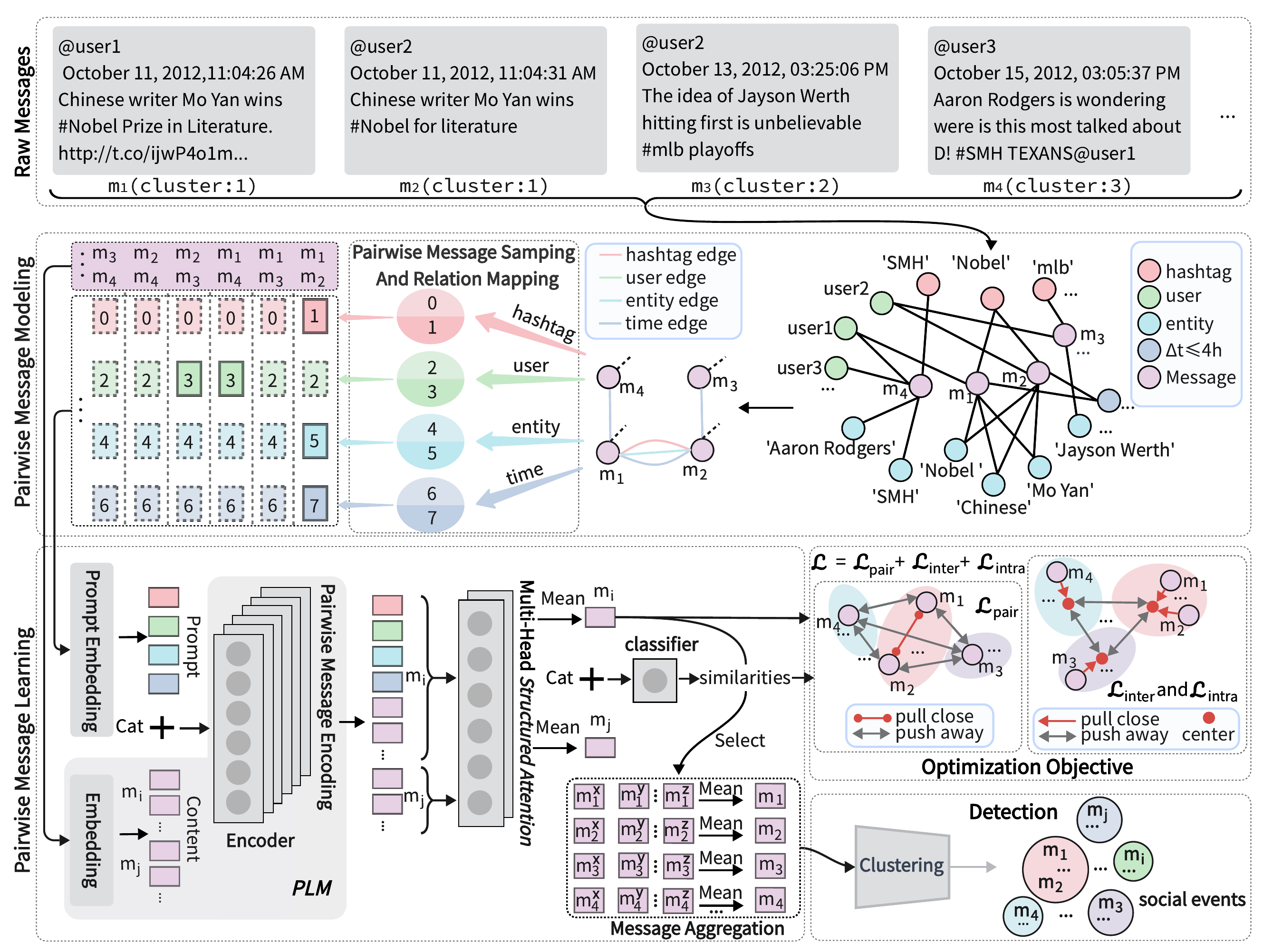}
    \caption{The architecture of the $\mathrm{RPLM}_{SED}$.
    \revised{
    \textbf{Raw Message} shows real messages from Twitter.
    \textbf{Pairwise Message Modeling} (Section \ref{subsec:Pairwise Message Sampling}) initially constructs a multi-relational message graph, then sample massage to form pairs and map relations to the multi-relational sequence.
    \textbf{Pairwise Message Learning} (Section \ref{sec:proposed_model}) embeds and encodes message content and relational sequence, then extracts message representations from the encoding vector. the messages' similarity is assessed by the classifier. 
    In the detection phase, candidate representations of each message are \textbf{selected} to \textbf{aggregate} the final representations.
    \textbf{Optimization Objective} (Section \ref{sec:Optimization Objective}) comprises pairwise cross-entropy, inter-cluster and intra-cluster losses, jointly guiding model training.
        }   
    }
    \label{fig:model}
\end{figure}
\subsection{Incremental event detection life-cycle}
\label{Incremental event detection life-cycle}
As shown in Figure~\ref{fig:life-cycle}, to detect events from social streams incrementally, we propose a three-stage event detection framework, including initial model training, event detection, and  model maintenance:
\begin{enumerate}
    \item \textbf{Initial model training}: we train the initial social event detection model with the first message block \(M_0\) using the proposed $\mathrm{RPLM}_{SED}$. 
    \item \textbf{Event detection}: event detection is performed on incoming message blocks \(M_{t+1}, \cdots, M_{t+w}\) with the trained model \(f_t\). 
    Specifically, we use model \(f_t\) to detect events in each message block separately and evaluate the results.
    \item \textbf{Model maintenance}: the model enters the maintenance stage after each event detection stage. 
    We first perform a weighted fusion of the parameters of PLM in RPLM with the initial parameters of PLM, thereby striking a balance between retaining some existing knowledge and assimilating new knowledge.
    Subsequently, we use message blocks \(M_{t+1}, \cdots, M_{t+w}\) to maintain the model \(f_{t+w}\). 
\end{enumerate}

\subsection{Pairwise Message Modeling Strategy}
\label{subsec:Pairwise Message Sampling}
To jointly model social messages' textual content and structural information, we model social messages into message pairs with multi-relational sequences.
\textbf{Firstly}, we construct a Heterogeneous Information Network (HIN)~\cite{peng2019fine,peng2022reinforced}, where each social message is treated as a message node.
Concurrently, we extract a set of named entities, hashtags, and users (including users and mentioned users) as relational nodes from each message. 
Edges between the message nodes and the corresponding relational nodes denote their associations.
For instance, the message $m_1$ depicted in the `Raw Messages' part of Figure~\ref{fig:model} includes three entities (`Nobel', `Mo Yan', `Chinese'), one user (`user1'), and one hashtag (`Nobel'). 
Edges are established between $m_1$ and its related nodes. 
By merging duplicate nodes, a HIN graph is formed.
In this graph, message nodes, entity nodes, user nodes, and hashtag nodes are represented by $m$, $e$, $u$, and $h$, respectively.
\textbf{Subsequently}, the relational nodes in the heterogeneous information graph are converted into different types of edges between message nodes, thereby constructing a multi-relational message graph that includes single-type nodes and multiple-type relational edges. 
Additionally, edges are added between message nodes published within a 4-hour timeframe to capture temporal correlations. 
\textbf{Finally}, after establishing the multi-relational message graph, we further implement pairwise modeling for message nodes within the graph. 
\revised{For any given message node \( m_i \),  we sample \( y \) positive and negative samples to form message pairs in the training phase and sample \( n \)  nodes in the detection phase, where \( n \) is three times \( y \).
In the training stage, we prefer to select message nodes of the same event with richer connections to \( m_i \) as positive sampling message nodes, whereas messages from different events with sparser connections to  \( m_i \) are chosen as negative sampling message nodes.
This balanced and targeted sampling strategy helps to obtain more valuable sample pairs, enabling the model to learn more effective information during training.
In the detection phase, nodes that have richer connections to \( m_i \) are sampled to form message pairs.}
Considering that the multiple relations between messages may have varying impacts. 
\revised{These multiple relations are preserved comprehensively through a multi-relational sequence.}
We define an external vocabulary to represent multi-relational edges \({\{\mathcal{R}_t\} \mid}_{t=1}^T\) in a multi-relational message graph. 
Each type of relation/edge \(R_t\) occupies two distinct discrete values, indicating whether this relation/edge exists between messages.
\revised{Consequently,
The different states (exist or not) of different relations between messages correspond to unique discrete values.}
As illustrated in the `Pairwise Message Learning'  part of Figure~\ref{fig:model}, given a social message pair \((m_i, m_s)\),  we map the edge relations between them into the corresponding multi-relational sequence as follows:
\begin{equation} \label{eq:(2)}
	\begin{aligned}
		h_{i,s}=
		\begin{cases}
			1 & \text{if } \exists ~ r_{i,s}^h \in \mathcal{R}_h\\
			0 & \text{otherwise} 
		\end{cases}, 
        u_{i,s} &=
		\begin{cases}
			3 & \text{if } \exists ~ r_{i,s}^u \in \mathcal{R}_u\\
			2 & \text{otherwise} 
		\end{cases}, 
		e_{i,s} =
		\begin{cases}
			5 & \text{if } \exists ~ r_{i,s}^e \in \mathcal{R}_e\\
			4 & \text{otherwise} 
		\end{cases}, 
		d_{i,s} =
		\begin{cases}
			7 & \text{if } \exists ~ r_{i,s}^d \in \mathcal{R}_d\\
			6 & \text{otherwise} 
		\end{cases}, \\
      r_{i,s} &= \left\{h_{i,s}, u_{i,s}, e_{i,s}, d_{i,s} \,|\, i,s \in |\mathcal{G}_k[\mathcal{M}]|\right\}.
	\end{aligned}
\end{equation}
Here, $r_{i,s}^h$, $r_{i,s}^u$, $r_{i,s}^e$, and $r_{i,s}^d$ denote the hashtag, user, entity, and temporal edges between message node $m_i$ and $m_s$, respectively. 
Correspondingly, $\mathcal{R}_h$, $\mathcal{R}_u$, $\mathcal{R}_e$, and $\mathcal{R}_d$ represent the sets of hashtag, user, entity, and temporal edges in the multi-relational message graph $\mathcal{G}$. 
$r_{i,s}$ represents the multi-relational sequence between $m_i$ and $m_s$.
We represent one such sample pair and the batch of all nodes in the pairwise modeled multi-relational message graph as follows:
\begin{align}
    d_j &= \{ r_{i,s}~,c_{i,s}~, m_i~, m_s \,|\, m_i, m_s \in \mathcal{G}_k[\mathcal{M}] \} ,\label{eq:(3)}\\
    D_k &= \{ d_j \,|\, 1 \leq j \leq |\mathcal{G}_k[\mathcal{M}]| \}, \label{eq:(4)}
\end{align}
where \( m_s \) is the sampled message node. 
The \( c_{i,s} \) denotes the true cluster relation label for a message pair, which is assigned a value of 1 if the pair of messages belongs to the same event, and 0 otherwise. 
\( r_{i,s} \) represents the multi-relational sequence between \( m_i \) and \( m_s \).
\( \mathcal{G}_k[\mathcal{M}] \) and \( |\mathcal{G}_k[\mathcal{M}]| \) denote the set of message nodes in the multi-relational message graph \( \mathcal{G} \) and its size, respectively.
The \( d_j \) is a data pair of samples, and \( D_k \) is the batch of data after pairwise modeling for the \( k \)-th time block.

\subsection{Multi-relational Prompt-based Pairwise Message Learning Mechanism}
\label{sec:proposed_model}
Owing to current GNN-based SED methods to learn message representations from explicit message relations,  a challenge arises where GNNs cannot directly learn effective knowledge when messages of the same event share no common attributes yet have the same semantics. 
Conversely, when messages from two different events share common attributes, GNN-based methods may introduce noise from irrelevant messages, thereby affecting message representations.
Furthermore, GNNs struggle to handle situations with missing structural relations.
Consequently, we propose a new multi-relational prompt-based pairwise message learning mechanism.
The core of the mechanism lies in harnessing the powerful semantic and contextual relationship understanding abilities of PLMs to learn the deeper connections and differences between messages from both textual content and structural information, thereby obtaining more detailed and comprehensive message representations.
As illustrated in the `Pairwise Message Learning' part of Figure~\ref{fig:model}, this mechanism comprises three key components: \textit{Pairwise Message Encoding}, \textit{Multi-Head Structured Attention}, and \textit{Similarity-based Representation Aggregation}.
The \textit{Pairwise Message Encoding} module employs a PLM as the encoder to learn message representations from both message content embeddings and multi-relational prompt embeddings.
Subsequently, we utilize the \textit{Multi-Head Structured Attention} to extract more critical and clearer message representations.
Finally, the final message representations are obtained through \textit{Similarity-based Representation Aggregation}.

\subsubsection{Pairwise Message Encoding}
To incorporate the multi-relational sequence between messages as prompt information into the encoding process of PLMs, we transform it into multi-relational prompt embeddings. 
Considering that the multi-relational sequence serves as soft prompts, it reveals the relations between messages without directly conveying semantic information.
Sharing the embedding space of the PLMs with message content may blur the semantic boundary between multi-relational prompts and textual content, thereby diminishing their guiding value. 
We design an independent and learnable prompt embedding layer to distinguish between message content embeddings and multi-relational prompt embeddings.
This design prevents potential conflicts in the embedding space, enabling the model to adaptively learn optimized representations of different relational prompts during training, thereby enhancing the model's capability to understand and process complex relationships in social messages.

In the encoding phase, we combine the embeddings of multi-relational prompts with pairwise message content embeddings, delving into the connections and distinctions in semantic and structural information between messages. 
Multiple messages are sampled for each message to form pairs, allowing each message to be repeatedly learned. 
This process enhances the learned message representations, making them more closely related to messages from the same event. 
Conversely, it makes message representations more distinctly differentiated from messages of different events.
The specific encoding operation for the message pair $m_i$ and $m_s$ is as follows:
\begin{equation} \label{eq:(5)}
    x_i, x_s = \mathbf{Extract}(\Phi(\Psi(r_i^s);\Psi^{plm}(m_i,m_s))),
\end{equation}
where $\Phi(\cdot)$ denotes the encoding operation of PLMs, $\Psi(\cdot)$ and $\Psi^{plm}(\cdot,\cdot)$ denote the multi-relational embedding operation and the message pair embedding operation.
The $r_i^s$ represents the multi-relational sequence of the message pair.
The $\mathbf{Extract}(\cdot)$ operation utilizes token type IDs to separately extract the encoded representation $x_i$ of message $m_i$ and the encoded representation $x_s$ for sampled message $m_s$ from the encoded representation of message pair. 
It is essential to note that, to ensure that $x_i$ encompasses both message content and structural information, the multi-relational prompt encoded representation is treated as a part of $x_i$.

\subsubsection{Multi-Head Structured Attention}
To further extract key features from message representations, we introduce the multi-head structured attention mechanism~\cite{vaswani2017attention,kim2017structured}.
This mechanism can attend to different subspaces of the message representations. 
It highlights those message features closely related to the event while ignoring the irrelevant and noisy parts. 
This process is defined as:
\begin{equation} \label{eq:(6)}
    \hat{x}_i= \mathbf{Mean}(\Lambda(x_i,o)), \quad \hat{x}_s =\mathbf{Mean}(\Lambda(x_s,o)).
\end{equation}
Here, $\mathbf{Mean}(\cdot)$ represents the operation of averaging representations, and $\Lambda(\cdot,\cdot)$ denotes the structured attention computation operation, with $o$ being the number of attention heads. 
Finally, the resulting $\hat{x}_i$ and $\hat{x}_s$ are concatenated and fed into a classifier to discern the similarity between the two messages:
\revised{\begin{equation} \label{eq:(7)}
    q_{i,s} = \sigma(\mathbf{W}^\mathsf{T} \cdot (\hat{x}_i;\hat{x}_s) + b),
\end{equation}}
where $\mathbf{W}$ is a weight matrix of dimension $(\cdot,1)$, and $b$ is a bias term.
The linear layer maps the concatenated representation to a scalar, which is then mapped to the interval $(0,1)$ through the sigmoid function $\sigma(\cdot)$, representing the similarity score $q_i$ between two messages. 
Simultaneously, $\hat{x}_i$ will also participate as a candidate representation of the message $m_i$ in the message representation aggregation.
We will introduce this in detail in Section~\ref{subsec:Message Representation Aggregation}.

\subsubsection{Similarity-based Representation Aggregation}
\label{subsec:Message Representation Aggregation}
We propose a similarity-based representation aggregation approach to ensure that the final representation of $m_i$ is as similar as possible to the message representations of the same event. 
The essence of this approach lies in identifying and filtering out ambiguous or noisy candidate message representations generated by negative sample pairs. 
Subsequently, aggregating the candidate message representations extracted from pairs of messages with high similarity makes the resulting final message representation more robust and stable.

In the detection phase, we sample $n$ other message nodes for each message node $m_i$. 
Consequently, for each specific message $m_i$, we obtain a set of $n$ candidate representations \revised{$X_i = [\hat{x}_{i,1}, \hat{x}_{i,2}, \cdots, \hat{x}_{i,n}]$}, along with the corresponding set of similarity \revised{$Q_i = [q_{i,1}, q_{i,2}, \cdots, q_{i,n}]$}. 
Subsequently, select the indices from $Q_i$ that satisfy the conditions and average the candidate representations corresponding to these indices to obtain the final representation of $m_i$.
We formalize this process as follows:
\revised{\begin{equation} \label{eq:(8)}
    \begin{aligned}
        I_i = \{ j \mid q_{i,j} > \alpha , j \in \{1, 2, \ldots, n\}\}, \quad
        \overline{x}_i=\frac{1}{|I_i|} \sum_{j \in I_i} \hat{x}_{i,j}.
    \end{aligned}
\end{equation}}
Here, $I_i$ represents the set of indices obtained by filtering the set $Q_i$, $\alpha$ is the similarity retention threshold, and $\overline{x}_i$ denotes the final representation of message $m_i$.
Particularly, we choose the candidate representation corresponding to the highest similarity index in $Q_i$ as the final representation for those messages in the similarity set $Q_i$ with only one or no indices meeting the condition. 

\subsection{Optimization Objective}
\label{sec:Optimization Objective}
During the model training process, we enforce intra-event message proximity and inter-event message separation through \textit{Pairwise Cross-Entropy Loss}. 
Additionally, to enhance intra-cluster cohesion and inter-cluster dispersion, we design an additional \textit{Clustering Constraint}.
\subsubsection{Pairwise Cross-Entropy Loss}
Considering the pairwise training characteristic of the model, we construct the pairwise cross-entropy loss.
This loss minimizes the divergence between the similarity of messages and the true cluster relationship labels, thereby drawing messages within the same cluster closer while pushing messages from different clusters further apart.
Moreover, the pairwise cross-entropy loss also optimizes the model's classifier, enabling it to predict whether two messages belong to the same event more accurately.
The pairwise cross-entropy loss is defined as:
\begin{equation} \label{eq:(9)}
    L_{pair} = \frac{1}{B} \sum_{i=1}^B c_i \cdot \log(q_i) + (1-c_i) \cdot \log(1-q_i),
\end{equation}
where $B$ denotes the batch size, $c_i$ represents the true cluster relation label of the message pair, and $q_i$ represents the similarity of messages as predicted by the classifier.

\subsubsection{Clustering Constraint}
Given that the pairwise cross-entropy loss treats messages as objects to be pushed away or pulled closer, it may not effectively separate the various event clusters or ensure compactness within clusters.
Hence, we further incorporate a new clustering constraint to enhance the distinguishability of message representations.
This clustering constraint is composed of two parts: intra-cluster loss and inter-cluster loss.
The intra-cluster loss aims to ensure that message representations from the same event are closer.
Specifically, we initialize an updatable central feature matrix, mapping real event labels to the row indices of the matrix, where each row represents the central feature representation of a particular event. 
With the input of message representations, we treat the first incoming message representation as the initial central feature representation of that event category.
In subsequent processes, the central feature representation is updated in a weighted manner.
This process is formulated as follows:
\begin{equation} \label{eq:(10)}
    h_j =
    \begin{cases}
        \hat{x}_i& \text{if } \mathbf{Sum}(\hat{h}_j) = 0 \\
        \beta \cdot \hat{h}_j + (1- \beta) \cdot \hat{x}_i & \text{otherwise}
    \end{cases} , 
\end{equation}
\begin{equation} \label{eq:(11)}
\mathbf{H}_k = [h_{1},h_{2},\cdots, h_{j},\cdots,h_{|E_k|}]^T,
\end{equation}
where $\hat{x}_i$ represents a message representation, and $\beta$ is the retention weight for the central feature. 
$\hat{h}_j$ and $h_j$ are the central feature representations before and after updating, respectively. 
$\mathbf{Sum}(\cdot)$ is the feature summation operation, $\mathbf{H}_k$ is the center feature matrix, and $|E_k|$ is the number of event categories in the current message block. 
After updating the center feature matrix, we average the Euclidean distance between all message representations in the batch with their corresponding central feature representations to derive the final loss value.
The intra-cluster loss is defined as:
\begin{equation} \label{eq:(12)}
    L_{intra} =  \frac{1}{B} \sum_{i=1}^B \mathbf{D}(x_i, h_j),
\end{equation}
where $\mathbf{D}(\cdot,\cdot)$ calculates the euclidean distance. $B$ denotes the batch size, and $h_j$ represents message representation $x_i$ corresponding central feature representation.
Moreover, the inter-cluster loss aims to push apart the centroids of different event clusters.
It is challenging to push away all messages from different clusters for each message, but we can achieve this by pushing apart the centroids of different clusters.
We randomly shuffle the order of the central feature matrix and compute its Euclidean distance to the original matrix, utilizing a distance threshold to keep them apart.
The inter-cluster loss is defined as:
\begin{equation} \label{eq:(13)}
    H_k' = \text{Shuffled}(H_k),  \quad
    L_{inter} =   \frac{1}{|H_k|} \sum_{n = 0}^{|H_k|} \mathbf{Max} \{\gamma - \mathbf{D}(H_{k,n}, H_{k,n}'), 0\}.
\end{equation}
Where $\text{Shuffled}(\cdot)$ denotes the shuffling operation performed on $H_k$ once, i.e., random rearrangement of its order, $\mathbf{D}(\cdot,\cdot)$ calculates the euclidean distance, and the \(\mathbf{Max}\{,\}\) operation selects the larger of two elements.
$\gamma$ represents the distance threshold. 
$H_{k,n}$ and $H_{k,n}'$ represent the $n$-th row of the original central feature matrix and the $n$-th row of the central feature matrix after shuffling the order.
Through the constraints and guidance of inter-cluster and intra-cluster losses, representations of messages within the same cluster become more cohesive, while those of different clusters become more dispersed. 
The pairwise cross-entropy loss, inter-cluster loss, and intra-cluster loss complement each other.
They jointly constrain and guide the model training, thereby achieving higher-quality message representations.
The overall loss is defined as:
\begin{equation} \label{eq:(14)}
L = \kappa \cdot L_{pair} + \lambda \cdot L_{intra} + \mu \cdot L_{inter},
\end{equation}
where $\kappa$, $\lambda$, and $ \mu$ represent the weights of the pairwise cross-entropy loss, intra-cluster loss, and inter-cluster loss, respectively.
\begin{algorithm}[p]
        \caption{\revised{\textbf{$\mathrm{RPLM}_{SED}$}: Relational Prompt-based Pre-trained Language Models for Social Event Detection.}}
        \label{alg:RPLM}
	\SetAlgoLined
	\KwIn{A social stream $S = M_0, M_1, M_2,\cdots$, the ground truth label set: $L = L_0, L_1, L_2, \cdots$, the window size: $w$, and the number of mini-batches $B$.}
	\KwOut{Sets of events: $E_0, E_1, E_2, \cdots$.}
	
	\For{$t=0, 1, 2, \cdots$}{
 
            \tcc{Detect events from $M_t$}
            \If{$t~>~0$}{
                $\mathcal{G} \gets$ Construct multi-relational message graph;\\
                \For{$a=1, 2, 3, \cdots, n$}{
    				$(m_i,m_s) \gets \{m_i, \text{Sample}(\mathcal{G}[\mathcal{M}_t])~|~\forall m_i \in M_t \}$; \\
                        \tcp{Sampling messages that have richer connections with $m_i$ to form pairs.}
                        $r_{i,s} \gets$  Eq.~\ref{eq:(2)};\\
    				$d\gets$ Eq.~\ref{eq:(3)};\\
    				Append $d$ to $D$ (Eq.~\ref{eq:(4)});
                }
                $D, Train = False \Rightarrow$ Execute~\text{Algorithm~\ref{alg:MTD}};
            }
            
            \tcc{Initial model training or the model maintenance.}
            \If{$t \% w == 0$}{
                    \eIf{$t == 0$}{
                        $M, L^{gt} = M_0, L_0$; \tcp{$M_0$ is used to train initial model $f_0$.}
                    }{
                        $M = M_{t-w+1} + \cdots + M_t$; \tcp{ Using messages in the window to maintain the model.}
                        $L^{gt} = L_{t-w+1} + \cdots + L_t$;\\
                        $\theta_t^{PLM} = \zeta \cdot \theta_{t-w}^{PLM} + (1 - \zeta) \cdot \theta_0^{PLM}$;\tcp{Parameter weighted update of PLM in $\mathrm{RPLM}_{SED}$.} 
                        ($\theta_{t-w}^{PLM}$ and $\theta_0^{PLM}$ represent the PLM parameters in $\mathrm{RPLM}_{SED}$ and the initial parameters of the PLM, respectively. $\zeta$ denotes the retention weight threshold.)
                    }
                    
                    \tcc{Pairwise message modeling.}
                    $\mathcal{G} \gets$ Construct multi-relational message graph;\\
                    \For{$a=1, 2, 3, \cdots, y$}{ 
    				$(m_i, m^+_s), (m_i, m^-_s) \gets\{m_i, \text{Sample}(\mathcal{G}[\mathcal{M}], L^{gt})~|~\forall m_i \in M\} $;\\
                        \tcp{Utilizing ground truth labels to choose positive and negative samples.}
                        $r_{i,s} \gets$  Eq.~\ref{eq:(2)};\\
    				$d^+$, $d^- \gets$  Eq.~\ref{eq:(3)};\\
    				Append $d^+$, $d^-$ to $D$ (Eq.~\ref{eq:(4)});
    		      }
                    $D, Train = True \Rightarrow$ Execute~\text{Algorithm~\ref{alg:MTD}}; 
    	}
        }
\end{algorithm}
\begin{algorithm}[h]
        \caption{\revised{Model Training and Detection.}}
	\label{alg:MTD}
	\SetAlgoLined
    \tcc{Multi-relational prompt-based pairwise message learning.}
    \For{$b=1, 2, 3, \cdots, B$}{
                get representations of the message and sampled message from each message pair via Eq.~\ref{eq:(5)},  Eq.~\ref{eq:(6)};\\
                Get the similarity of the two in each message pair via Eq.~\ref{eq:(7)};\\
	   \eIf{Train}{
    		Update central feature matrix $H \gets$ Eq.~\ref{eq:(10)};\\
    		$L_{pair}, L_{intra}, L_{inter} \gets$Eq.~\ref{eq:(9)}, Eq.~\ref{eq:(12)}, Eq.~\ref{eq:(13)};\\
                $L \gets$ Eq.~\ref{eq:(14)};\\
    		Back propagate to update model parameters;
		}{
    		Update sets of each message candidate representations: $X_1, X_2, \cdots, X_i$;\\
                Update corresponding similarity sets: $Q_1, Q_2, \cdots, Q_i$;\\
		}
    }
  \tcc{Similarity-based representation aggregation and Clustering.}
    \If{not Train}{
        \For{$X,Q~in~ [X_1, X_2, X_3, \cdots, X_i], [Q_1, Q_2, Q_3, \cdots, Q_i]$}{
		 $\overline{x}_i \gets$ Eq.~\ref{eq:(8)};\\
		Append final message representation $\overline{x}_i$ into the clustering feature matrix $Matrix$;\\
	}
	$E_t=\{e_0, e_1, e_2, \cdots\} \gets Cluster(Matrix)$;  
    }
\end{algorithm}

\subsection{Proposed $\mathrm{RPLM}_{SED}$}
\label{subsec:Proposed RPLMSED}
As depicted in Algorithm~\ref{alg:RPLM}, the life cycle of $\mathrm{RPLM}_{SED}$ is divided into three stages: the initial training phase, the detection phase, and the maintenance phase. 
During the initial training stage, we train the model using the initial message block (\revised{Algorithm~\ref{alg:RPLM}, lines 17, 25-33}). 
Subsequently, the model performs event detection on message blocks within the window (\revised{Algorithm~\ref{alg:RPLM}, lines 3-13}).
To retain existing knowledge while adapting to new data, we perform a weighted update of the parameters in the PLMs within the model before the maintenance phase (\revised{Algorithm~\ref{alg:RPLM}, line 21}), then maintain the model by utilizing the message blocks within the window (\revised{Algorithm~\ref{alg:RPLM}, lines 19, 20, 25-33}).
Once the model maintenance is completed, it will proceed to detect events in message blocks within the next window.

\revised{Lines 4-11 and 25-32 of Algorithm~\ref{alg:RPLM} separately introduce the pairwise message modeling strategy in the training and detection stages, including message node sampling and multiple relation mapping. 
Lines 2-14 of Algorithm~\ref{alg:MTD} explain the multi-relational prompt-based pairwise message learning mechanism.}
After obtaining message representations (\revised{Algorithm~\ref{alg:MTD}, line 3}) and their respective similarity (\revised{Algorithm~\ref{alg:MTD}, line 4}), during the training phase, we construct the pairwise cross-entropy loss based on the similarity of messages and actual event relations between them.
Additionally, the intra-cluster loss and inter-cluster loss (\revised{Algorithm~\ref{alg:MTD}, line 7}) are constructed by the central feature matrix (\revised{Algorithm~\ref{alg:MTD}, line 6}) and message representations. 
The model is optimized through backpropagation of the total loss (\revised{Algorithm~\ref{alg:MTD}, line 8, 9}). 
During the detection phase, we update the candidate representation set and the similarity set for each message based on the obtained candidate representations of messages and their corresponding similarities. (\revised{Algorithm~\ref{alg:MTD}, lines 11, 12}).
Subsequently, we obtain the final representation of each message through the similarity-based representation aggregation approach (\revised{Algorithm~\ref{alg:MTD}, lines 17-20}).
\revised{Finally, events in a message block are detected, utilizing all messages' final representations through clustering algorithms (Algorithm~\ref{alg:MTD}, line 21).}
Herein, we employ two distinct clustering algorithms: the distance-based  \textit{K-Means}~\cite{macqueen1967some} and the density-based \textit{HDBSCAN}~\cite{campello2013density} algorithms. 
\textit{HDBSCAN} is an enhanced version of the \textit{DBSCAN}~\cite{ester1996density} algorithm, incorporating concepts from hierarchical clustering.
This method does not require pre-specifying the number of clusters. 
Instead, it identifies the optimal clustering outcome by constructing a density hierarchy. 
Compared to the original \textit{DBSCAN}, \textit{HDBSCAN} offers greater flexibility in parameter settings, effectively adapts to data from different density regions, and exhibits stronger robustness. 
In practice, density-based clustering algorithms are more suitable as they do not require pre-setting the number of clusters. 
This allows them to better adapt to incremental data changes and meet the needs for continuous or dynamic detection.

\subsection{Time Complexity}
\label{subsec:Time Complexity}
\revised{The time complexity of $\mathrm{RPLM}_{SED}$ consists of pairwise message modeling, multi-relational prompt-based pairwise message learning mechanism, and message representation clustering.
During pairwise message modeling,  constructing a HIN graph and transforming it into a multi-relational message graph necessitates traversing all edges, resulting in a time complexity of $O(N_e)$, where $N_e$ is the number of edges in the graph. 
The time complexity for sampling messages to form pairs is $O(N \cdot n) = O(N)$, where $N$ is the number of messages in the graph, and $n$ is the number of message pairs sampled for each message.
Thus, the total time complexity of this phase is $O(N_e + N) = O(N_e)$.
In the multi-relational prompt-based pairwise message learning mechanism, firstly, embedding operations for message pair content and multi-relation sequences require time $O(N \cdot n \cdot s) = O(N)$, where $s$ is the average input sequence length.
Secondly, the PLM comprises multiple layers of \textit{Transformers}, where each layer's primary source of time complexity stems from the self-attention mechanism. 
The time complexity in the encoding phase is $O(N \cdot n \cdot s^2 ) = O(N \cdot s^2)$.
Notably, as the scale of PLM increases, higher layer counts and hidden dimensions lead to increased time complexity.
Thirdly, the time complexities of the structured attention mechanism and classifier are $O(N \cdot n \cdot s ) = O(N)$ and $O(N)$, respectively.
Finally, the time complexity for selecting and aggregating final message representations is $O(N \cdot n) = O(N)$. 
Therefore, the overall time complexity of the pairwise message learning mechanism is $O(N + N \cdot s^2 + N + N + N) = O(N \cdot s^2)$.
In the clustering stage, the time complexity for clustering message representations using the K-means algorithm is $O(I \cdot K \cdot N \cdot d) = O(N)$, where $I$ and $K$ denote the number of iterations and the given number of clusters, respectively.
In contrast, the time complexity of clustering with the HDBSCAN algorithm is $O(N \cdot \log N)$.}

\section{Experimental Setup}
\label{sec: Experimental Setup}
This section provides a detailed overview of the experimental setup, including dataset selection, comparison baselines, evaluation metrics, and experiment parameter configurations.
Specifically, in Section~\ref{sec:Datasets}, we describe the details of the datasets employed in this study.
In Section~\ref{sec: baselines}, we list and discuss all the baseline methods included in the comparison, aiming to conduct a comprehensive comparative analysis. 
Furthermore, in Section~\ref{sec: setting}, we present the hardware and software configurations for our experiments. 
We also offer a comprehensive description of the parameter settings for $\mathrm{RPLM}_{SED}$ and the implementation approaches for baseline methods, ensuring the reproducibility of our experimental results.
Finally, Section~\ref{sec:Evaluation Metrics} elaborates on the evaluation metrics adopted for the experiments.

\subsection{Datasets}
\label{sec:Datasets}
As shown in Table~\ref{tab:statisitcs}, we evaluate the performance of the $\mathrm{RPLM}_{SED}$ model on three publicly available social event datasets: Events2012, Events2018, and Arabic-Twitter. 
We test the model's performance in offline and incremental scenarios using the Events2012 and Events2018 datasets and explore the model's performance under low-resource language conditions using the Arabic Twitter dataset. 
Moreover, to observe the model's performance on long-tail distribution issues, we extracted 100 event categories from both the Events2012 and Events2018 datasets. 
Considering that for the long-tail recognition task, an imbalanced training set alongside balanced validation and test sets should be provided, we select an additional 20 and 30 tweets for each event for the validation and test sets, respectively.
After filtering out unavailable and duplicate tweets, the detailed information of these datasets is as described:
\begin{itemize}
    \item \textbf{\textit{Events2012}}~\cite{mcminn2013building}. 
    The Events2012 dataset contains 68,841 annotated English tweets covering 503 different event categories, encompassing tweets over a consecutive 29-day period.
    The \textbf{\textit{Events2012\_100}} contains 100 events with a total of 15,019 tweets, where the maximal event comprises 2,377 tweets, and the minimally has 55 tweets, with an imbalance ratio of approximately 43.
    
    \item \textbf{\textit{Events2018}}~\cite{mazoyer2020french}. 
    The Events2018 includes 64,516 annotated French tweets covering 257 different event categories, with data spanning over a consecutive 23-day period.
    The \textbf{\textit{Events2018\_100}} contains 100 events with a total of 19,944 tweets, where the maximal event comprises 4,189 tweets and the minimally has 27 tweets, an imbalance ratio of approximately 155.
    \item \textbf{\textit{Arabic-Twitter}}~\cite{alharbi2021kawarith}. The Arabic-Twitter dataset comprises 9,070 annotated Arabic tweets, covering seven catastrophic-class events from various periods.
\end{itemize}
\begin{table}[h]
    \centering
    \begin{threeparttable}
        \caption{Statistics of the datasets.}
        \label{tab:statisitcs}
        \begin{tabularx}{\textwidth}{c|*6{>{\centering\arraybackslash}X}}
        \toprule
        \makecell{Dataset\\[1ex]} & \makecell{Messages\\[1ex]} & \makecell{Events\\[1ex]} & \makecell{Timespan\\[1ex]} & \makecell{\revised{Max. tweet}\\ \revised{count/Event}} & \makecell{\revised{Min. tweet}\\ \revised{count/Event}} & \makecell{\revised{Avg. tweet}\\ \revised{count/Event}} \\
        \midrule
         Events2012 & 68841 & 503 & 29 days& 7133 & 1 & 137\\ 
        Events2018 & 64516 & 257 & 23 days &12567 & 1 & 251\\ 
        Events2012\_100 &15019 & 100 & - & 2377 & 55 & 150\\  
        Events2018\_100 &19944 & 100 & - & 4189 & 27 & 199\\ 
        Arabic-Twitter & 9070 & 7& - & 2883 & 519 & 1296\\  
        \bottomrule
        \end{tabularx}
    \end{threeparttable}
\end{table}

\subsection{Baselines}
\label{sec: baselines}

To verify the effectiveness of our proposed $\mathrm{RPLM}_{SED}$, we compare it with popular social event detection methods in recent years.
These baseline methods include:

\begin{itemize}
    \item \textbf{PP-GCN}~\cite{peng2019fine}: 
    PP-GCN is an offline fine-grained social event detection method based on Graph Convolutional Networks (GCN)~\cite{kipf2016semi}.
    \item \textbf{EventX}~\cite{liu2020story}: 
    EventX is a fine-grained event detection method based on community detection suitable for online scenarios.
    \item \textbf{KPGNN}~\cite{cao2021knowledge}: 
    KPGNN is a knowledge-preserving incremental social event detection framework based on heterogeneous GNN, leveraging the inductive learning capability of GNN to represent and detect events effectively.
    \item \textbf{QSGNN}~\cite{ren2022known}: 
    QSGNN focuses on open-set social event detection tasks. It extends knowledge from known to unknown by utilizing the best-known samples (i.e., setting stricter constraints in inter-class distances and directional relationships) and reliable knowledge transfer (generating and selecting high-quality pseudo labels).
    \item \textbf{FinEvent}~\cite{peng2022reinforced}: 
    FinEvent is an event detection framework that combines reinforcement learning with GNNs, achieving incremental and cross-lingual social event detection. 
    Concurrently, this framework proposes the Deep Reinforcement Learning Guided DBSCAN Model (DRL-DBSCAN)~\cite{zhang2022automating}. 
    In this paper, the FinEvent baseline, based on the K-means clustering method, is denoted as \textbf{FinEven$\mathrm{t}^{\mathrm{k}}$}, while the FinEvent baseline leveraging DGL-DBSCAN is referred to as \textbf{FinEven$\mathrm{t}^{\mathrm{d}}$}.
    \revised{\item \textbf{CLKD}~\cite{ren2021transferring}: 
    CLKD is a cross-lingual social event detection framework based on knowledge distillation, jointing knowledge distillation, and GNNs to improve social event detection performance in low-resource language. 
    \item \textbf{ETGNN}~\cite{ren2022evidential}: 
    ETGNN is an evidential temporal-aware graph neural network for social event detection, which integrates view-specific graph quality and temporal information uncertainty into social event detection.
    \item $\mathrm{\textbf{UCL}}_{\textbf{SED}}$~\cite{ren2023uncertainty}:
    $\mathrm{UCL}_{SED}$ is an uncertainty-guided class imbalance learning framework for imbalanced social event detection tasks. 
    It focuses on boundary learning in latent space and classifier learning with high-quality uncertainty estimation.
    \item \textbf{MVGAN}~\cite{cui2021mvgan}: 
    MVGAN is a multi-view graph attention network for social event detection. 
    It enriches event semantics through node aggregation and multi-view fusion in heterogeneous message graphs.}

\end{itemize}

Additionally, we enumerate the methods employed in previous works that utilize word embedding models and pre-trained models for event detection while also proposing new approaches for implementing incremental social event detection  through PLMs:

\begin{itemize}
    \item \textbf{Word2Vec}~\cite{mikolov2013efficient}: 
    Word2Vec is a pre-trained word embedding technique. It employs the average of the embeddings of all words within a message to serve as the message embedding.
    \item \textbf{BERT}~\cite{devlin-etal-2019-bert}: 
    BERT is a pre-trained model for natural language processing (NLP) proposed by researchers at Google in 2018.
    The cornerstone of this model's innovation lies in its utilization of the encoder architecture from the Transformer model to train context-based word embeddings, incorporating a Masked Language Model (MLM) design.
    Consistent with previous methodologies, we employ BERT to obtain message representations for clustering directly.
    \textbf{BERT-FT} achieves streaming event detection by applying fine-tuning to BERT.
    Given the inapplicability of cross-entropy in open event sets, the encoding process is optimized through online triplet loss and the implementation of proposed clustering constraints.
    Analogous to standard incremental event detection \revised{methods}, BERT-FT, after its training or maintenance, detects events within message blocks inside the window.
    \item \textbf{RoBERTa}~\cite{liu2019roberta}: 
    RoBERTa is an enhanced version of the BERT model,  introduced by the Facebook AI research team in 2019. 
    It is trained using larger datasets, increased batch sizes, and more granular parameter tuning. 
    Compared to BERT, this model adopts a dynamic masking strategy and implements more refined adjustments to hyperparameters. 
    Similarly, we utilize RoBERTa to encode messages for clustering directly. 
    \textbf{RoBERTa-FT} shares the same training approach as BERT-FT, with the distinction between them lying solely in using different PLMs.
    \item \textbf{BERT-LR}: 
    BERT-LR is a variant of $\mathrm{RPLM}_{SED}$ that fine-tunes BERT within $\mathrm{RPLM}_{SED}$ using Low-Rank Adaptation (LoRA)~\cite{Edward2022Lora}, an efficient fine-tuning technique for PLMs. 
    The core concept of this approach is to introduce low-rank weights into the parameters of the pre-trained model to balance parameter efficiency and performance during model fine-tuning.
    \textbf{RoBERTa-LR} is another variant of $\mathrm{RPLM}_{SED}$, which, in contrast to BERT-LR, is based on RoBERTa.
\end{itemize}

\subsection{Experimental Setting and Implementation}
\label{sec: setting}
In the English dataset Events2012, we select the RoBERTa (RoBERTa-large)~\cite{liu2019roberta} released by Facebook AI in 2019 as the backbone architecture for the $\mathrm{RPLM}_{SED}$ model. 
In the French dataset Events2018 and the Arabic dataset Arabic-Twitter, we use the multilingual version of the TwHIN-BERT (TwHIN-BERT-large)~\cite{zhang2022twhin}.
TwHIN-BERT is a multilingual tweet language model that is trained on 7 billion Tweets from over 100 distinct languages. 
It is trained with not only text-based self-supervision (e.g., MLM) but also with a social objective based on the rich social engagements within a Twitter Heterogeneous Information Network (TwHIN). 
During the data preprocessing phase, for each message in the training set, we sample 60 positive and negative samples. 
For each message in the validation and test sets, 180 samples are sampled. 
The experimental setup includes setting the number of training epochs to 2, the window size to 3, the learning rate to $2e-5$, and the batch size to 35.
During the model maintenance phase, the parameter update weight $\zeta$ is set to 0.4, the inter-cluster loss distance threshold $\gamma$ is defined as 10, and the loss weights $\kappa$, $\lambda$, and $\mu$ are set to 1, 0.01, and 0.005, respectively.
The similarity threshold $\alpha$ for the message feature aggregation stage is set to 0.9, and the number of heads in structured attention mechanism $o$ is set to 2.
The embedding and encoding dimensions of the model are consistent with the chosen PLM. 
In terms of optimizer selection, the entire model is trained using Adam. 
All experiments are repeated 5 times to ensure the stability of the results, and the mean and standard deviation of the experiments are reported.

We implement all models using Python 3.8 and PyTorch 2.0. 
All experiments are conducted on a 32-core Intel Core i9-13900K@3.80GHz, with 128GB RAM and an NVIDIA GeForce RTX 4090 GPU. 
For PP-GCN\footnote{\url{https://github.com/RingBDStack/PPGCN}}, KPGNN\footnote{\url{https://github.com/RingBDStack/KPGNN}}, QSGNN\footnote{\url{https://github.com/RingBDStack/open-set-social-event-detection}} and FinEvent\footnote{\url{https://github.com/RingBDStack/FinEvent}}, we follow their open-source implementations.
We implement EventX with Python 3.7. 
To ensure fairness in comparison, unless specifically noted, we uniformly apply the K-Means~\cite{macqueen1967some} clustering algorithm to all baselines \revised{and $\mathrm{RPLM}_{SED}$}, excluding EventX, setting the total number of categories equal to the number of event categories in the real world.
Additionally, we train the FinEvent framework from scratch on the French dataset, rather than utilizing its proposed cross-lingual social message representation learning method, `Crlme'. 
Furthermore, for the DGL-DBSCAN method proposed by FinEvent, we conduct a more comprehensive comparison by further clustering the message representations generated by $\mathrm{RPLM}_{SED}$ through the HDBSCAN~\cite{campello2013density} clustering method, and conduct a comprehensive comparison and analysis of the results.
\revised{Where the minimum cluster size parameter of HDBSCAN is set to 8.}

\subsection{Evaluation Metrics}
\label{sec:Evaluation Metrics}
To evaluate the effectiveness of $\mathrm{RPLM}_{SED}$, we adopt three widely-used metrics to assess the quality of clustering: Normalized Mutual Information (NMI)~\cite{estevez2009normalized}, Adjusted Mutual Information (AMI)~\cite{vinh2010information}, and the Adjusted Rand Index (ARI)~\cite{vinh2010information}. 
The NMI metric is employed to gauge the clustering quality, primarily focusing on the congruence between our model's clustering output and the actual ground truth categories:
\begin{equation}
   \text{NMI} = \frac{2 \cdot \text{MI}(P, T)}{H(P) + H(T)},
\end{equation}
where \(P\) and \(T\) represent the predicted label set and the true label set, respectively. \(MI(P, T)\) denotes the mutual information between them, and \(H(p)\) and \(H(T)\) respectively denote their entropies.
AMI represents an enhancement of the NMI metric. 
This improvement is achieved by adjusting for the expected mutual information derived from random label assignments.
Therefore, AMI provides values that are closer to 0 in scenarios where labeling is performed randomly, thereby offering a more accurate assessment of the information shared between clusters:
\begin{equation}
   \text{AMI} = \frac{\text{MI}(P, T) - \mathbb{E}[\text{MI}(P, T)]}{\max\{H(P), H(T)\} - \mathbb{E}[\text{MI}(P, T)]},
\end{equation}
where \(\mathbb{E}[\text{MI}(P, T)]\) represents the expected value of the mutual information between \(P\) and \(T\).
The Rand Index measures the similarity between two data clustering outcomes. 
It is particularly used to compare the similarity between algorithmic clustering results and benchmark clusterings. 
The Rand Index values range from 0 to 1, where 1 indicates identical clustering. 
ARI improves upon the original Rand Index by ensuring the expected value of random label assignments is 0 and adjusts the range to be between -1 and 1:
\begin{equation}
    \text{ARI} = \frac{\text{RI} - \text{ERI}}{\text{MaxRI} - \text{ERI}},
\end{equation}
where \(\text{RI}\) is the original Rand Index, \(\text{ERI}\) represents the Expected Rand Index, and \(\text{MaxRI}\) denotes the maximum value in the Rand Index.

\section{Results And Discussion}
\label{sec: Results And Discussion}

This section primarily revolves around several key questions to empirically evaluate the performance of the $\mathrm{RPLM}_{SED}$ model:
\begin{itemize}
    \item Q1:
     How does the performance of the proposed PLM-based method $\mathrm{RPLM}_{SED}$ compare to current mainstream SED methods in both offline and online detection scenarios (Sect~\ref{sec: effectiveness})?
    \item Q2:
     How do different models fare in the low-resource scenario (Section~\ref{sec:Low-Resource Language})?
     \item Q3:
     How do different models address the long-tail distribution issue of social data (Section~\ref{sec:Long-tail Recognition Task})?    
     \item Q4:
     How do various PLM-based approaches perform in online scenarios (Section~\ref{sec:PLM-based Model for Social Event Detection})?
    \item Q5:
     How does the performance of the $\mathrm{RPLM}_{SED}$ model, based on different PLMs, manifest (Section~\ref{subsec:RPLM-Event with Different PLMs})?
     \item Q6:
     \revised{How does the time consumption of $\mathrm{RPLM}_{SED}$ based on different scale PLMs (Section~\ref{sec: report time})?}
    \item Q7:
     How do different components and hyperparameters in $\mathrm{RPLM}_{SED}$ influence overall effectiveness (Section~\ref{sec:Ablation Study}, Section~\ref{subsec:Hyperparameter Study})?
    \item Q8:
     How does the performance of $\mathrm{RPLM}_{SED}$ compare to the strongest baseline in a specific scenario (Section~\ref{sec: case study})?
    \item Q9:
     \revised{How does the performance of  $\mathrm{RPLM}_{SED}$ based on different cluster algorithms (Section~\ref{sec: Clustering Algorithm Study})? }
\end{itemize}

\subsection{Model Effectiveness}
\label{sec: effectiveness}
In this section, we compare and analyze $\mathrm{RPLM}_{SED}$ and all baseline methods under offline and online detection scenarios. 
To ensure consistency in the evaluation, $\mathrm{RPLM}_{SED}^\mathrm{k}$ is compared with baseline methods that employ the K-Means clustering approach.
In contrast, $\mathrm{RPLM}_{SED}^\mathrm{d}$ is separately compared with FinEven$\mathrm{t}^{\mathrm{d}}$.
\begin{table}[h]
    \centering
    \setlength{\tabcolsep}{2pt} 
    \begin{threeparttable}
        \caption{Evaluation on the Closed Set. The best results are bolded, and the second-best results are underlined.}
        \label{ClosedSet}
        \begin{tabularx}{\textwidth}{c|c|*{8}{>{\centering\arraybackslash}X}|c}
        \toprule
         Blocks &Metrics & PP-GCN & EventX & KPGNN & QSGNN & \text{FinEven$\mathrm{t}^{\mathrm{k}}$}  & \text{FinEven$\mathrm{t}^{\mathrm{d}}$} & $\mathrm{RPLM}_{SED}^\mathrm{k}$ & $\mathrm{RPLM}_{SED}^\mathrm{d}$ & Improve\\
        \midrule
        \multirow{3}{*}{\makecell{$M_0$ in \\Events\\2012}}
        & NMI & $.70 \pm .02$ & $.68 \pm .00$ & $.76 \pm .02$ & $.79 \pm .01$ & $.84 \pm .01$ & $\underline{.86 \pm .01}$ & \underline{$.86 \pm .01$} & $\textbf{.87} \pm \textbf{.01}$ & $\uparrow .01$ \\ 
       & AMI & $.56 \pm .01$ & $.29 \pm .00$ & $.64 \pm .02$ & $.68 \pm .01$ & $.72 \pm .01$ & $.77 \pm .01$ & $\underline{.81 \pm .00}$ & $\textbf{.84} \pm \textbf{.00}$ & $\uparrow .07$ \\ 
       & ARI & $.20 \pm .01$ & $.05 \pm .00$ & $.34 \pm .01$ & $.38 \pm .01$ & $.40 \pm .02$ & $\underline{.56 \pm .02}$ & $.41 \pm .01$ & $\textbf{.80} \pm \textbf{.02}$ & $\uparrow .24$ \\ 
        \midrule
        \multirow{3}{*}{\makecell{$M_0$ in \\Events\\2018}}
       & NMI & $.60 \pm .01$ & $.57 \pm .00$ & $.66 \pm .03$ & $.71 \pm .02$ & $.73 \pm .02$ & $.78 \pm .01$ & $\textbf{.93} \pm \textbf{.00}$ & $\underline{.92 \pm .00}$ & $\uparrow .15$ \\ 
       & AMI & $.49 \pm .02$ & $.56 \pm .00$ & $.60 \pm .02$ & $.64 \pm .02$ & $.69 \pm .01$ & $.74 \pm .01$ & $\textbf{.92} \pm \textbf{.01}$ & $\underline{.91 \pm .01}$ & $\uparrow .18$ \\ 
       & ARI & $.17 \pm .01$ & $.01 \pm .00$ & $.27 \pm .03$ & $.33 \pm .02$ & $.35 \pm .02$ & $.61 \pm .02$ & $\underline{.94 \pm .01}$ & $\textbf{.95} \pm \textbf{.01}$ & $\uparrow .34$ \\ 
        \bottomrule
        \end{tabularx}
    \end{threeparttable}
\end{table}
\begin{table}[thp]
    \centering
    \begin{threeparttable}
        \setlength{\tabcolsep}{3pt} 
        \caption{Incremental evaluation NMIs on Events2012. The best results are bolded, and the second-best results are underlined.}
        \label{NMI2012}
        \begin{tabularx}{\textwidth}{c|*{8}{>{\centering\arraybackslash}X}|c}
        \toprule
        Blocks  & PP-GCN & EventX & KPGNN & QSGNN & \text{FinEven$\mathrm{t}^{\mathrm{k}}$}  & \text{FinEven$\mathrm{t}^{\mathrm{d}}$} & $\mathrm{RPLM}_{SED}^\mathrm{k}$ & $\mathrm{RPLM}_{SED}^\mathrm{d}$ & Improve\\
        \midrule
         $M_1$ & $.23 \pm .00$ & $.36 \pm .00$ & $.39 \pm .00$ & $.43 \pm .01$ & $.38 \pm .01$ & $\underline{.84 \pm .01}$ & $.46 \pm .01$ & $\textbf{.91} \pm \textbf{.02}$ & $\uparrow .07$ \\ 
        $M_2$ & $.57 \pm .00$ & $.68 \pm .00$ & $.79 \pm .01$ & $.81 \pm .02$ & $.81 \pm .00$ & $.84 \pm .01$ & $\underline{.89 \pm .02}$ & $\textbf{.91} \pm \textbf{.01}$ & $\uparrow .07$ \\ 
        $M_3$ & $.55 \pm .01$ & $.63 \pm .00$ & $.67 \pm .00$ & $.78 \pm .01$ & $.71 \pm .01$ & $\underline{.89 \pm .00}$ & $.88 \pm .00$ & $\textbf{.93} \pm \textbf{.00}$ & $\uparrow .04$ \\ 
        $M_4$ & $.46 \pm .01$ & $.63 \pm .00$ & $.67 \pm .02$ & $.71 \pm .02$ & $.71 \pm .01$ & $.71 \pm .01$ & $\underline{.78 \pm .01}$ & $\textbf{.83} \pm \textbf{.01}$ & $\uparrow .12$ \\ 
        $M_5$ & $.48 \pm .00$ & $.59 \pm .00$ & $.73 \pm .00$ & $.75 \pm .00$ & $.76 \pm .00$ & $.83 \pm .00$ & $\textbf{.85} \pm \textbf{.01}$ & $\underline{.85 \pm .02}$ & $\uparrow .02$ \\ 
        $M_6$ & $.57 \pm .01$ & $.70 \pm .00$ & $.82 \pm .01$ & $.83 \pm .01$ & $.84 \pm .00$ & $.83 \pm .00$ & $\underline{.87 \pm .01}$ & $\textbf{.92} \pm \textbf{.00}$ & $\uparrow .08$ \\ 
        $M_7$ & $.37 \pm .00$ & $.51 \pm .00$ & $.55 \pm .01$ & $.57 \pm .01$ & $.56 \pm .00$ & $\underline{.73 \pm .01}$ & $.68 \pm .01$ & $\textbf{.88} \pm \textbf{.01}$ & $\uparrow .15$ \\ 
        $M_8$ & $.55 \pm .02$ & $.71 \pm .00$ & $.80 \pm .00$ & $.79 \pm .01$ & $.87 \pm .01$ & $.87 \pm .02$ & $\underline{.88 \pm .01}$ & $\textbf{.88} \pm \textbf{.00}$ & $\uparrow .01$ \\ 
        $M_9$ & $.51 \pm .00$ & $.67 \pm .00$ & $.74 \pm .02$ & $.77 \pm .02$ & $.78 \pm .02$ & $.79 \pm .01$ & $\underline{.87 \pm .01}$ & $\textbf{.92} \pm \textbf{.01}$ & $\uparrow .13$ \\ 
        $M_{10}$ & $.55 \pm .02$ & $.68 \pm .00$ & $.80 \pm .01$ & $.82 \pm .02$ & $.81 \pm .01$ & $.82 \pm .01$ & $\underline{.90 \pm .00}$ & $\textbf{.91} \pm \textbf{.01}$ & $\uparrow .09$ \\ 
        $M_{11}$ & $.50 \pm .01$ & $.65 \pm .00$ & $.74 \pm .01$ & $.75 \pm .01$ & $.76 \pm .00$ & $.75 \pm .00$ & $\underline{.87 \pm .01}$ & $\textbf{.88} \pm \textbf{.01}$ & $\uparrow .12$ \\ 
        $M_{12}$ & $.45 \pm .01$ & $.61 \pm .00$ & $.68 \pm .01$ & $.70 \pm .01$ & $.76 \pm .01$ & $.67 \pm .01$ & $\underline{.85 \pm .01}$ & $\textbf{.92} \pm \textbf{.00}$ & $\uparrow .16$ \\ 
        $M_{13}$ & $.47 \pm .01$ & $.58 \pm .00$ & $.69 \pm .01$ & $.68 \pm .02$ & $.67 \pm .00$ & $\underline{.79 \pm .00}$ & $.77 \pm .01$ & $\textbf{.91} \pm \textbf{.01}$ & $\uparrow .12$ \\ 
        $M_{14}$ & $.44 \pm .01$ & $.57 \pm .00$ & $.69 \pm .00$ & $.68 \pm .01$ & $.74 \pm .00$ & $\underline{.82 \pm .00}$ & $.82 \pm .01$ & $\textbf{.88} \pm \textbf{.00}$ & $\uparrow .06$ \\ 
        $M_{15}$ & $.39 \pm .01$ & $.49 \pm .00$ & $.58 \pm .00$ & $.59 \pm .01$ & $.64 \pm .00$ & $.69 \pm .01$ & $\underline{.75 \pm .01}$ & $\textbf{.83} \pm \textbf{.01}$ & $\uparrow .14$ \\ 
        $M_{16}$ & $.55 \pm .01$ & $.62 \pm .00$ & $.79 \pm .01$ & $.78 \pm .01$ & $.80 \pm .00$ & $\underline{.90 \pm .01}$ & $.84 \pm .00$ & $\textbf{.93} \pm \textbf{.01}$ & $\uparrow .03$ \\ 
        $M_{17}$ & $.48 \pm .00$ & $.58 \pm .00$ & $.70 \pm .01$ & $.71 \pm .01$ & $.73 \pm .00$ & $\underline{.83 \pm .00}$ & $.80 \pm .01$ & $\textbf{.86} \pm \textbf{.01}$ & $\uparrow .03$ \\ 
        $M_{18}$ & $.47 \pm .01$ & $.59 \pm .00$ & $.68 \pm .02$ & $.70 \pm .01$ & $.72 \pm .01$ & $.74 \pm .01$ & $\underline{.82 \pm .01}$ & $\textbf{.83} \pm \textbf{.00}$ & $\uparrow .09$ \\ 
        $M_{19}$ & $.51 \pm .02$ & $.60 \pm .00$ & $.73 \pm .01$ & $.73 \pm .00$ & $.76 \pm .02$ & $.66 \pm .01$ & $\underline{.88 \pm .01}$ & $\textbf{.91} \pm \textbf{.00}$ & $\uparrow .15$ \\ 
        $M_{20}$ & $.51 \pm .01$ & $.67 \pm .00$ & $.72 \pm .02$ & $.73 \pm .02$ & $.73 \pm .00$ & $.80 \pm .00$ & $\textbf{.85} \pm \textbf{.00}$ & $\underline{.82 \pm .01}$ & $\uparrow .05$ \\ 
        $M_{21}$ & $.41 \pm .02$ & $.53 \pm .00$ & $.60 \pm .00$ & $.61 \pm .01$ & $.65 \pm .01$ & $\underline{.74 \pm .01}$ & $\textbf{.76} \pm \textbf{.00}$ & $.74 \pm .02$ & $\uparrow .02$ \\
        \bottomrule
        \end{tabularx}
    \end{threeparttable}
\end{table}

\subsubsection{Offline Evaluation}
In the offline detection scenario, the training, validation, and test sets share the same events.
For the Events2012 and Events2018 datasets, experiments are conducted using data from the first week, with 70\% for training, 20\% for testing, and 10\% for validation.
As illustrated in Table \ref{ClosedSet}, the $\mathrm{RPLM}_{SED}$ outperforms all baseline methods across all metrics (NMI, AMI, ARI). 
Compared to EventX, on the Events2012 dataset, $\mathrm{RPLM}_{SED}^\mathrm{k}$ shows improvements of 26\%, 180\%, and 710\% in NMI, AMI, and ARI, respectively.
On the Events2018 dataset, $\mathrm{RPLM}_{SED}^\mathrm{k}$ exhibits enhancements of 64\%, 64\%, and 9350\% in NMI, AMI, and ARI, respectively. 
$\mathrm{RPLM}_{SED}$ significantly outperforms EventX on both datasets, especially in terms of ARI, indicating that EventX tends to produce a higher number of clusters than the actual events. 
In contrast, $\mathrm{RPLM}_{SED}$ demonstrates greater stability.
Compared with PP-GCN, KPGNN, QSGNN, and FinEven$\mathrm{t}^{\mathrm{k}}$, $\mathrm{RPLM}_{SED}^\mathrm{k}$ shows significant improvements on the Events2012 dataset, with increases of 2\%-23\% in NMI, 13\%-45\% in AMI, and 1\%-103\% in ARI.
Furthermore, on the Events2018 dataset, improvements include 27\% to 55\% in NMI, 33\% to 87\% in AMI, and 170\% to 456\% in ARI.
Compared to FinEven$\mathrm{t}^{\mathrm{d}}$ on the Events2012 dataset, $\mathrm{RPLM}_{SED}^\mathrm{d}$ exhibits improvements of 1\%, 9\%, and 43\% in NMI, AMI, and ARI, respectively. 
On the Events2018 dataset, the increases are 18\%, 23\%, and 56\% in NMI, AMI, and ARI, respectively.
The superior performance of $\mathrm{RPLM}_{SED}$ compared to GNN-based methods can be attributed to its approach of not relying on explicit structural relations for representation learning, which also enhances its robustness. 
Furthermore, by concurrently leveraging the structure and semantics of messages, $\mathrm{RPLM}_{SED}$ is capable of capturing the interplay between structural and semantic information of messages, thereby facilitating a more accurate determination of event relations between messages.
Conversely, GNN-based models are impacted during the learning process by the presence of noisy edges between different event messages or the absence of effective connections among messages of the same event, thereby affecting the quality of message representation.
Additionally, the static initialization of message nodes impedes their ability to dynamically refine the embeddings of message content throughout the training process in the manner exhibited by $\mathrm{RPLM}_{SED}$.
\begin{table}[h]
    \centering
    \begin{threeparttable}
        \setlength{\tabcolsep}{3pt} 
        \caption{Incremental evaluation AMIs on Events2012. The best results are bolded, and the second-best results are underlined.}
        \label{AMI2012}
        \begin{tabularx}{\textwidth}{c|*{8}{>{\centering\arraybackslash}X}|c}
        \toprule
        Blocks  & PP-GCN & EventX & KPGNN &QSGNN & \text{FinEven$\mathrm{t}^{\mathrm{k}}$}  & \text{FinEven$\mathrm{t}^{\mathrm{d}}$} & $\mathrm{RPLM}_{SED}^\mathrm{k}$ & $\mathrm{RPLM}_{SED}^\mathrm{d}$ & Improve\\
        \midrule
        $M_1$ & $.21 \pm .00$ & $.06 \pm .00$ & $.37 \pm .00$ & $.41 \pm .02$ & $.36 \pm .01$ & $\underline{.84 \pm .01}$ & $.44 \pm .01$ & $\textbf{.91} \pm \textbf{.01}$ & $\uparrow .07$ \\ 
        $M_2$ & $.55 \pm .02$ & $.29 \pm .02$ & $.78 \pm .01$ & $.80 \pm .01$ & $.77 \pm .00$ & $.84 \pm .01$ & $\underline{.88 \pm .01}$ & $\textbf{.91} \pm \textbf{.00}$ & $\uparrow .07$ \\ 
        $M_3$ & $.52 \pm .01$ & $.18 \pm .01$ & $.74 \pm .00$ & $.76 \pm .01$ & $.82 \pm .01$ & $\underline{.89 \pm .01}$ & $.87 \pm .00$ & $\textbf{.93} \pm \textbf{.00}$ & $\uparrow .04$ \\ 
        $M_4$ & $.42 \pm .01$ & $.19 \pm .01$ & $.64 \pm .01$ & $.68 \pm .01$ & $.67 \pm .02$ & $.69 \pm .00$ & $\underline{.76 \pm .01}$ & $\textbf{.81} \pm \textbf{.01}$ & $\uparrow .12$ \\ 
        $M_5$ & $.46 \pm .01$ & $.14 \pm .00$ & $.71 \pm .01$ & $.73 \pm .00$ & $.74 \pm .00$ & $.82 \pm .00$ & $\underline{.83 \pm .01}$ & $\textbf{.84} \pm \textbf{.02}$ & $\uparrow .02$ \\ 
        $M_6$ & $.52 \pm .02$ & $.27 \pm .00$ & $.79 \pm .01$ & $.80 \pm .01$ & $.81 \pm .00$ & $.82 \pm .02$ & $\underline{.85 \pm .01}$ & $\textbf{.91} \pm \textbf{.00}$ & $\uparrow .09$ \\ 
        $M_7$ & $.34 \pm .00$ & $.13 \pm .00$ & $.51 \pm .01$ & $.54 \pm .00$ & $.53 \pm .00$ & $\underline{.72 \pm .00}$ & $.66 \pm .01$ & $\textbf{.88} \pm \textbf{.01}$ & $\uparrow .16$ \\ 
        $M_8$ & $.49 \pm .02$ & $.21 \pm .00$ & $.76 \pm .00$ & $.75 \pm .01$ & $.84 \pm .01$ & $\textbf{.87} \pm \textbf{.01}$ & $.86 \pm .01$ & $\underline{.86 \pm .00}$ & $\downarrow .01$ \\ 
        $M_9$ & $.46 \pm .02$ & $.19 \pm .00$ & $.71 \pm .02$ & $.75 \pm .02$ & $.75 \pm .00$ & $.78 \pm .01$ & $\underline{.86 \pm .01}$ & $\textbf{.91} \pm \textbf{.00}$ & $\uparrow .13$ \\ 
        $M_{10}$ & $.51 \pm .02$ & $.24 \pm .00$ & $.78 \pm .01$ & $.80 \pm .03$ & $.78 \pm .00$ & $.81 \pm .00$ & $\underline{.88 \pm .01}$ & $\textbf{.90} \pm \textbf{.01}$ & $\uparrow .09$ \\ 
        $M_{11}$ & $.46 \pm .01$ & $.24 \pm .00$ & $.71 \pm .01$ & $.72 \pm .01$ & $.73 \pm .00$ & $.74 \pm .00$ & $\underline{.85 \pm .01}$ & $\textbf{.87} \pm \textbf{.01}$ & $\uparrow .13$ \\ 
        $M_{12}$ & $.42 \pm .01$ & $.16 \pm .00$ & $.66 \pm .01$ & $.68 \pm .00$ & $.75 \pm .01$ & $.67 \pm .02$ & $\textbf{.84} \pm \textbf{.00}$ & $\underline{.77 \pm .00}$ & $\uparrow .09$ \\ 
        $M_{13}$ & $.43 \pm .01$ & $.16 \pm .00$ & $.67 \pm .01$ & $.66 \pm .01$ & $.64 \pm .00$ & $\underline{.79 \pm .00}$ & $.75 \pm .01$ & $\textbf{.91} \pm \textbf{.00}$ & $\uparrow .12$ \\ 
        $M_{14}$ & $.41 \pm .01$ & $.14 \pm .00$ & $.65 \pm .00$ & $.66 \pm .01$ & $.72 \pm .00$ & $\underline{.82 \pm .01}$ & $.81 \pm .01$ & $\textbf{.88} \pm \textbf{.01}$ & $\uparrow .06$ \\ 
        $M_{15}$ & $.35 \pm .01$ & $.07 \pm .00$ & $.54 \pm .00$ & $.55 \pm .01$ & $.61 \pm .00$ & $.67 \pm .01$ & $\underline{.73 \pm .01}$ & $\textbf{.82} \pm \textbf{.01}$ & $\uparrow .15$ \\ 
        $M_{16}$ & $.52 \pm .01$ & $.19 \pm .00$ & $.77 \pm .01$ & $.76 \pm .02$ & $.75 \pm .01$ & $\underline{.90 \pm .00}$ & $.82 \pm .00$ & $\textbf{.93} \pm \textbf{.01}$ & $\uparrow .03$ \\ 
        $M_{17}$ & $.45 \pm .00$ & $.18 \pm .00$ & $.68 \pm .01$ & $.69 \pm .01$ & $.71 \pm .02$ & $\underline{.82 \pm .00}$ & $.79 \pm .00$ & $\textbf{.86} \pm \textbf{.01}$ & $\uparrow .04$ \\ 
        $M_{18}$ & $.45 \pm .01$ & $.16 \pm .00$ & $.66 \pm .02$ & $.68 \pm .01$ & $.70 \pm .00$ & $.74 \pm .00$ & $\underline{.81 \pm .01}$ & $\textbf{.82} \pm \textbf{.01}$ & $\uparrow .08$ \\ 
        $M_{19}$ & $.48 \pm .02$ & $.16 \pm .00$ & $.71 \pm .01$ & $.70 \pm .01$ & $.75 \pm .01$ & $.66 \pm .00$ & $\underline{.86 \pm .01}$ & $\textbf{.90} \pm \textbf{.01}$ & $\uparrow .15$ \\ 
        $M_{20}$ & $.45 \pm .02$ & $.18 \pm .00$ & $.68 \pm .02$ & $.69 \pm .02$ & $.68 \pm .00$ & $.78 \pm .00$ & $\textbf{.82} \pm \textbf{.00}$ & $\underline{.80 \pm .01}$ & $\uparrow .04$ \\ 
        $M_{21}$ & $.38 \pm .02$ & $.10 \pm .00$ & $.57 \pm .00$ & $.58 \pm .00$ & $.63 \pm .01$ & $.64 \pm .01$ & $\textbf{.74} \pm \textbf{.01}$ & $\underline{.71 \pm .01}$ & $\uparrow .10$ \\ 
        \bottomrule
        \end{tabularx}
    \end{threeparttable}
\end{table}
\begin{table}[b]
    \centering
    \begin{threeparttable}
        \setlength{\tabcolsep}{3pt} 
        \caption{Incremental evaluation ARIs on Events2012. The best results are bolded, and the second-best results are underlined.}
        \label{ARI2012}
        \begin{tabularx}{\textwidth}{c|*{8}{>{\centering\arraybackslash}X}|c}
        \toprule
        Blocks  & PP-GCN & EventX & KPGNN &QSGNN & \text{FinEven$\mathrm{t}^{\mathrm{k}}$} & \text{FinEven$\mathrm{t}^{\mathrm{d}}$} & $\mathrm{RPLM}_{SED}^\mathrm{k}$ & $\mathrm{RPLM}_{SED}^\mathrm{d}$ & Improve\\
        \midrule
        $M_1$ & $.05 \pm .00$ & $.01 \pm .00$ & $.07 \pm .01$ & $.09 \pm .01$ & $.05 \pm .00$ & $\underline{.90 \pm .00}$ & $.07 \pm .01$ & $\textbf{.96} \pm \textbf{.01}$ & $\uparrow .06$ \\ 
        $M_2$ & $.67 \pm .03$ & $.45 \pm .02$ & $.76 \pm .02$ & $.78 \pm .02$ & $.67 \pm .01$ & \underline{$.90 \pm .01$} & $.83 \pm .02$ & $\textbf{.91} \pm \textbf{.01}$ & $\uparrow .00$ \\ 
        $M_3$ & $.47 \pm .01$ & $.09 \pm .01$ & $.58 \pm .01$ & $.62 \pm .01$ & $.58 \pm .00$ & $\underline{.89 \pm .01}$ & $.66 \pm .01$ & $\textbf{.94} \pm \textbf{.00}$ & $\uparrow .05$ \\ 
        $M_4$ & $.24 \pm .01$ & $.07 \pm .01$ & $.29 \pm .01$ & $.31 \pm .02$ & $.27 \pm .02$ & $.27 \pm .01$ & $\underline{.39 \pm .02}$ & $\textbf{.61} \pm \textbf{.00}$ & $\uparrow .32$ \\ 
        $M_5$ & $.34 \pm .00$ & $.04 \pm .00$ & $.47 \pm .03$ & $.48 \pm .02$ & $.43 \pm .01$ & $.63 \pm .02$ & $\underline{.67 \pm .01}$ & $\textbf{.74} \pm \textbf{.02}$ & $\uparrow .11$ \\ 
        $M_6$ & $.55 \pm .03$ & $.14 \pm .00$ & $.72 \pm .03$ & $.65 \pm .03$ & $.65 \pm .00$ & $\underline{.74 \pm .00}$ & $.65 \pm .01$ & $\textbf{.93} \pm \textbf{.00}$ & $\uparrow .19$ \\ 
        $M_7$ & $.11 \pm .02$ & $.02 \pm .00$ & $.12 \pm .00$ & $.13 \pm .02$ & $.09 \pm .01$ & $\underline{.45 \pm .01}$ & $.19 \pm .01$ & $\textbf{.91} \pm \textbf{.01}$ & $\uparrow .46$ \\ 
        $M_8$ & $.43 \pm .04$ & $.09 \pm .00$ & $.60 \pm .01$ & $.57 \pm .02$ & $.65 \pm .02$ & $\underline{.72 \pm .01}$ & $.68 \pm .02$ & $\textbf{.79} \pm \textbf{.01}$ & $\uparrow .07$ \\ 
        $M_9$ & $.31 \pm .02$ & $.07 \pm .00$ & $.46 \pm .02$ & $.48 \pm .03$ & $.43 \pm .00$ & $\underline{.68 \pm .00}$ & $.56 \pm .01$ & $\textbf{.95} \pm \textbf{.01}$ & $\uparrow .27$ \\ 
        $M_{10}$ & $.50 \pm .07$ & $.13 \pm .00$ & $.70 \pm .06$ & $.65 \pm .02$ & $.62 \pm .02$ & $.74 \pm .00$ & $\underline{.84 \pm .01}$ & $\textbf{.93} \pm \textbf{.01}$ & $\uparrow .19$ \\ 
        $M_{11}$ & $.38 \pm .02$ & $.16 \pm .00$ & $.49 \pm .03$ & $.50 \pm .03$ & $.42 \pm .01$ & $.60 \pm .00$ & $\underline{.61 \pm .06}$ & $\textbf{.91} \pm \textbf{.01}$ & $\uparrow .31$ \\ 
        $M_{12}$ & $.34 \pm .03$ & $.07 \pm .00$ & $.48 \pm .01$ & $.47 \pm .01$ & $.44 \pm .00$ & $.26 \pm .00$ & $\textbf{.69} \pm \textbf{.03}$ & $\underline{.48 \pm .03}$ & $\uparrow .21$ \\ 
        $M_{13}$ & $.19 \pm .01$ & $.04 \pm .00$ & $.29 \pm .03$ & $.28 \pm .02$ & $.21 \pm .02$ & $\underline{.75 \pm .02}$ & $.32 \pm .03$ & $\textbf{.97} \pm \textbf{.00}$ & $\uparrow .22$ \\ 
        $M_{14}$ & $.29 \pm .01$ & $.10 \pm .00$ & $.42 \pm .02$ & $39 \pm .01$ & $.43 \pm .01$ & $\underline{.81 \pm .01}$ & $.53 \pm .02$ & $\textbf{.85} \pm \textbf{.01}$ & $\uparrow .04$ \\ 
        $M_{15}$ & $.15 \pm .00$ & $.01 \pm .00$ & $.17 \pm .00$ & $.15 \pm .01$ & $.16 \pm .00$ & $\underline{.46 \pm .00}$ & $.31 \pm .02$ & $\textbf{.78} \pm \textbf{.00}$ & $\uparrow .32$ \\ 
        $M_{16}$ & $.51 \pm .03$ & $.08 \pm .00$ & $.66 \pm .05$ & $.58 \pm .04$ & $.56 \pm .01$ & $\underline{.88 \pm .01}$ & $.61 \pm .01$ & $\textbf{.95} \pm \textbf{.01}$ & $\uparrow .07$ \\ 
        $M_{17}$ & $.35 \pm .03$ & $.12 \pm .00$ & $.43 \pm .05$ & $.40\pm .02$ & $.36 \pm .01$ & $\underline{.81 \pm .01}$ & $.44 \pm .01$ & $\textbf{.89} \pm \textbf{.00}$ & $\uparrow .08$ \\ 
        $M_{18}$ & $.39 \pm .03$ & $.08 \pm .00$ & $.47 \pm .04$ & $.48 \pm .01$ & $.44 \pm .01$ & $.52 \pm .01$ & $\underline{.56 \pm .01}$ & $\textbf{.67} \pm \textbf{.02}$ & $\uparrow .15$ \\ 
        $M_{19}$ & $.41 \pm .02$ & $.07 \pm .00$ & $.51 \pm .03$ & $.45 \pm .03$ & $.44 \pm .00$ & $.35 \pm .01$ & $\underline{.62 \pm .03}$ & $\textbf{.90} \pm \textbf{.01}$ & $\uparrow .39$ \\ 
        $M_{20}$ & $.41 \pm .01$ & $.11 \pm .00$ & $.51 \pm .04$ & $.47\pm .02$ & $.43 \pm .02$ & $\underline{.71 \pm .01}$ & $.64 \pm .02$ & $\textbf{.75} \pm \textbf{.02}$ & $\uparrow .04$ \\ 
        $M_{21}$ & $.20 \pm .03$ & $.01 \pm .00$ & $.20 \pm .01$ & $.24 \pm .01$ & $.23 \pm .00$ & $\textbf{.48} \pm \textbf{.00}$ & $.37 \pm .02$ & $\underline{.44 \pm .01}$ & $\downarrow .04$ \\ 
        \bottomrule
        \end{tabularx}
    \end{threeparttable}
\end{table}
\begin{table}[t]
    \centering
    \begin{threeparttable}
        \setlength{\tabcolsep}{3pt} 
        \caption{Incremental evaluation NMIs on Events2018. The best results are bolded, and the second-best results are underlined.}
        \label{NMI2018}
        \begin{tabularx}{\textwidth}{c|*{8}{>{\centering\arraybackslash}X}|c}
        \toprule
        Blocks & PP-GCN & EventX & KPGNN & QSGNN & \text{FinEven$\mathrm{t}^{\mathrm{k}}$}  & \text{FinEven$\mathrm{t}^{\mathrm{d}}$} & $\mathrm{RPLM}_{SED}^\mathrm{k}$ & $\mathrm{RPLM}_{SED}^\mathrm{d}$ & Improve\\
        \midrule
        $M_1$ & $.49 \pm .01$ & $.34 \pm .00$ & $.54 \pm .01$ & $.57 \pm .01$ & $.57 \pm .01$ & $.70 \pm.01$ & $\underline{.81 \pm .01}$ & $\textbf{.89} \pm \textbf{.01}$ & $\uparrow .19$ \\ 
        $M_2$ & $.45 \pm .00$ & $.37 \pm .00$ & $.56 \pm .02$ & $.58 \pm .01$ & $.60 \pm .01$ & $.74 \pm .01$ & $\underline{.77 \pm .01}$ & $\textbf{.84} \pm \textbf{.01}$ & $\uparrow .10$ \\ 
        $M_3$ & $.56 \pm .03$ & $.37 \pm .00$ & $.52 \pm .03$ & $.57 \pm .01$ & $.62 \pm .02$ & $.64 \pm .00$ & $\underline{.73 \pm .01}$ & $\textbf{.76} \pm \textbf{.01}$ & $\uparrow .12$ \\ 
        $M_4$ & $.54 \pm .03$ & $.39 \pm .00$ & $.55 \pm .01$ & $.58 \pm .03$ & $.58 \pm .01$ & $.72 \pm .01$ & $\textbf{.77} \pm \textbf{.01}$ & $\underline{.75 \pm .00}$ & $\uparrow .05$ \\ 
        $M_5$ & $.54 \pm .02$ & $.53 \pm .00$ & $.58 \pm .02$ & $.61 \pm .02$ & $.59 \pm .02$ & $.64 \pm .00$ & $\textbf{.72} \pm \textbf{.01}$ & $\underline{.67 \pm .01}$ & $\uparrow .08$ \\ 
        $M_6$ & $.52 \pm .02$ & $.44 \pm .00$ & $.59 \pm .03$ & $.60 \pm .01$ & $.57 \pm .01$ & $.67 \pm .00$ & $\textbf{.75} \pm \textbf{.01}$ & $\underline{.73 \pm .02}$ & $\uparrow .08$ \\ 
        $M_7$ & $.56 \pm .04$ & $.42 \pm .00$ & $.63 \pm .02$ & $.64 \pm .01$ & $.67 \pm .01$ & $.78 \pm .01$ & $.\underline{79 \pm .00}$ & $\textbf{.88} \pm \textbf{.00}$ & $\uparrow .10$ \\ 
        $M_8$ & $.56 \pm .03$ & $.54 \pm .00$ & $.58 \pm .02$ & $.57 \pm .02$ & $.62 \pm .02$ & $.66 \pm .02$ & $\textbf{.80} \pm \textbf{.01}$ & $\underline{.77 \pm .01}$ & $\uparrow .14$ \\ 
        $M_9$ & $.54 \pm .02$ & $.45 \pm .00$ & $.48 \pm .02$ & $.52 \pm .02$ & $.48 \pm .01$ & $.57 \pm .01$ & $\textbf{.61} \pm \textbf{.00}$ & $\underline{.58 \pm .02}$ & $\uparrow .04$ \\ 
        $M_{10}$ & $.56 \pm .06$ & $.52 \pm .00$ & $.57 \pm .01$ & $.60 \pm .01$ & $.60 \pm .01$ & $.65 \pm .01$ & $\textbf{.77} \pm \textbf{.01}$ & $\textbf{.77} \pm \textbf{.01}$ & $\uparrow .12$ \\ 
        $M_{11}$ & $.59 \pm .03$ & $.48 \pm .00$ & $.54 \pm .01$ & $.60 \pm .01$ & $.54 \pm .01$ & $.63 \pm .00$ & $\underline{.66 \pm .00}$ & $\textbf{.68} \pm \textbf{.01}$ & $\uparrow .05$ \\ 
        $M_{12}$ & $.60 \pm .02$ & $.51 \pm .00$ & $.55 \pm .04$ & $.61 \pm .02$ & $.59 \pm .02$ & $.70 \pm .01$ & $.\textbf{80 }\pm \textbf{.01}$ & $\underline{.77 \pm .01}$ & $\uparrow .10$ \\ 
        $M_{13}$ & $.61 \pm .01$ & $.44 \pm .00$ & $.60 \pm .02$ & $.59 \pm .04$ & $.64 \pm .01$ & $.67 \pm .00$ & $\textbf{.70} \pm \textbf{.00}$ & $\underline{.68 \pm .00}$ & $\uparrow .03$ \\ 
        $M_{14}$ & $.60 \pm .02$ & $.52 \pm .00$ & $.66 \pm .01$ & $.68 \pm .02$ & $.65 \pm .00$ & $.65 \pm .00$ & $\underline{.69 \pm .00}$ & $\textbf{.70} \pm \textbf{.00}$ & $\uparrow .05$ \\ 
        $M_{15}$ & $.57 \pm .03$ & $.49 \pm .00$ & $.60 \pm .01$ & $.63 \pm .02$ & $.60 \pm .01$ & $.70 \pm .01$ & $\underline{.71 \pm .01}$ & $\textbf{.73} \pm \textbf{.01}$ & $\uparrow .03$ \\ 
        $M_{16}$ & $.53 \pm .02$ & $.39 \pm .00$ & $.52 \pm .02$ & $.51 \pm .03$ & $.61 \pm .01$ & $\underline{.75 \pm .01}$ & $.74 \pm .02$ & $\textbf{.95} \pm \textbf{.02}$ & $\uparrow .20$ \\ 
        \bottomrule
        \end{tabularx}
    \end{threeparttable}
\end{table}
\begin{table}[t]
    \centering
    
    \begin{threeparttable}
        \setlength{\tabcolsep}{3pt} 
        \caption{Incremental evaluation AMIs on Events2018. The best results are bolded, and the second-best results are underlined.}
        \label{AMI2018}
        \begin{tabularx}{\textwidth}{c|*{8}{>{\centering\arraybackslash}X}|c}
        \toprule
        Blocks & PP-GCN & EventX & KPGNN & QSGNN & \text{FinEven$\mathrm{t}^{\mathrm{k}}$}  & \text{FinEven$\mathrm{t}^{\mathrm{d}}$} & $\mathrm{RPLM}_{SED}^\mathrm{k}$ & $\mathrm{RPLM}_{SED}^\mathrm{d}$ & Improve\\
        \midrule
        $M_1$ & $.48 \pm .00$ & $.11 \pm .00$ & $.54 \pm .01$ & $.56 \pm .01$ & $.55 \pm .01$ & $.70 \pm .00$ & $\underline{.81 \pm .01}$ & $\textbf{.89} \pm \textbf{.01}$ & $\uparrow .19$ \\ 
        $M_2$ & $.44 \pm .02$ & $.12 \pm .00$ & $.55 \pm .01$ & $.57 \pm .01$ & $.60 \pm .01$ & $.74 \pm .00$ & $\underline{.77 \pm .01}$ & $\textbf{.84} \pm \textbf{.00}$ & $\uparrow .10$ \\ 
        $M_3$ & $.55 \pm .03$ & $.11 \pm .00$ & $.55 \pm .02$ & $.56 \pm .02$ & $.62 \pm .01$ & $.64 \pm .00$ & $\underline{.73 \pm .01}$ & $\textbf{.77} \pm \textbf{.01}$ & $\uparrow .13$ \\ 
        $M_4$ & $.54 \pm .04$ & $.14 \pm .00$ & $.55 \pm .01$ & $.57 \pm .03$ & $.60 \pm .01$ & $.67 \pm .00$ & $\textbf{.76} \pm \textbf{.01}$ & $\underline{.75 \pm .02}$ & $\uparrow .09$ \\ 
        $M_5$ & $.53 \pm .02$ & $.24 \pm .00$ & $.57 \pm .01$ & $.59 \pm .01$ & $.57 \pm .01$ & $.64 \pm .00$ & $\textbf{.70} \pm \textbf{.01}$ & $\underline{.67 \pm .01}$ & $\uparrow .06$ \\ 
        $M_6$ & $.50 \pm .03$ & $.15 \pm .00$ & $.57 \pm .02$ & $.59 \pm .01$ & $.56 \pm .01$ & $.57 \pm .00$ & $\textbf{.75} \pm \textbf{.01}$ & $\underline{.72 \pm .01}$ & $\uparrow .18$ \\ 
        $M_7$ & $.55 \pm .04$ & $.12 \pm .00$ & $.61 \pm .02$ & $.63 \pm .01$ & $.62 \pm .02$ & $.67 \pm .00$ & $\underline{.79 \pm .00}$ & $\textbf{.87} \pm \textbf{.01}$ & $\uparrow .20$ \\ 
        $M_8$ & $.55 \pm .02$ & $.21 \pm .00$ & $.57 \pm .02$ & $.55 \pm .02$ & $.61 \pm .01$ & $.62 \pm .00$ & $\textbf{.78} \pm \textbf{.02}$ & $\underline{.76 \pm .02}$ & $\uparrow .16$ \\ 
        $M_9$ & $.48 \pm .03$ & $.16 \pm .00$ & $.46 \pm .02$ & $.46 \pm .02$ & $.47 \pm .01$ & $.52 \pm .00$ & $\textbf{.60} \pm \textbf{.01}$ & $\underline{.57 \pm .02}$ & $\uparrow .08$ \\ 
        $M_{10}$ & $.55 \pm .04$ & $.19 \pm .00$ & $.56 \pm .02$ & $.58 \pm .01$ & $.59 \pm .01$ & $.60 \pm .00$ & $\underline{.75 \pm .02}$ & $\textbf{.75} \pm \textbf{.01}$ & $\uparrow .15$ \\ 
        $M_{11}$ & $.57 \pm .02$ & $.18 \pm .00$ & $.53 \pm .01$ & $.59 \pm .02$ & $.52 \pm .01$ & $.54 \pm .00$ & $\underline{.64 \pm .00}$ & $\textbf{.68} \pm \textbf{.01}$ & $\uparrow .14$ \\ 
        $M_{12}$ & $.58 \pm .02$ & $.20 \pm .00$ & $.56 \pm .02$ & $.59 \pm .02$ & $.62 \pm .02$ & $.59 \pm .00$ & $\textbf{.79} \pm \textbf{.01}$ & $\underline{.77 \pm .01}$ & $\uparrow .17$ \\ 
        $M_{13}$ & $.59 \pm .02$ & $.15 \pm .00$ & $.60 \pm .02$ & $.58 \pm .03$ & $.65 \pm .01$ & $.64 \pm .00$ & $\textbf{.69} \pm \textbf{.00}$ & $\underline{.67 \pm .00}$ & $\uparrow .04$ \\ 
        $M_{14}$ & $.59 \pm .01$ & $.22 \pm .00$ & $.65 \pm .00$ & $.67 \pm .02$ & $.67 \pm .01$ & $.65 \pm .00$ & $\underline{.68 \pm .00}$ & $\textbf{.70} \pm \textbf{.01}$ & $\uparrow .03$ \\ 
        $M_{15}$ & $.55 \pm .03$ & $.22 \pm .00$ & $.58 \pm .02$ & $.61 \pm .00$ & $.57 \pm .01$ & $.65 \pm .00$ & $\underline{.70 \pm .01}$ & $\textbf{.71} \pm \textbf{.01}$ & $\uparrow .06$ \\ 
        $M_{16}$ & $.52 \pm .02$ & $.10 \pm .00$ & $.50 \pm .01$ & $.50 \pm .03$ & $.59 \pm .01$ & $.68 \pm .00$ & $\underline{.73 \pm .02}$ & $\textbf{.94} \pm \textbf{.02}$ & $\uparrow .27$ \\ 
        \bottomrule
        \end{tabularx}
    \end{threeparttable}
\end{table}

\subsubsection{Online Evaluation}
In the online detection scenario, we utilize the data from the first week as the $M_0$ to train the initial model, constructing subsequent message blocks daily. 
It is crucial to note that for offline baselines like PP-GCN, direct adaptation to the online scenario is not feasible.
We circumvent this by retraining the model using previous message blocks as the training set and predicting the current message block. 
Unlike offline detection, the $\mathrm{RPLM}_{SED}$ method enters the maintenance phase after predicting the message block within the window. 
During the maintenance phase, $\mathrm{RPLM}_{SED}$ initially updates the parameters of the PLMs section through a parameter weighting method, thereby enabling rapid adaptation to new knowledge while preserving historical knowledge. 
Subsequently, the model is maintained using message blocks within the current window. 

As shown in Tables~\ref{NMI2012} to~\ref{ARI2018}, $\mathrm{RPLM}_{SED}$ significantly outperforms all baseline methods in incremental event detection, consistently leading in evaluation metrics across all message blocks. 
Overall, in comparison with EventX, on the Events2012 dataset, $\mathrm{RPLM}_{SED}^\mathrm{k}$ achieves improvements of 35\%, 360\%, and 480\% in terms of NMI, AMI, and ARI, respectively. 
On the Events2018 dataset, $\mathrm{RPLM}_{SED}^\mathrm{k}$ shows enhancements of 64\%, 346\%, and 4105\% in NMI, AMI, and ARI, respectively. 
This can be attributed to the fact that EventX relies on community detection whilst neglecting the rich semantic information present in social messages.
Compared to PP-GCN, KPGNN, QSGNN, and FinEven$\mathrm{t}^{\mathrm{k}}$, $\mathrm{RPLM}_{SED}^\mathrm{k}$ exhibits improvements ranging from 12\% to 71\%, 12\% to 81\%, and 21\% to 56\% in NMI, AMI, and ARI, respectively, on the Events2012 dataset.
On the Events2018 dataset, $\mathrm{RPLM}_{SED}^\mathrm{k}$ demonstrates enhancements of 22\% to 34\%, 22\% to 36\%, and 55\% to 78\% in NMI, AMI, and ARI, respectively.
Compared to FinEven$\mathrm{t}^{\mathrm{d}}$, $\mathrm{RPLM}_{SED}^\mathrm{d}$ achieves improvements of $11\%$, $11\%$, and $28\%$ in terms of NMI, AMI, and ARI, respectively, on the Events2012 dataset.
On the Events2018 dataset, $\mathrm{RPLM}_{SED}^\mathrm{d}$ shows enhancements of $11\%$, $12\%$, and $32\%$ in NMI, AMI, and ARI, respectively. 
Regardless of the clustering method employed, our model maintained optimal performance.
These improvements further underscore the effectiveness of $\mathrm{RPLM}_{SED}$.
Within the model, the multi-relational prompt-based pairwise message learning mechanism acquires more effective message representations by learning the connections and differences between messages in pairs. 
Subsequently, through the similarity-based representation aggregation method, message representations are filtered and aggregated, further enhancing the robustness of message representations.
Additionally, converting the structural relations between messages into multi-relational prompts allows the model to better cope with scenarios characterized by a scarcity of structural information.
Furthermore, at each maintenance stage, the parameters of the PLM within $\mathrm{RPLM}_{SED}$ are updated through a weighted method. 
This endows $\mathrm{RPLM}_{SED}$with enhanced capabilities for knowledge adaptation, preservation, and expansion, thereby exhibiting stronger generalization abilities and stability when dealing with social data streams.

\begin{figure}[h]
    \centering
    \subfigure[KPGNN ($M_9$).]{
        \includegraphics[width=0.24\linewidth, trim={0 0.2cm 0 0}, clip]{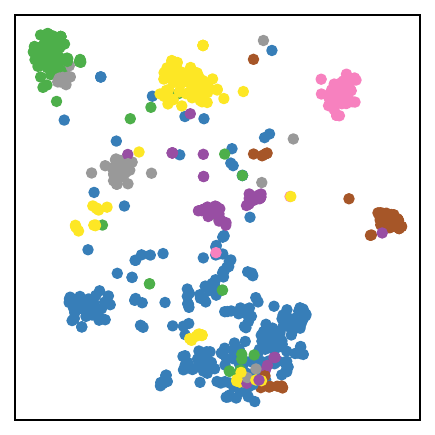}
    }
    \hspace{-10pt}
    \subfigure[QSGNN ($M_9$).]{
        \includegraphics[width=0.24\linewidth, trim={0 0.2cm 0 0}, clip]{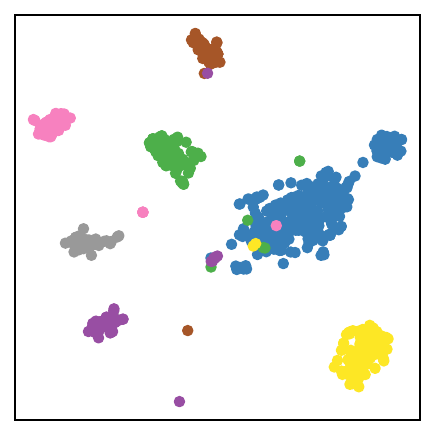}
    }
    \hspace{-10pt}
    \subfigure[FinEvnet ($M_9$).]{
        \includegraphics[width=0.24\linewidth, trim={0 0.2cm 0 0}, clip]{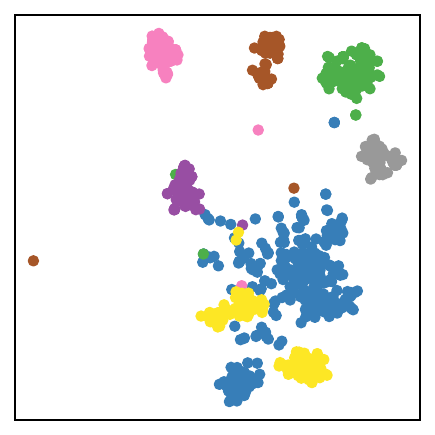}
    }
    \hspace{-10pt}
    \subfigure[$\mathrm{RPLM}_{SED}$ ($M_9$).]{
        \includegraphics[width=0.24\linewidth, trim={0 0.2cm 0 0}, clip]{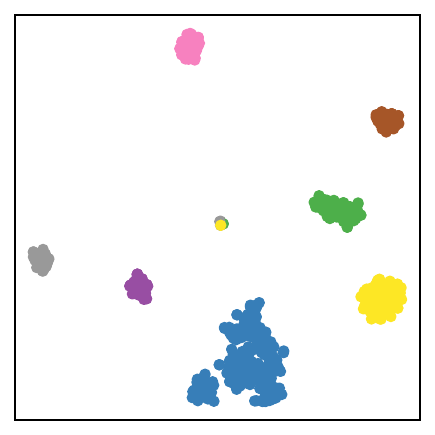}
    }
    \vspace{-10pt}
    
     \subfigure[KPGNN ($M_{12}$).]{
        \includegraphics[width=0.24\linewidth, trim={0 0.2cm 0 0}, clip]{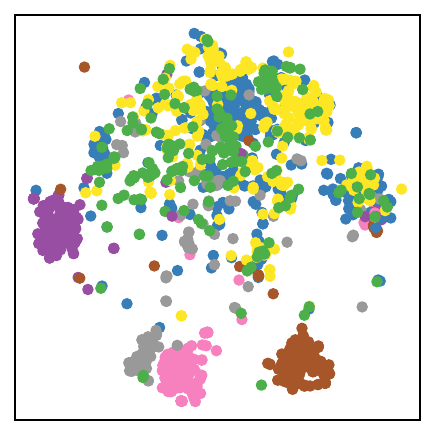}
    }
    \hspace{-10pt}
    \subfigure[QSGNN ($M_{12}$).]{
        \includegraphics[width=0.24\linewidth, trim={0 0.2cm 0 0}, clip]{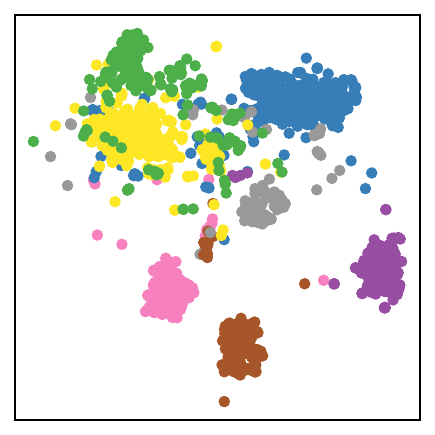}
    }
    \hspace{-10pt}
    \subfigure[FinEvent ($M_{12}$).]{
        \includegraphics[width=0.24\linewidth, trim={0 0.2cm 0 0}, clip]{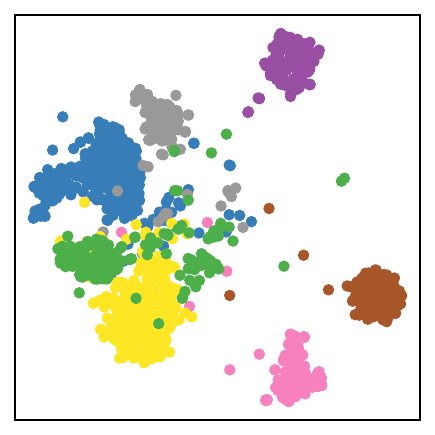}
    }
    \hspace{-10pt}
    \subfigure[$\mathrm{RPLM}_{SED}$ ($M_{12}$).]{
        \includegraphics[width=0.24\linewidth, trim={0 0.2cm 0 0}, clip]{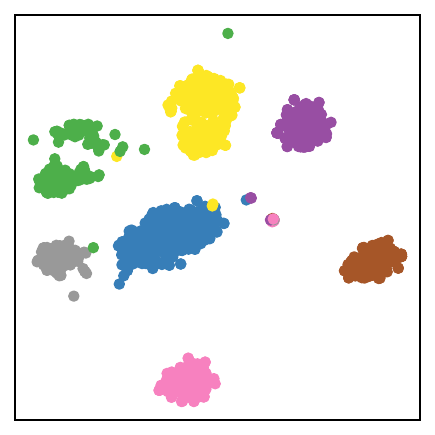}
    }
    \vspace{-10pt}
    
     \subfigure[KPGNN ($M_{19}$).]{
        \includegraphics[width=0.24\linewidth, trim={0 0.2cm 0 0}, clip]{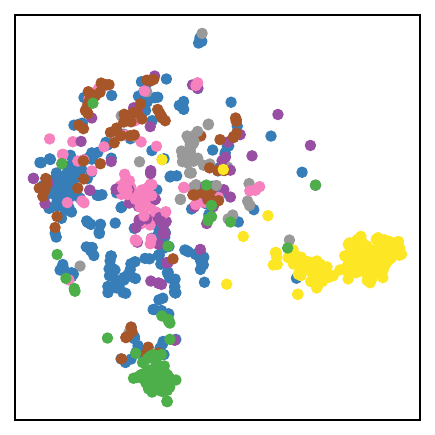}
    }
    \hspace{-10pt}
    \subfigure[QSGNN ($M_{19}$).]{
        \includegraphics[width=0.24\linewidth, trim={0 0.2cm 0 0}, clip]{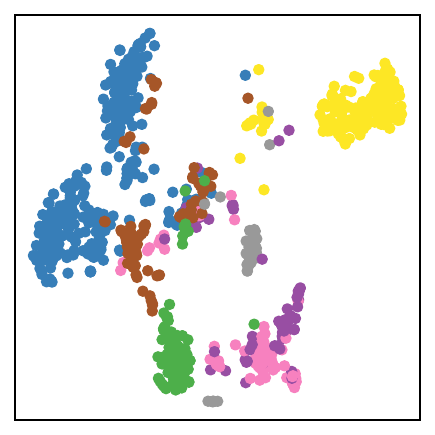}
    }
    \hspace{-10pt}
    \subfigure[FinEvent ($M_{19}$).]{
        \includegraphics[width=0.24\linewidth, trim={0 0.2cm 0 0}, clip]{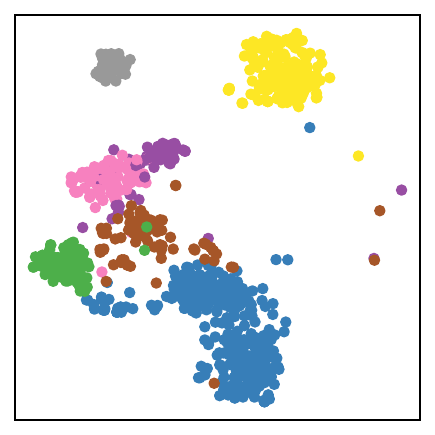}
    }
    \hspace{-10pt}
    \subfigure[$\mathrm{RPLM}_{SED}$ ($M_{19}$).]{
        \includegraphics[width=0.24\linewidth, trim={0 0.2cm 0 0}, clip]{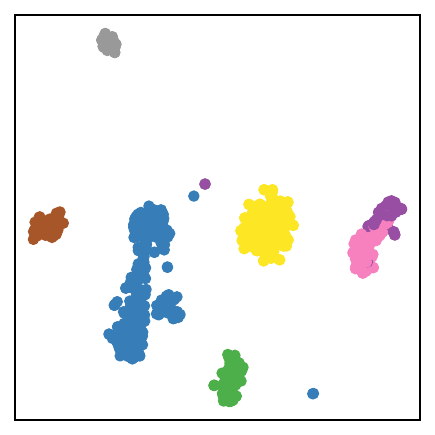}
    }
    \vspace{15pt}
    \caption{\revised{Message representation visualization in the detection stage. Each row represents the t-SNE dimensionality reduction visualization of message representations for the same block obtained by different methods. Messages from the same event are represented with the same color.}}
    \label{fig:visualization}
\end{figure}

As depicted in Figure~\ref{fig:visualization}, to intuitively compare and further demonstrate the effectiveness of our proposed $\mathrm{RPLM}_{SED}$ model, we select message representations of three message blocks ($M_2$, $M_9$, and $M_{19}$) from the Events2012 dataset. 
These are subsequently subjected to dimensionality reduction and visualization using t-SNE. 
Considering the long-tail distribution characteristic of social data, our attention is focused on the top seven events \revised{of each message block} by quantity. 
From the visualization results obtained from the $\mathrm{RPLM}_{SED}$ (Figure~\ref{fig:visualization} (d), (h), and (l)), it is clearly observable that event clusters exhibit stronger cohesion. 
Simultaneously, the separation between distinct event clusters is also more pronounced.
This reflects the importance of the proposed clustering constraint, which effectively enhances the distinguishability of message representations, thereby improving the accuracy of event detection.
\begin{table}[h]
    \centering
    \begin{threeparttable}
        \setlength{\tabcolsep}{3pt} 
        \caption{Incremental evaluation ARIs on Events2018. The best results are bolded, and the second-best results are underlined.}
        \label{ARI2018}
        \begin{tabularx}{\textwidth}{c|*{8}{>{\centering\arraybackslash}X}|c}
        \toprule
        Blocks & PP-GCN & EventX & KPGNN&QSGNN & \text{FinEven$\mathrm{t}^{\mathrm{k}}$} & \text{FinEven$\mathrm{t}^{\mathrm{d}}$} & $\mathrm{RPLM}_{SED}^\mathrm{k}$ & $\mathrm{RPLM}_{SED}^\mathrm{d}$ & Improve\\
        \midrule
        $M_1$ & $.27 \pm .03$ & $.01 \pm .00$ & $.29 \pm .02$ & $.30 \pm .01$ & $.32 \pm .02$ & $.60 \pm .01$ & $\underline{.85 \pm .01}$ & $\textbf{.94} \pm \textbf{.01}$ & $\uparrow .34$ \\ 
        $M_2$ & $.21 \pm .01$ & $.01 \pm .00$ & $.37 \pm .01$ & $.38 \pm .02$ & $.42 \pm .02$ & $.58 \pm .01$ & $\underline{.79 \pm .02}$ & $\textbf{.89} \pm \textbf{.01}$ & $\uparrow .31$ \\ 
        $M_3$ & $.38 \pm .03$ & $.01 \pm .00$ & $.39 \pm .04$ & $.36 \pm .02$ & $.47 \pm .01$ & $.50 \pm .02$ & $\underline{.76 \pm .01}$ & $\textbf{.81} \pm \textbf{.00}$ & $\uparrow .31$ \\ 
        $M_4$ & $.35 \pm .05$ & $.01 \pm .00$ & $.36 \pm .04$ & $.36 \pm .01$ & $.40 \pm .03$ & $.51 \pm .02$ & $\textbf{.64} \pm \textbf{.01}$ & $\underline{.64 \pm  .02}$ & $\uparrow .13$ \\ 
        $M_5$ & $.30 \pm .01$ & $.03 \pm .00$ & $.37 \pm .02$ & $.36 \pm .02$ & $.41 \pm .01$ & $.48 \pm .01$ & $\underline{.49 \pm .03}$ & $\textbf{.49} \pm \textbf{.01}$ & $\uparrow .01$ \\ 
        $M_6$ & $.27 \pm .02$ & $.01 \pm .00$ & $.35 \pm .04$ & $.36 \pm .01$ & $.35 \pm .02$ & $.46 \pm .02$ & $\textbf{.63} \pm \textbf{.02}$ & $\underline{.60 \pm .02}$ & $\uparrow .17$ \\ 
        $M_7$ & $.38 \pm .05$ & $.01 \pm .00$ & $.37 \pm .02$ & $.38 \pm .02$ & $.43 \pm .01$ & $.61 \pm .01$ & $.\underline{79 \pm .00}$ & $\textbf{.92} \pm \textbf{.01}$ & $\uparrow .31$ \\ 
        $M_8$ & $.38 \pm .04$ & $.01 \pm .00$ & $.38 \pm .02$ & $.35 \pm .02$ & $.45 \pm .02$ & $.47 \pm .01$ & $\textbf{.75} \pm \textbf{.02}$ & $\underline{.67 \pm .01}$ & $\uparrow .28$ \\ 
        $M_9$ & $.32 \pm .04$ & $.01 \pm .00$ & $.23 \pm .02$ & $.30 \pm .04$ & $.26 \pm .01$ & $.39 \pm .01$ & $\textbf{.44} \pm \textbf{.01}$ & $\underline{.41 \pm .01}$ & $\uparrow .04$ \\ 
        $M_{10}$ & $.37 \pm .04$ & $.02 \pm .00$ & $.38 \pm .02$ & $.37 \pm .02$ & $.44 \pm .03$ & $\underline{.52 \pm .01}$ & $.51 \pm .02$ & $\textbf{.55} \pm \textbf{.00}$ & $\uparrow .03$ \\ 
        $M_{11}$ & $.37 \pm .04$ & $.01 \pm .00$ & $.25 \pm .02$ & $.24 \pm .03$ & $.27 \pm .01$ & $\underline{.40 \pm .01}$ & $.29 \pm .01$ & $\textbf{.63} \pm \textbf{.02}$ & $\uparrow .23$ \\ 
        $M_{12}$ & $.39 \pm .03$ & $.02 \pm .00$ & $.46 \pm .02$ & $.40 \pm .04$ & $.45 \pm .01$ & $\underline{.57 \pm .02}$ & $\textbf{.77} \pm \textbf{.01}$ & $.48 \pm .01$ & $\uparrow .20$ \\ 
        $M_{13}$ & $.39 \pm .01$ & $.01 \pm .00$ & $.36 \pm .05$ & $.33 \pm .03$ & $.42 \pm .03$ & $.49 \pm .04$ & $\underline{.61 \pm .01}$ & $\textbf{.64} \pm \textbf{.01}$ & $\uparrow .15$ \\ 
        $M_{14}$ & $.39 \pm .06$ & $.03 \pm .00$ & $.50 \pm .03$ & $.47 \pm .03$ & $.50 \pm .03$ & $\textbf{.52} \pm \textbf{.02}$ & $.46 \pm .02$ & $\underline{.50 \pm .01}$ & $\downarrow .02$ \\ 
        $M_{15}$ & $.40 \pm .06$ & $.02 \pm .00$ & $.37 \pm .02$ & $.32 \pm .02$ & $.38 \pm .02$ & $.49 \pm .01$ & $\underline{.51 \pm .01}$ & $\textbf{.63} \pm \textbf{.01}$ & $\uparrow .14$ \\ 
        $M_{16}$ & $.26 \pm .03$ & $.01 \pm .00$ & $.26 \pm .02$ & $.25 \pm .02$ & $.30 \pm .03$ & $\underline{.46 \pm .02}$ & $.38 \pm .03$ & $\textbf{.97} \pm \textbf{.01}$ & $\uparrow .51$ \\
        \bottomrule
        \end{tabularx}
    \end{threeparttable}
\end{table}

\subsection{Performance Evaluation on Low-Resource Language}
\label{sec:Low-Resource Language}
Low-resource social event detection often faces the challenge of scarce training resources.
Recent research introduced the CLKD~\cite{ren2021transferring} method, utilizing a knowledge distillation-based framework~\cite{gupta2016cross} to address this challenge. 
Specifically, CLKD initially employs pre-trained word embedding algorithms~\cite{mikolov2013efficient,pennington2014glove,bojanowski2017enriching} to generate initial embeddings of messages, followed by mapping the data of low-resource languages into the word embedding space of high-resource languages through linear~\cite{LampleCRDJ18,artetxe-etal-2018-robust} or non-linear~\cite{mohiuddin-etal-2020-lnmap} mapping techniques. 
Subsequently, a teacher model trained in a high-resource language (e.g., English) guides the student model in low-resource languages (e.g., Arabic) for learning and training. 
However, similar to other GNN-based methods, the backbone encoder of CLKD is still GNN. 
GNNs learn representations from explicit message graphs, within which some noisy edges are inevitably present. 
This results in the knowledge learned by the teacher model not being entirely effective and accurate, which may adversely affect the learning process of the student model.
To validate our hypothesis, we conduct experiments on the low-resource Arabic-Twitter dataset.
We choose TwHIN-BERT (TwHIN-BERT-large) as the backbone architecture for $\mathrm{RPLM}_{SED}$ and implement CLKD  by using open-source code\footnote{\url{https://github.com/RingBDStack/CLKD}}. 
\begin{table}[h]
    \centering
    \begin{threeparttable}
        \caption{Evaluation on  Arabic-Twitter. The best results are bolded, and the second-best results are underlined.}
        \label{Arabic}
        \begin{tabularx}{\textwidth}{c|*7{>{\centering\arraybackslash}X}|c}
        \toprule
        Metrics & PP-GCN & EventX & KPGNN & QSGNN & CLKD& \text{FinEvent} & \text{$\mathrm{RPLM}_{SED}$} & Improve\\
        \midrule
         NMI & $.32 \pm .02$ & $.28 \pm .00$ & $.75 \pm .02$ & $.78 \pm .02$ & $.81 \pm .02$ & $\underline{.82 \pm .01}$ & $\textbf{.93} \pm \textbf{.01}$ & $\uparrow .11$ \\ 
        AMI & $.32 \pm .02$ & $.03 \pm .00$ & $.74 \pm .01$ & $.77 \pm .01$ & $.81 \pm .02$ & $\underline{.82 \pm .01}$ & $\textbf{.93} \pm \textbf{.01}$ & $\uparrow .11$ \\ 
        ARI & $.31 \pm .02$ & $.03 \pm .00$ & $.75 \pm .02$ & $.79 \pm .02$ & $.84 \pm .02$ & $\underline{.86 \pm .01}$ & $\textbf{.95} \pm \textbf{.01}$ & $\uparrow .09$ \\ 
        \bottomrule
        \end{tabularx}
    \end{threeparttable}
\end{table}
For other baseline methods, we use pre-trained word embedding algorithm\footnote{\url{https://github.com/bakrianoo/aravec}} to convert Arabic messages into initial embeddings before training. 
To ensure a fair comparison, all methods utilize the K-Means clustering method.

As shown in Table~\ref{Arabic}, $\mathrm{RPLM}_{SED}$ outperforms the best results in the baseline by 13.41\%, 13.41\%, and 10.47\% on NMI, AMI, and ARI, respectively.
This is attributed to the inherent model advantage of $\mathrm{RPLM}_{SED}$ based on Multilingual Pretrained Language Models (MPLMs).
Compared to GNNs, MPLMs can understand and encode different languages.
This advantage enables $\mathrm{RPLM}_{SED}$ based on MPLMs to obtain better message embeddings when dealing with low-resource data, as well as stronger capabilities for encoding contextual messages.
Secondly, the pairwise learning mechanism, by consistently encoding both the content and structural information of messages, leverages more effective information.
This enables the model to obtain robust and stable message representations, even in situations where training resources are limited.

\subsection{Performance Evaluation on Long-tail Recognition Task}
\label{sec:Long-tail Recognition Task}
In real-world social data, the characteristics of a long-tail distribution are particularly evident. 
There are few head-dominated event classes, whereas low-frequency tail classes are more numerous. 
This imbalance in training data makes it difficult for tail events to be correctly classified.
In recent research, ETGNN~\cite{ren2022evidential} employs the Dempster-Shafer theory~\cite{sentz2002combination} to integrate the uncertainty and temporal information of graphs constructed from different perspectives, thereby enhancing the accuracy of event detection.
Furthermore, building upon ETGNN, the $\mathrm{UCL}_{SED}$~\cite{ren2023uncertainty} framework is proposed. 
Compared to ETGNN, $\mathrm{UCL}_{SED}$ introduces uncertainty-guided contrastive learning loss, pushing boundary-ambiguous event classes further away to improve the distinguishability of representations.
We also compare with MVGAN~\cite{cui2021mvgan}, which learns message representations from semantic and temporal views of messages and proposes a hashtag-based multi-view graph attention mechanism to fuse representations.

We conduct experiments on two long-tail datasets, Events2012\_100 and Events2018\_100.
It is noteworthy that to accommodate the classification task, an event classifier is additionally incorporated into $\mathrm{RPLM}_{SED}$. 
For the acquisition of message labels, we first filter out unreliable labels from their corresponding similarity sets based on similarity and then determine their final labels through a voting mechanism.
In terms of evaluation metrics, we select Accuracy (ACC) and F1 value (F1), which are widely used in classification tasks.
As illustrated in Table~\ref{tab:long tail}, compared with the optimal results among the baseline methods, $\mathrm{RPLM}_{SED}$ exhibits an improvement of 4.34\% in ACC and 5.43\% in F1, respectively, on Events2012\_100. 
On Events2018\_100, the improvements are 19.92\% and 16.46\%, respectively.
Recalling the pairwise message modeling process, we sample multiple messages for each message to construct message pairs. 
Utilizing the multi-relational prompt-based pairwise message learning mechanism, each message is learned multiple times from different message pairs. 
This allows messages of tail events to effectively learn a common representation of similar messages in terms of semantic and structural relations, as well as the subtle differences between different classes of messages. 
Additionally, the clustering constraint forces the representations of each event class to be as dispersed as possible in the latent space while ensuring that representations of messages within the same class are as close as possible. 
This further enhances the distinguishability of message representations, thereby enabling the event classifier to predict labels with greater accuracy.
Furthermore, for each message, by filtering out unreliable labels from its corresponding set of candidate labels based on similarity, the label subsequently obtained through voting from the remaining labels is more reliable.
\begin{table}[h]
    \centering
    \setlength{\tabcolsep}{2pt} 
    \begin{threeparttable}
        \caption{Evaluation on the Long-tail Set. The best results are bolded, and the second-best results are underlined.}
        \label{tab:long tail}
        \renewcommand\arraystretch{1}
        \begin{tabularx}{\textwidth}{c|c|*{8}{>{\centering\arraybackslash}X}|c}
        \toprule
        Dataset&Metrics & PPGCN & KPGNN & QSGNN & FinEvent& MVGAN & ETGNN &$\mathrm{UCL}_{SED}$ & $\mathrm{RPLM}_{SED}$ &Improve\\
        \midrule
        \multirow{2}{*}{\makecell{Events\\2012\_100}}
        ~ & ACC & $.64\pm .00$ & $.73 \pm .01$ & $.75\pm .01$ & $.81\pm .01$ &$.82\pm .00$ & $.87\pm .01$ & \underline{$.92 \pm .01$} & $\textbf{.96}\pm \textbf{.01}$ & $\uparrow .04$ \\ 
        ~ & F1 & $.55 \pm .00$ & $.60 \pm .00$ & $.69\pm .01$ & $.80\pm .01$ &$.82\pm .01$ & $.87\pm .00$ & \underline{$.92\pm .01$ }& $\textbf{.97}\pm \textbf{.01}$ & $\uparrow .05$ \\  
        \midrule
        \multirow{2}{*}{\makecell{Events\\2018\_100}}
        ~ & ACC & $.71 \pm .01$ & $.77 \pm .01$ & $.77\pm .01$ & $.63\pm .00$ &$.68\pm .01$ & $.61\pm .01$ & \underline{$.79\pm .01$ }& $\textbf{.90}\pm \textbf{.01}$ & $\uparrow .11$ \\ 
        ~ & F1 & $.51 \pm .00$ & $.62 \pm .01$ & $.75\pm .01$ & $.63\pm .01$ & $.68\pm .01$& $.60\pm .01$ & \underline{$.79\pm .01$ }& $\textbf{.92}\pm \textbf{.01}$ & $\uparrow .12$ \\ 
      
        \bottomrule
        \end{tabularx}
    \end{threeparttable}
\end{table}

\subsection{PLM-based Model for Social Event Detection}
\label{sec:PLM-based Model for Social Event Detection}
In the preceding sections, we comprehensively compare $\mathrm{RPLM}_{SED}$ against the currently dominant social event detection methods based on GNNs across various scenarios.
To our knowledge, $\mathrm{RPLM}_{SED}$ is the inaugural model leveraging PLMs to achieve high-quality social event detection in diverse scenarios, including offline, online, low-resource settings, and long-tail distributions.
To further explore the performance of Pre-trained Language Models (PLMs) in the social event detection tasks, we design experiments utilizing the pre-trained word embedding technology Word2Vec and two mainstream PLMs, BERT and RoBERTa.
Specifically, \revised{for Word2Vec, BERT, and RoBERTa, which only use the message content as input, the final representation is obtained by averaging the embeddings of each word in the message.
Additionally, we implement two distinct strategies for the original BERT and RoBERTa without fine-tuning and fine-tuning strategies.}
Furthermore, we propose variant models BERT-LR and RoBERTa-LR for $\mathrm{RPLM}_{SED}$ based on BERT and RoBERTa, respectively, which incorporate fine-tuning through LoRA.
In practical applications, given the infeasibility of the K-Means clustering method due to its requirement for pre-specifying the number of events, we consistently employed the HDBSCAN clustering method. 
It is important to note that $\mathrm{RPLM}_{SED}$ is predicated on RoBERTa-base, and similarly, all other models also utilize their respective base versions.

\begin{figure}[h]
    \centering
    \includegraphics[width=1\linewidth]{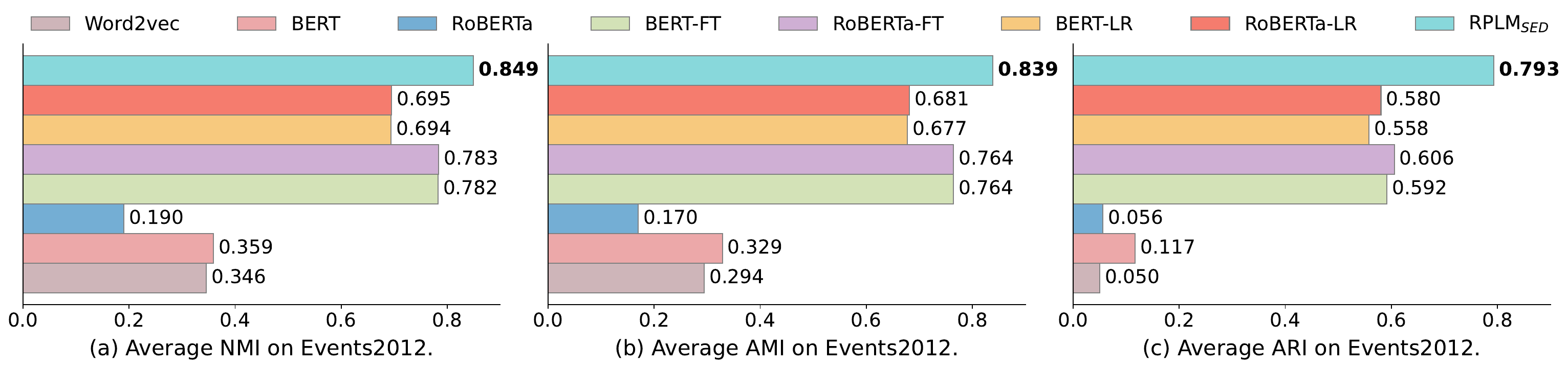}
    \caption{Overall Performance of Various PLM-based Models on Events2012 in Online Scenario. \revised{The horizontal axis represents the average metric values of different methods, and the best results are bolded.}}
    \label{fig:PLM-based Model}
\end{figure}

As illustrated in Figure~\ref{fig:PLM-based Model}, among the methods for directly obtaining message representations from Word2Vec, BERT, and RoBERTa, BERT demonstrates the relatively best performance.
This may be attributed to the higher compatibility between the knowledge acquired during the BERT pre-training process and the Events2012 dataset in the absence of model fine-tuning.
Upon fine-tuning, BERT-FT and RoBERTa-FT exhibit significant enhancements, achieving nearly equivalent model performances.
Nonetheless, a discernible performance gap remains when compared to $\mathrm{RPLM}_{SED}$, attributable to their exclusive reliance on the semantic information of messages while neglecting the abundant structural information within the messages.
Moreover, we observe that the performances of the variants BERT-LR and RoBERTa-LR fall short of those achieved through global fine-tuning of the models.
In online event detection scenarios, models necessitate continual updates to incorporate new knowledge while concurrently sustaining an existing knowledge base.
Utilizing LoRA to fine-tune models can effectively decrease the number of parameters requiring adjustments during the fine-tuning process, which aids in enhancing the efficiency of model updates and reducing computational costs. 
However, this strategy of reducing the volume of trainable parameters may adversely affect the model's long-term capabilities for knowledge learning, retention, and expansion, consequently leading to a decline in model performance.
Overall, RPLM's model performance significantly surpasses that of other methods, further illustrating the efficacy of the multi-relational prompt-based pairwise message learning mechanism. 
On the other hand, the experiments also validate the feasibility of utilizing PLMs for social event detection tasks.

\begin{figure}[b]
    \centering
    \subfigure{
        \includegraphics[width=0.95\linewidth, trim={0 1.2cm 0 1.2cm}, clip]{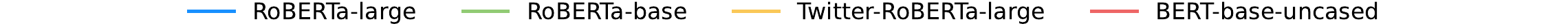}
    }
    \vspace{-10pt}

    \setcounter{subfigure}{0}
     \subfigure[NMI value of each message block on Events2012.]{
        \includegraphics[width=0.9\linewidth, trim={0.3cm 0.3cm 0.25cm 0}, clip]{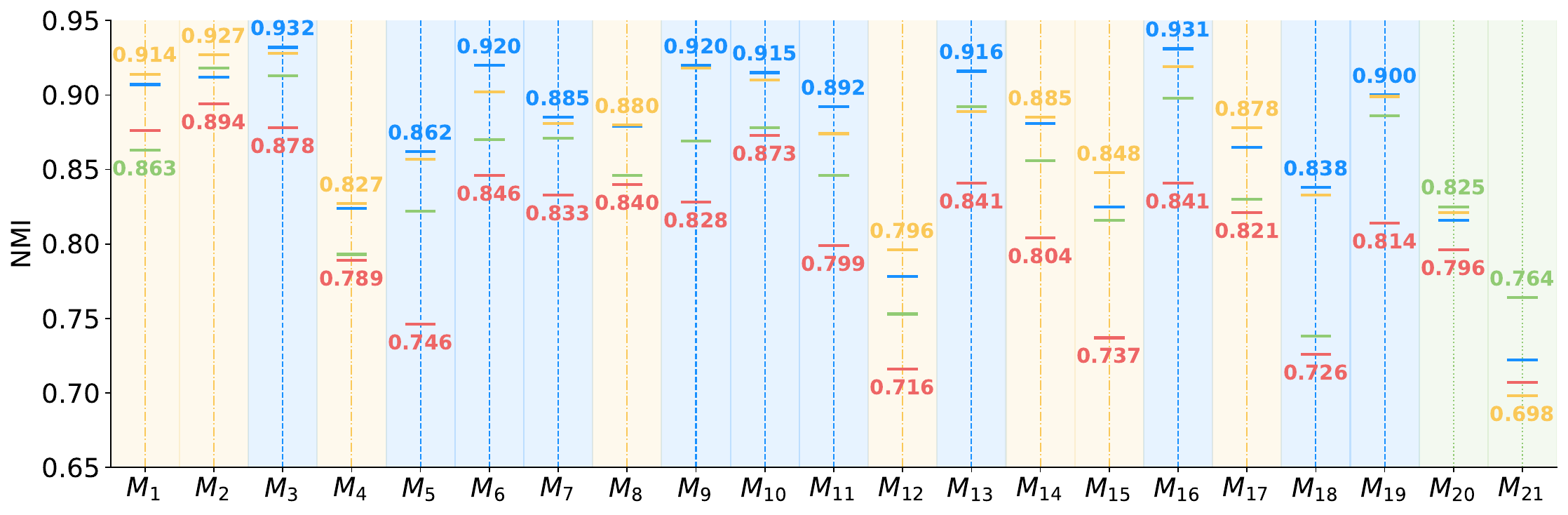}
    }
    \vspace{-10pt}
    
     \subfigure[AMI value of each message block on Events2012.]{
        \includegraphics[width=0.9\linewidth, trim={0.3cm 0.3cm 0.25cm 0}, clip]{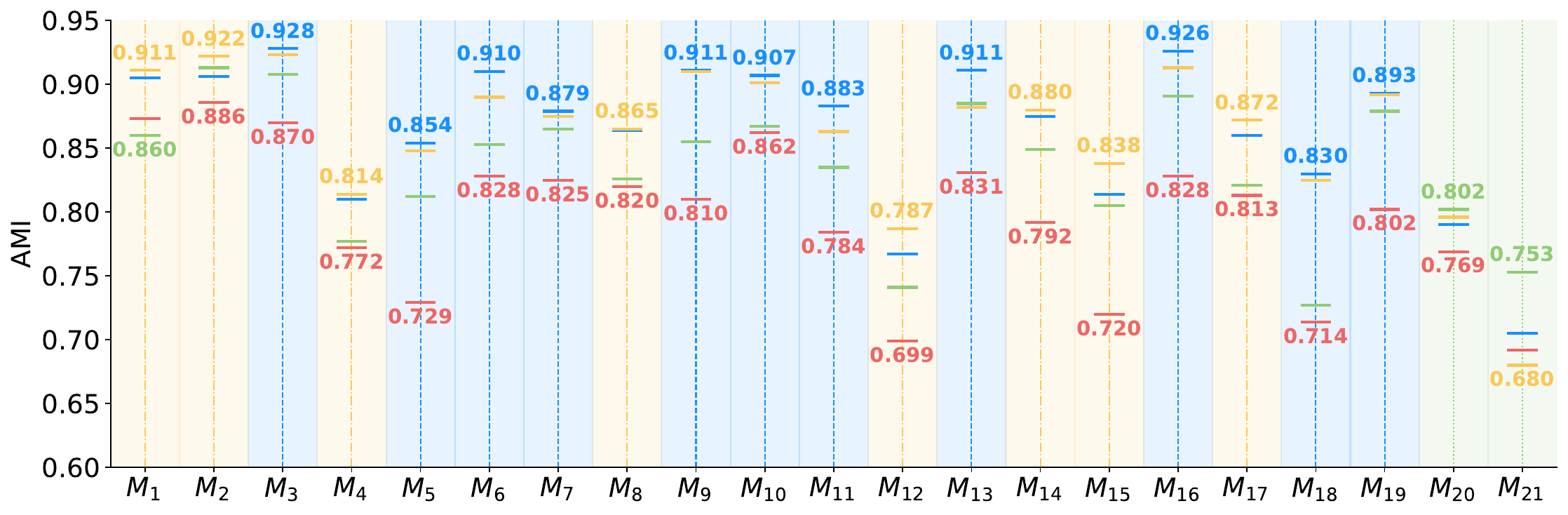}
    }
    \vspace{-10pt}
    
     \subfigure[ARI value of each message block on Events2012.]{
        \includegraphics[width=0.9\linewidth, trim={0.3cm 0.3cm 0.25cm 0}, clip]{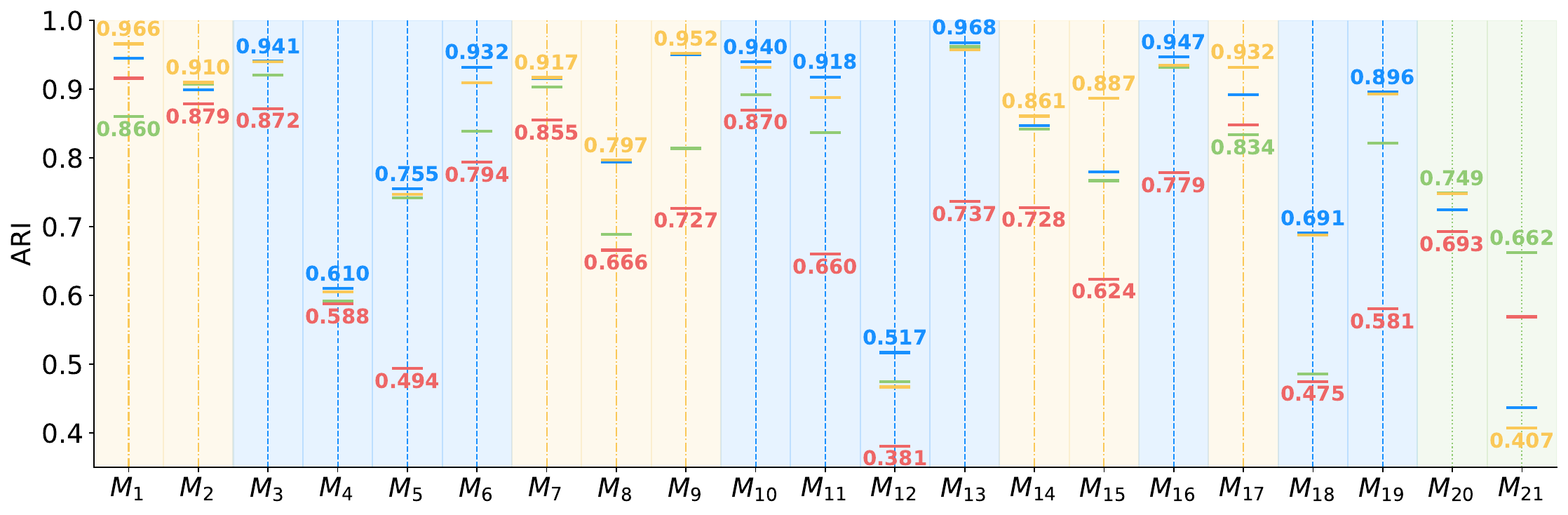}
    }

    \caption{\revised{$\mathrm{RPLM}_{SED}$ performance comparison based on different PLMs.
    The horizontal axis represents the message blocks, while the vertical axis denotes each message block's corresponding metric value. 
    Each message block's background color corresponds to the model's color with the highest score within that block.}}
    \label{fig:diff-plm}
\end{figure}
\subsection{$\mathrm{RPLM}_{SED}$ with Different PLMs}
\label{subsec:RPLM-Event with Different PLMs}
In investigating the variations in the performance of $\mathrm{RPLM}_{SED}$ under different PLMs, we conduct experiments on the Events2012 dataset utilizing four distinct models: RoBERTa-large~\cite{liu2019roberta}, RoBERTa-base, Twitter-RoBERTa-large~\cite{loureiro2023tweet}, and BERT-base-uncased~\cite{devlin-etal-2019-bert}. 
The experiments are designed to encompass three dimensions: model scale, pre-training corpora, and model architecture. 
To ensure consistency in comparison, the HDBSCAN method is utilized for clustering in all experiments.

As depicted in Figure~\ref{fig:diff-plm}, compared to $\mathrm{RPLM}_{SED}$ based on RoBERTa-base, the model based on RoBERTa-large exhibits overall improvements of $3.23\%$, $3.45\%$, and $4.68\%$ on the NMI, AMI, and ARI, respectively. 
Compared to RoBERTa-base, RoBERTa-large possesses a larger model size, more trainable parameters, and a broader training dataset, resulting in superior model performance. 
Naturally, the $\mathrm{RPLM}_{SED}$ based on RoBERTa-large requires greater computational resources and training time.
The primary distinction between Twitter-RoBERTa-large and RoBERTa-large lies in the difference in training corpora. 
Twitter-RoBERTa-large is specifically trained on Twitter data, whereas RoBERTa-large undergoes pre-training on a broader spectrum of text data. 
These two models have the same model scale and have been further fine-tuned on the same dataset, hence their overall performance is essentially consistent.
When contrasting RoBERTa-base with BERT-base-uncased, two distinct types of pre-trained language models, we observe that $\mathrm{RPLM}_{SED}$ based on RoBERTa-base exhibits superior performance. 
In terms of overall performance, the $\mathrm{RPLM}_{SED}$ based on RoBERTa-base surpasses the $\mathrm{RPLM}_{SED}$ based on BERT-base-uncased by 4.36\%, 4.81\%, and 12.15\% in NMI, AMI, and ARI, respectively.
This advantage can be attributed to the pre-training of the RoBERTa-base on a larger dataset and optimized training strategies.
Notably, even the $\mathrm{RPLM}_{SED}$ based on BERT-base-uncased outperforms the strongest baseline model, FinEvent, leading by 2.81\% in NMI, 2.51\% in AMI, and 8.75\% in ARI.

\begin{figure}[h]
    \centering
    \includegraphics[width=1\linewidth]{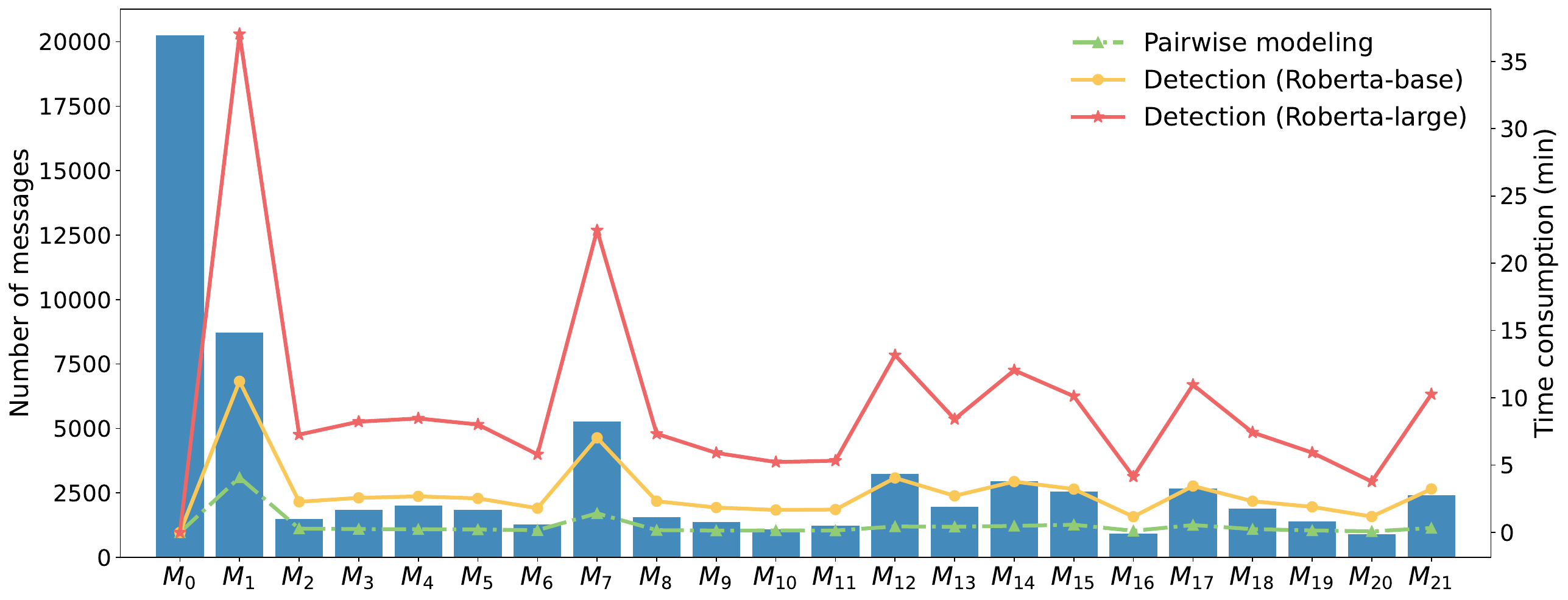}
    \caption{\revised{The number of messages in each message block (histogram) and the time consumption of pairwise modeling and detection for each block (line chart) in Events2012.}}
    \label{fig:time}
\end{figure}

\revised{
\subsection{Time Consumption of $\mathrm{RPLM}_{SED}$ based on Different Scale PLMs}
\label{sec: report time}
This section reports the time consumption of different parts of $\mathrm{RPLM}_{SED}$ based on RoBERTa-large and RoBERTa-base on the Events2012.
Figure~\ref{fig:time} shows the message number of each block, the time consumption for pairwise messages modeling during the detection phase, and the execution time for detection operations. 
$M_0$ is solely used for initial model training and incurs no time consumption during detection.
Additionally, Table~\ref{tab:time} presents statistics on message numbers within each window, the time required for pairwise message modeling during training, and the time consumption for model training or maintenance in each window.
Both Figures~\ref{fig:time} and Table~\ref{tab:time} show that the time expenditure for model training or detection correlates linearly with the number of messages in message blocks.
Secondly, whether in the training or detection phase, the time spent on pairwise message modeling is significantly less than the time spent on model training or detection.
Furthermore, the training and detection time of $\mathrm{RPLM}_{SED}$ based on RoBERTa-large is approximately 3.2 times that of $\mathrm{RPLM}_{SED}$ based on RoBERTa-base. 
This is attributed to RoBERTa-large's larger hidden dimensions and more Transformer layers, which enhance its performance compared to RoBERTa-base-based $\mathrm{RPLM}_{SED}$.
It should be noted that in addition to the employed PLM scale, factors such as message sampling numbers and hardware environment also markedly influence the overall time consumption of the model.}

\begin{table}[h]
    \centering
    \begin{threeparttable}
        \setlength{\tabcolsep}{4pt} 
        \caption{\revised{The number of messages in each window and the time consumption of pairwise modeling and training (or maintenance) for each window in Events2012.}}
        \label{tab:time}
        \begin{tabularx}{\textwidth}{c|*8{>{\centering\arraybackslash}c}}
        \toprule
         \multicolumn{9}{c}{Message Number}\\ 
         \midrule
        Window & $M_0$ & $M_1 - M_3$ & $M_4 - M_6$ & $M_7 - M_9$ & $M_{10} - M_{12}$& $M_{13} - M_{15}$ & $M_{16} - M_{18}$ & $M_{19} - M_{21}$\\
        \midrule
         Number &20254 &12048 &5120 &8201 &5565 &7477 &5473 & 4702\\ 
         \midrule
         \multicolumn{9}{c}{Time Consumption (min)}\\
         \midrule
         Window & $M_0$ & $M_1 - M_3$ & $M_4 - M_6$ & $M_7 - M_9$ & $M_{10} - M_{12}$& $M_{13} - M_{15}$ & $M_{16} - M_{18}$ & $M_{19} - M_{21}$\\
         \midrule
         Pairwise Modeling&4.55 &3.53 &0.52 &1.52 & 0.57& 1.20& 0.79& 0.47\\
         RoBERTa-base &55.06 &32.88 &14.73 &22.28 &15.08 &19.95 &14.7 & 12.58\\ 
         RoBERTa-large &175.20 &108.25 &46.35 &72.24 &50.78 &65.23 &48.80 &40.45 \\ 
        \bottomrule
        \end{tabularx}
    \end{threeparttable}
\end{table}
\subsection{Ablation Study}
\label{sec:Ablation Study}
In this section, we conduct ablation experiments on four critical components of the $\mathrm{RPLM}_{SED}$ model to demonstrate the indispensability of each component. 
To ensure consistency of results, all experiments are conducted on the Events2012 dataset and utilize the HDBSCAN clustering method.
To investigate the impact of different components on model performance, we sequentially conduct ablation experiments on four parts:  \textbf{Multi-relational Prompt Embedding}, \textbf{Similarity-based Representation Aggregation}, \textbf{Multi-Head Structured Attention}, and \textbf{Clustering Constraint}. 
The results of these experiments are reported in Table~\ref{tab:Ablation}.

\textbf{Multi-relational Prompt Embedding.}
As illustrated in `Prompt Embedding' within the `Pairwise Message Learning' part of Figure~\ref{fig:model}.
The task of multi-relational prompt embedding is to transform the structural relations between messages into multi-relational prompt embeddings.
When removing the multi-relational prompt embedding, we directly sample messages from the original message data to form message pairs.
As shown in Table~\ref{tab:Ablation}, after removing the multi-relational prompt embedding, the model experiences a decrease of $2.05\%$, $2.08\%$, and $2.07\%$ in NMI, AMI, and ARI, respectively.
Social messages contain rich structural information. 
After removing the multi-relational prompt embedding, due to the absence of multi-relational prompt guidance, the model can only learn representations from the semantics of messages, thereby affecting model performance.

\textbf{Similarity-based Representation Aggregation. }
We investigate the variations in model performance under different representation aggregation methods, including similarity-based representation aggregation method, top-k sampling representation aggregation method, and \revised{all representation average aggregation method}. 
For a set of candidate feature representations obtained for each message, under the top-k sampling representation aggregation method, we only sample the top 20 candidate representations with the highest similarities for averaging. 
This approach has certain limitations.
A smaller value of $k$ may lead to message representations that are not comprehensive enough. 
Conversely, a larger $k$ may include candidate representations extracted from negative sample pairs, which often carry noise and irrelevant information, thereby affecting the quality of the final message representations.
For the all-representation average aggregation method, averaging all candidate representations of messages may lead to overly smoothed representations and is also susceptible to noise from candidate representations extracted from negative sample pairs. 
Consequently, among the three methods, it performs the worst.
In contrast, the similarity-based representation aggregation approach averages candidate representations of messages with high confidence, effectively filtering out noisy candidate representations extracted from negative sample pairs while preventing message representations from becoming overly smoothed.
Overall, the similarity-based representation aggregation approach outperforms the top-k sampling representation aggregation by $1.04\%$, $1.05\%$, and $2.89\%$ in NMI, AMI, and ARI, respectively.
Additionally, it surpassed the all representation average aggregation method by 3.68\%, 3.84\%, and 8.16\% on NMI, AMI, and ARI, respectively.

\textbf{Multi-Head Structured Attention. }
The role of the multi-head structured attention mechanism is to extract key features from different subspaces of message representations while eliminating irrelevant features and noise to enhance the clarity of the representations. 
When the multi-head structured attention mechanism is removed, the resulting candidate message representations may contain noise, affecting the classifier's accuracy in discriminating message similarity. 
Additionally, it may also impact the quality of the final message representations.
Consequently, when the multi-head structured attention mechanism is removed, the values of NMI, AMI, and ARI decrease by $2.06\%$, $2.31\%$, and $2.31\%$, respectively. 
This reduction substantiates the pivotal role of the attention mechanism in bolstering both representation clarity and classifier precision.

\textbf{Clustering Constraint.} 
Our proposed clustering constraint consists of inter-cluster and intra-cluster loss, which enhances the distinguishability of message representations by optimizing the training process. 
Upon removing the clustering constraint, we find that relying solely on pairwise cross-entropy loss to constrain model training is insufficient. 
This insufficiency stems from the fact that pairwise cross-entropy loss merely treats messages as objects to be pulled closer or pushed apart, making it difficult to ensure cohesiveness within event clusters and separation between them. 
After the removal of the Clustering Constraint, the NMI, AMI, and ARI decreased by $6.29\%$, $6.71\%$, and $9.25\%$, respectively.
\begin{table}[h]
    \centering
    \setlength{\tabcolsep}{5pt} 
 \begin{threeparttable}
        \caption{Ablation study for Multi-relational Prompt Embedding, Similarity-based Representation Aggregation, Multi-Head Structured Attention, and Clustering Constraint. The best results are bolded, and the second-best results are underlined.}
        \label{tab:Ablation}
        \begin{tabularx}{\linewidth}{c*{7}{>{\centering\arraybackslash}c}}
            \toprule
             &\multirow{2}{*}{\makecell{Multi-relational\\Prompt Embedding}} &\multirow{2}{*}{\makecell{Representation \\Aggregation Strategy}} &\multirow{2}{*}{\makecell{Multi-head\\Structured Attention }} &\multirow{2}{*}{\makecell{Clustering \\ Constraint}}& \multicolumn{3}{c}{Avg. Metrics}  \\
             \cline{6-8}
            &&&&&NMI& AMI&ARI\\
            \midrule
            1&\checkmark&$Similarity$&$\checkmark$&$\checkmark$&$\textbf{0.874}$&$\textbf{0.865}$&$\textbf{0.822}$\\
            2&-&$Similarity$&$\checkmark$ &$\checkmark$&$0.856$&$0.847$&$\underline{0.805}$\\
            3&$\checkmark$&$Top-k$&$\checkmark$ &$\checkmark$&$\underline{0.865}$&$\underline{0.856}$&$0.799$\\
            4&$\checkmark$&$Average$&$\checkmark$ &$\checkmark$&$0.843$&$0.833$&$0.760$\\
            5&$\checkmark$&$Similarity$& - &$\checkmark$&$0.856$&$0.845$&$0.803$\\
            6&$\checkmark$&$Similarity$&$\checkmark$ &-&$0.819$&$0.807$&$0.746$\\
            \bottomrule
        \end{tabularx}
    \end{threeparttable}
\end{table}

\subsection{Hyperparameter Study}
\label{subsec:Hyperparameter Study}
\begin{figure}[t]
    \centering
    \subfigure[Intra-cluster Loss Weight ($\lambda$).]{
        \includegraphics[width=0.49\linewidth,trim={0 0.4cm 0 0}, clip]{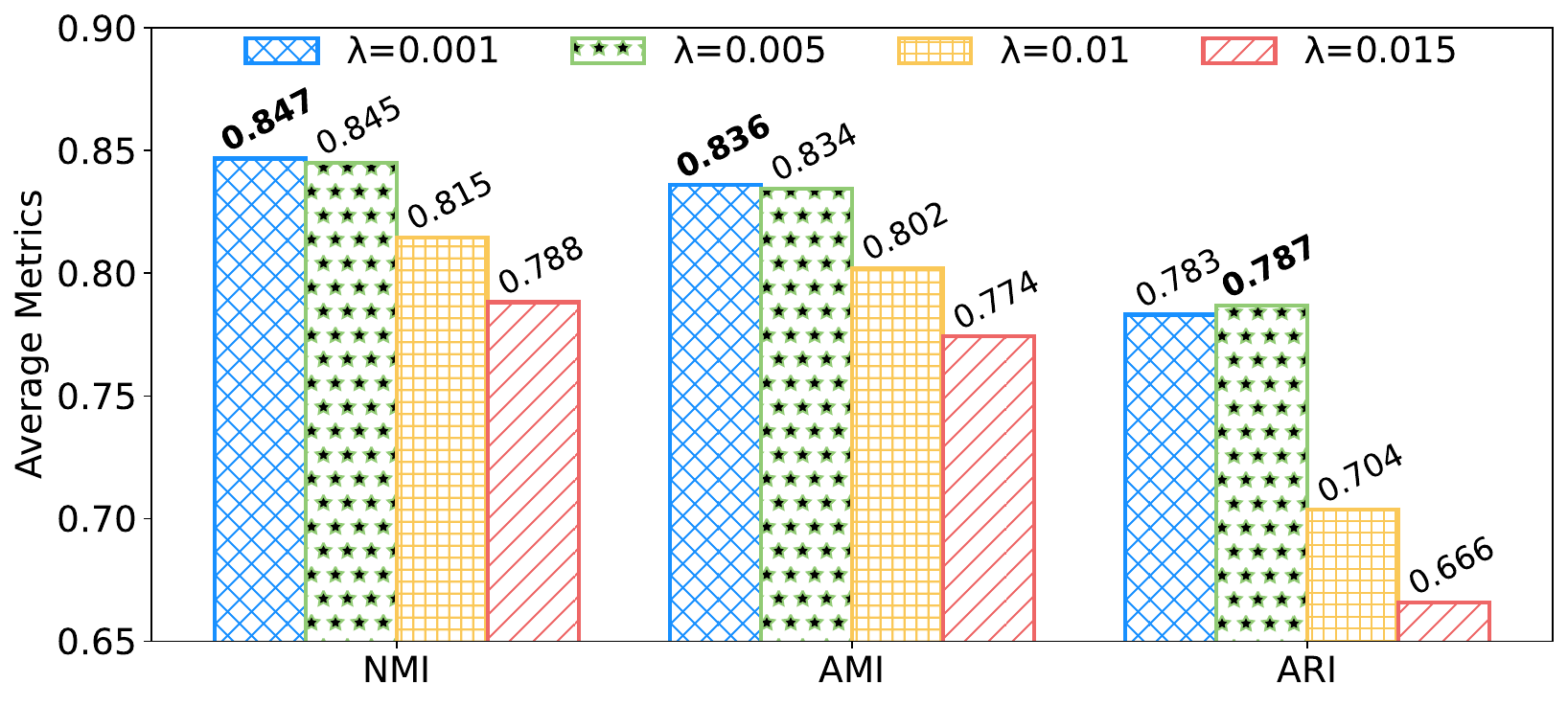}
    }
     \hspace{-10pt}
    \subfigure[Inter-cluster Loss Weight ($\mu$).]{
        \includegraphics[width=0.49\linewidth,trim={0 0.4cm 0 0}, clip]{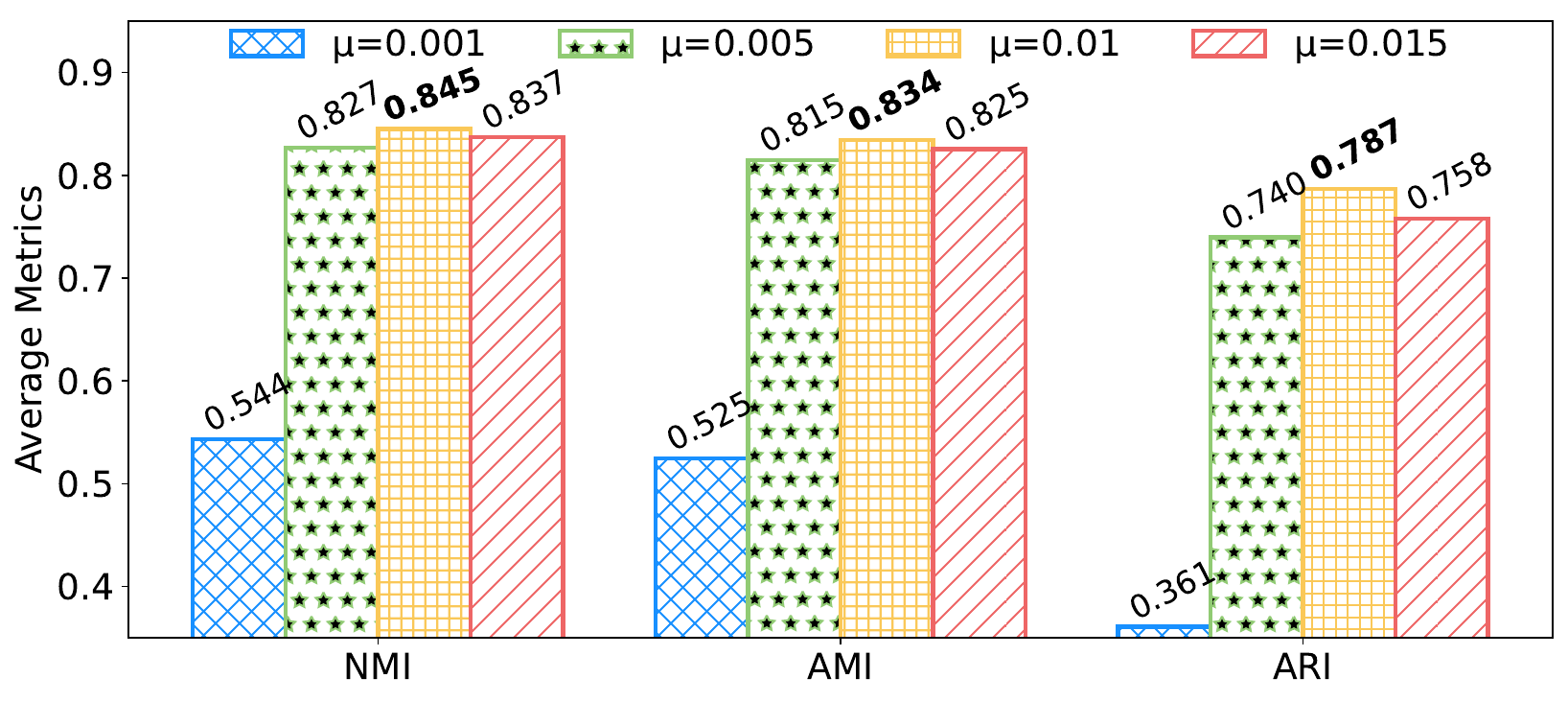}
    }
   \vspace{-10pt}
    
    \subfigure[Pairwise Cross-Entropy Loss Weight ($\kappa$).]{
        \includegraphics[width=0.49\linewidth,trim={0 0.4cm 0 0}, clip]{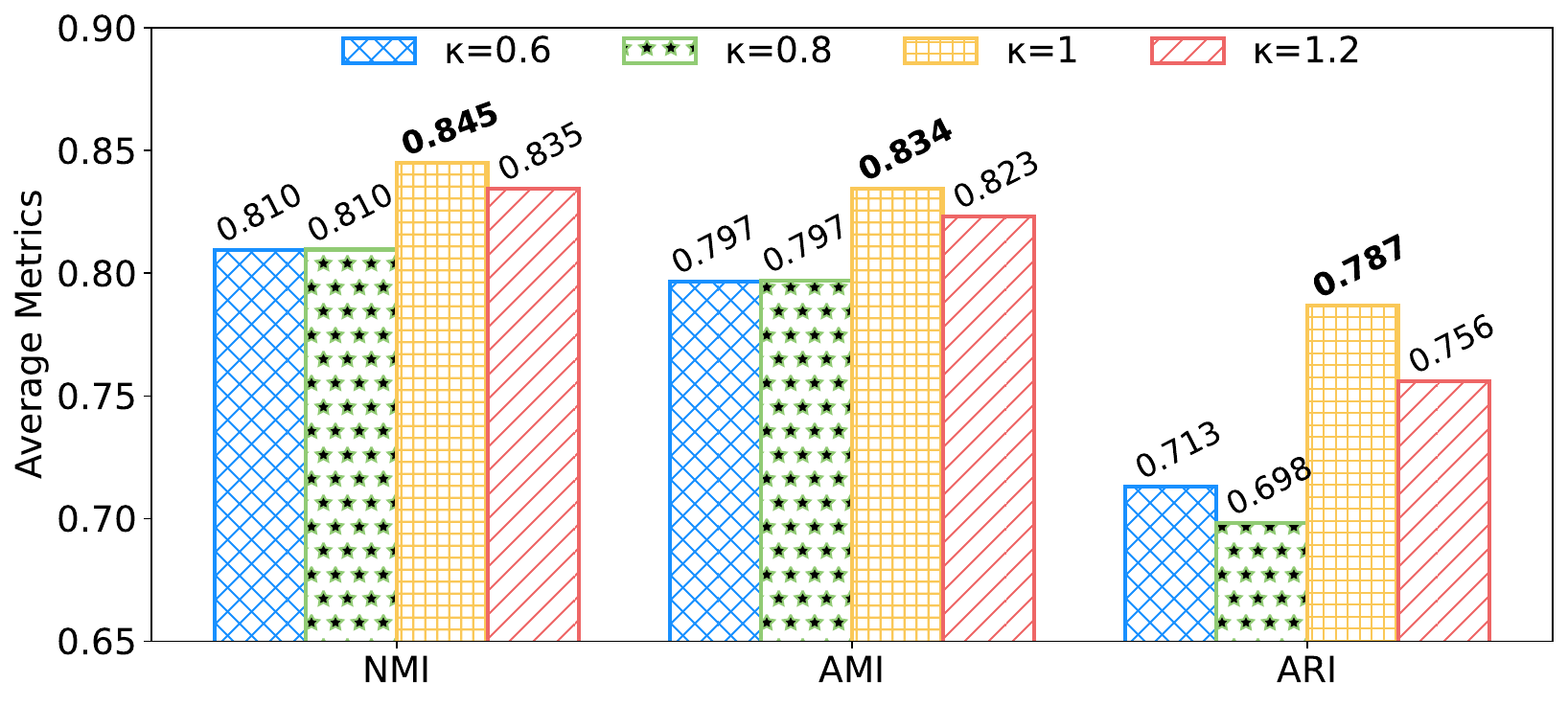}
    }
    \hspace{-10pt}
    \subfigure[Similarity Threshold ($\alpha$).]{
        \includegraphics[width=0.49\linewidth,trim={0 0.4cm 0 0}, clip]{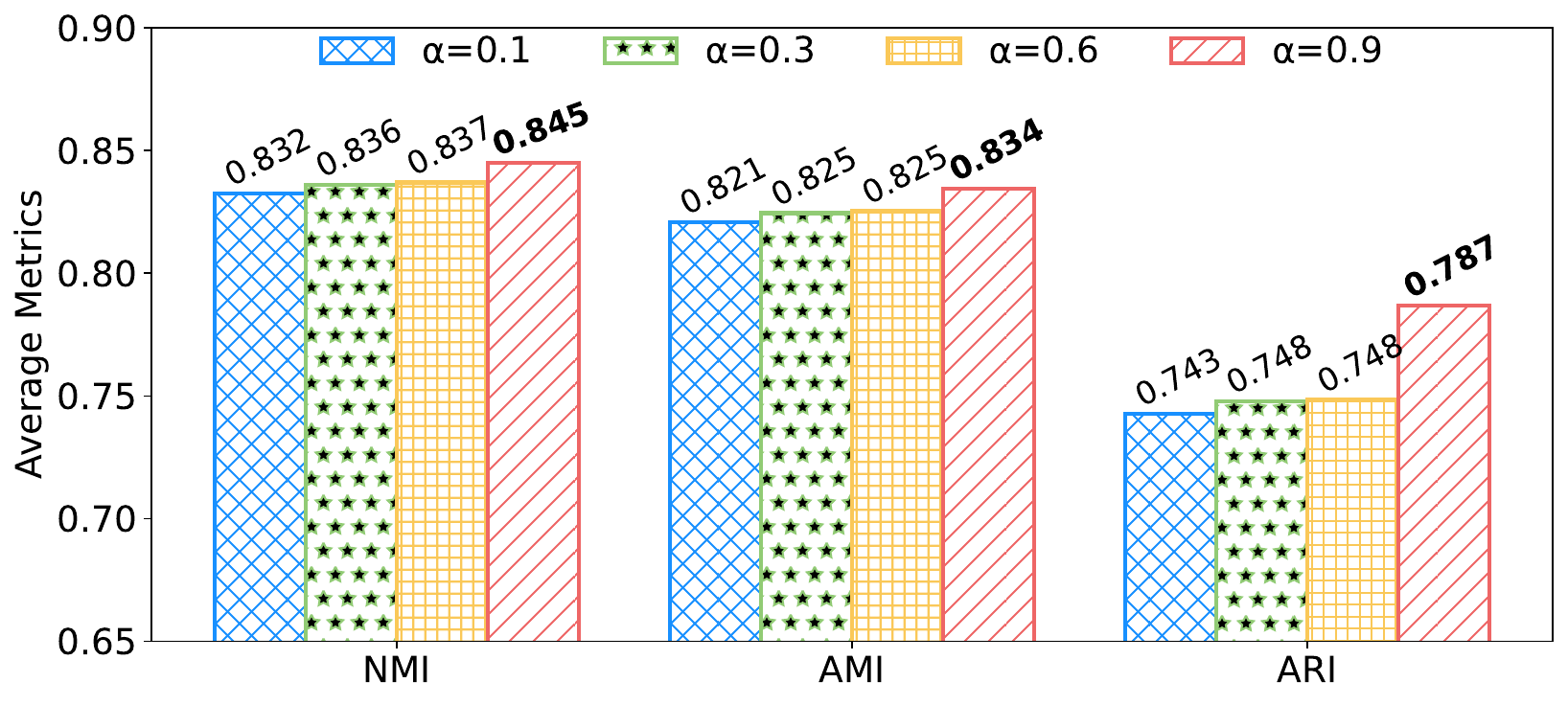}
    }
    \vspace{-10pt}
    
    \subfigure[Number of Structured Attention Heads ($o$).]{
        \includegraphics[width=0.49\linewidth,trim={0 0.4cm 0 0}, clip]{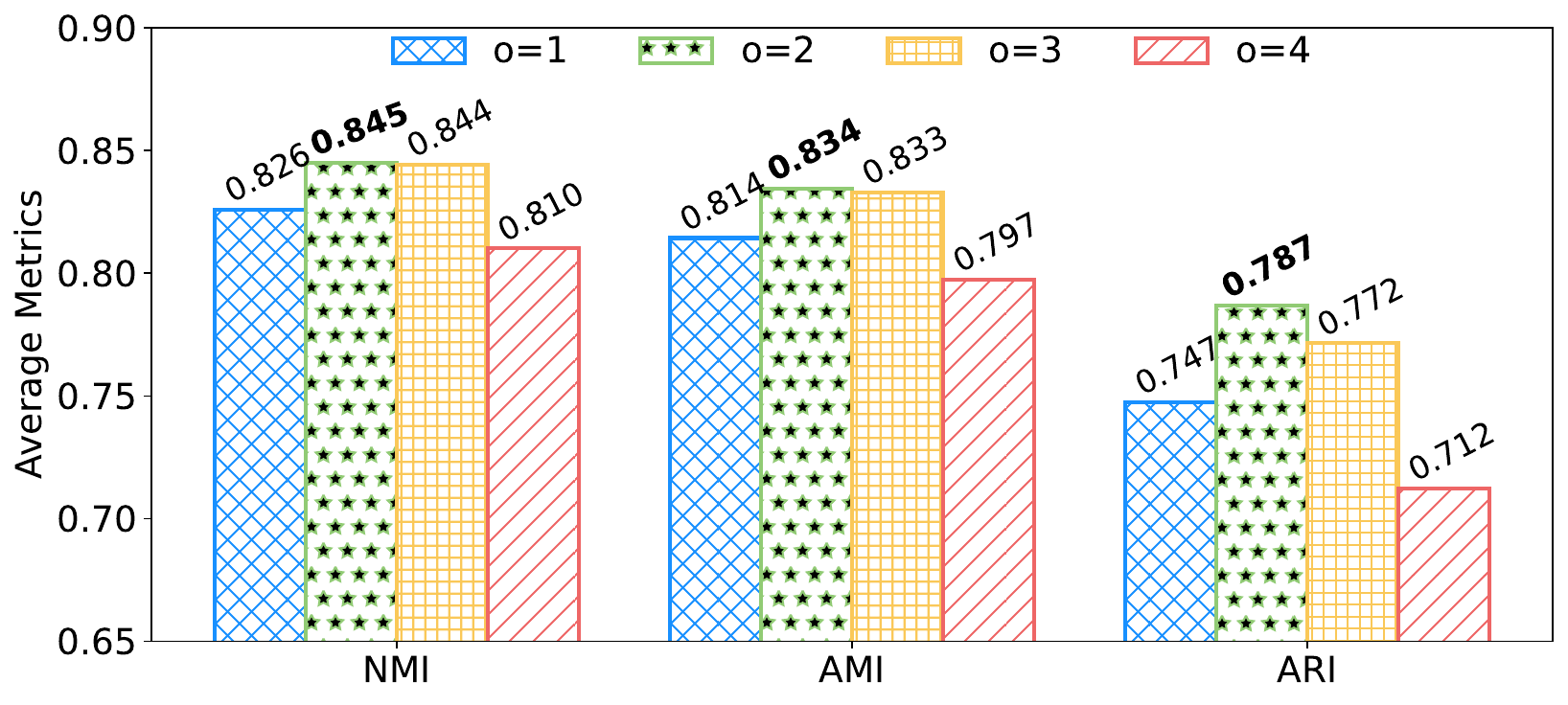}
    }
    \hspace{-10pt}
    \subfigure[Parameter Update Weight ($\zeta$).]{
        \includegraphics[width=0.49\linewidth,trim={0 0.4cm 0 0}, clip]{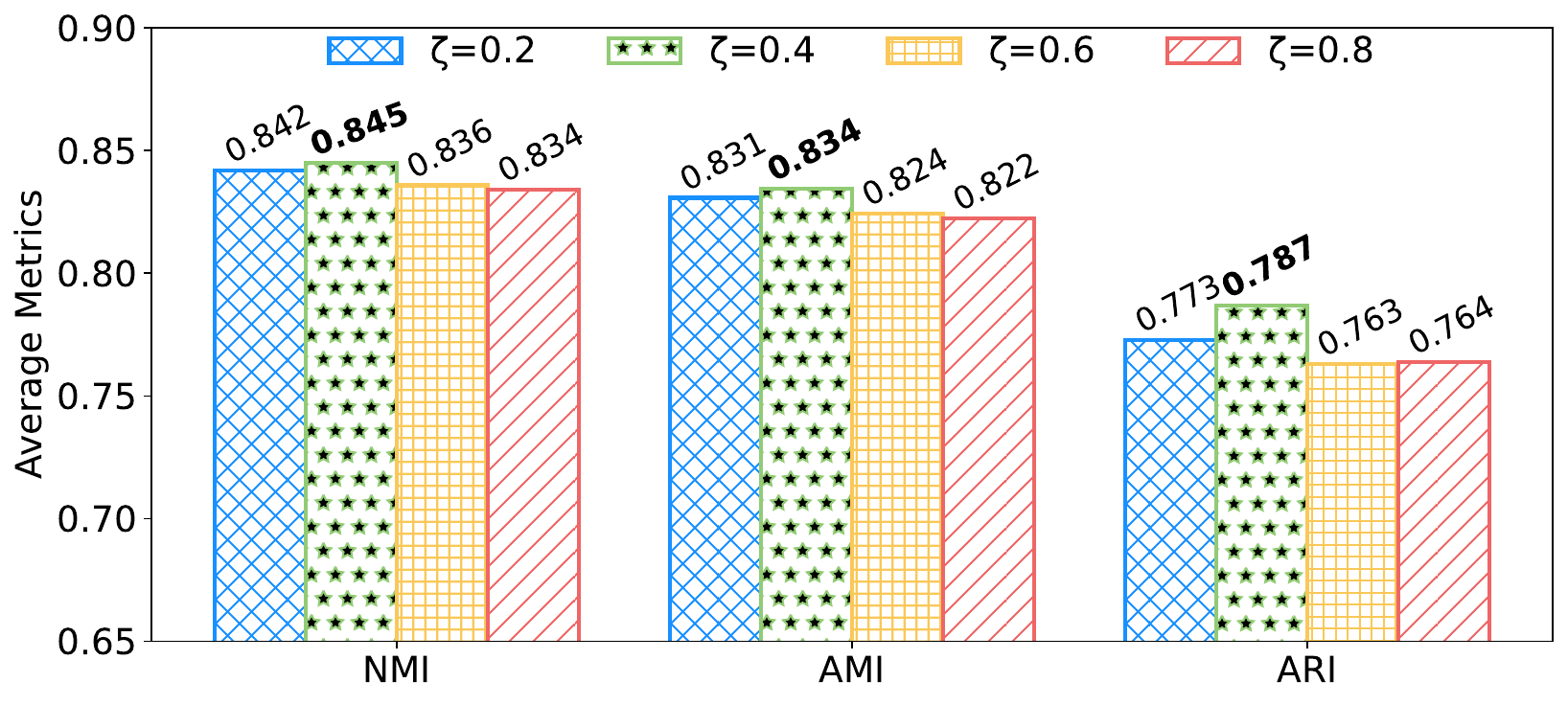}
    }
    \vspace{-10pt}
    
    \subfigure[Number of message sampled ($y$).]{
        \includegraphics[width=0.49\linewidth,trim={0 0.4cm 0 0}, clip]{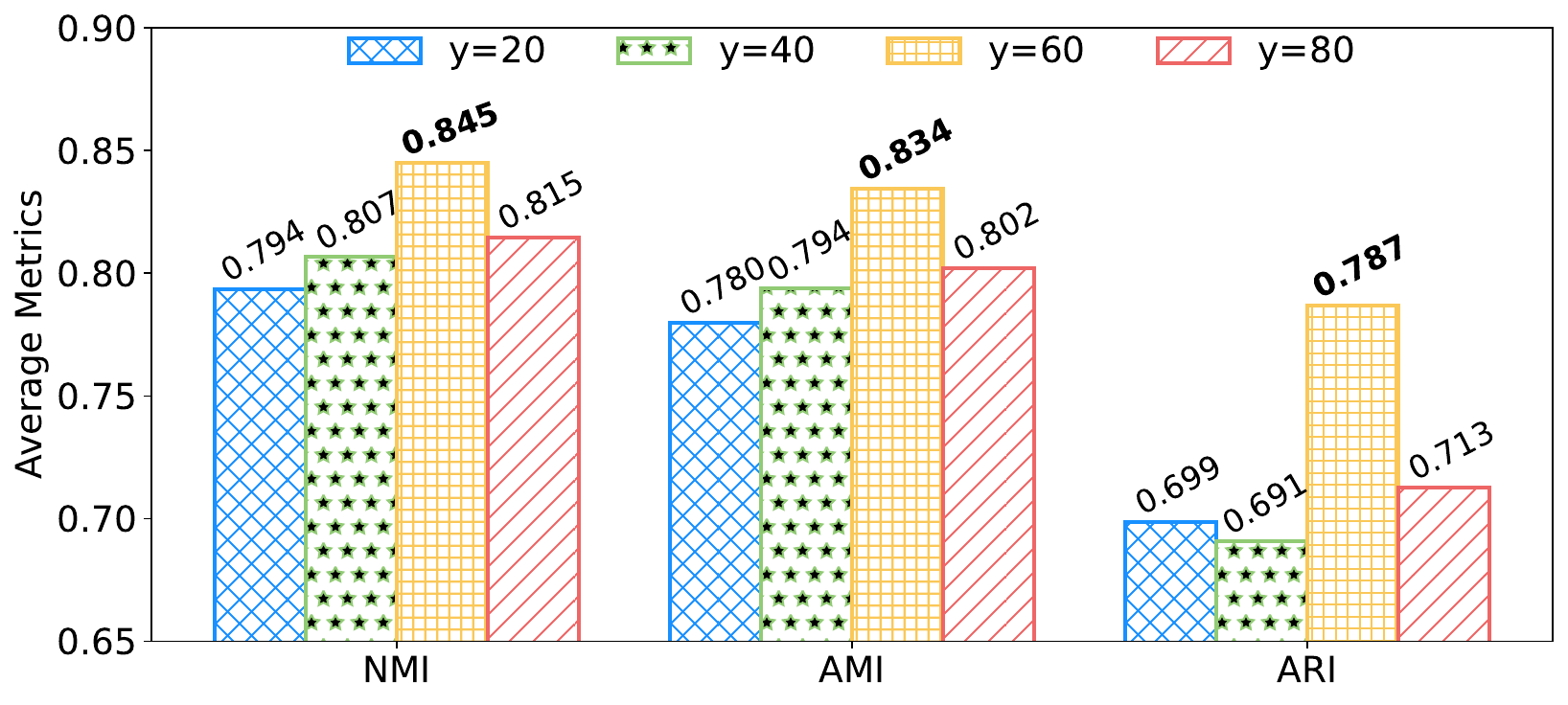}
    }
    \hspace{-10pt}
    \subfigure[Window Size ($w$).]{
        \includegraphics[width=0.49\linewidth,trim={0 0.4cm 0 0}, clip]{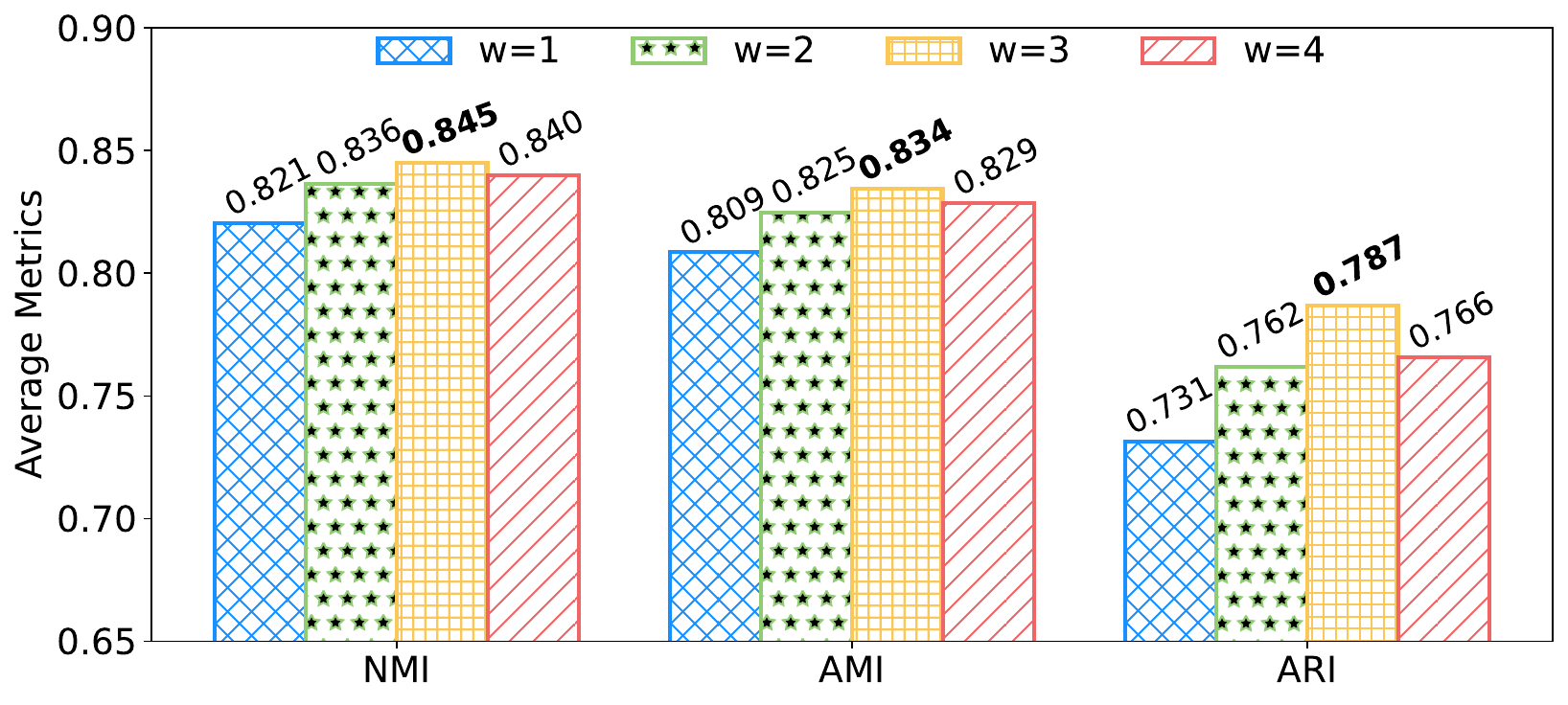}
    }
    \caption{\revised{$\mathrm{RPLM}_{SED}$ with different hyperparameters.
    The horizontal axis represents the metrics, while the vertical axis denotes
each hyperparameter’s corresponding average metric values. The best results are bolded.}}
    \label{fig:Hyperparameter}
\end{figure}

We explore the impact of \revised{eight} hyperparameters on model performance in the online detection scenario. 
\revised{
All experiments are conducted on the Events2012 dataset and utilize the HDBSCAN clustering method.
These hyperparameters include \textbf{Intra-cluster Loss Weight ($\lambda$)}, \textbf{Inter-cluster Loss Weight ($\mu$)}, \textbf{Pairwise Cross-Entropy Loss Weight ($\kappa$)}, \textbf{Similarity Threshold ($\alpha$)}, \textbf{Number of Structured Attention Heads ($o$)}, \textbf{Parameter Update Weight ($\zeta$)} during the maintenance phase, \textbf{Number of message sampled ($y$)}, and \textbf{Window Size ($w$)}. 
To ensure experimental consistency, all parameters except the one under study are kept fixed, with their initial values set to 0.005, 0.01, 1, 0.9, 2, 0.4, 60, and 3, respectively.}
Due to computational resource constraints, all hyperparameter experiments are conducted under the $\mathrm{RPLM}_{SED}$  based on RoBERTa-base. 
The experimental results are shown in Figure~\ref{fig:Hyperparameter}.

\textbf{Intra-cluster Loss Weight ($\lambda$).}
When examining the effect of different intra-cluster loss weights on model performance, we select four weights: 0.001, 0.005, 0.01, and 0.015. As shown in Figure~\ref{fig:Hyperparameter}(a), the model performance remains stable when $\lambda$ is set to 0.001 or 0.005. 
However, as $\lambda$ increases to 0.01 or higher, a linear decline in model performance is observed. 
This suggests that an excessively high intra-cluster loss weight causes the model to overly focus on making representations of the same event type more cohesive, neglecting the importance of separating different event types, thereby negatively impacting the final event detection accuracy.

\textbf{Inter-cluster Loss Weight ($\mu$).} 
We also choose 0.001, 0.005, 0.01, and 0.015 as candidate values for the inter-cluster loss weight. 
As shown in Figure~\ref{fig:Hyperparameter}(b), when $\mu$ is set too high (such as 0.015) or too low (such as 0.001), model performance decreases. 
Especially when $\mu$ is set too low, the model lacks sufficient constraint to push apart the centroids of different clusters, resulting in event clusters not being well separated. 
Conversely, when $\mu$ is set too high, the model places more emphasis on separating different event clusters, leading to a slight decrease in the cohesiveness of messages within clusters, thus mildly affecting the model's performance.

\revised{\textbf{Pairwise Cross-Entropy Loss Weight ($\kappa$).}
For pairwise cross-entropy loss weight, experiments are conducted on four values: 0.6, 0.8, 1.0, and 1.2.
Figure~\ref{fig:Hyperparameter}(c) shows a balanced allocation among the weights of pairwise cross-entropy loss, inter-cluster loss, and intra-cluster loss is crucial.
The model's overall performance decreases when the parameter $\kappa$ is set to 0.6 or 0.8. 
This is attributed to the insufficient constraint imposed by the pairwise cross-entropy loss and a too-small weight, which inversely amplifies the constraints of inter-cluster loss and intra-cluster loss.
This may cause similarity scores between messages obtained from the classifier to be inaccurate, thereby affecting the quality of the final message representation. 
Conversely, an excessively large $\kappa$ value weakens the constraints imposed by both inter-cluster and intra-cluster losses.
This, in turn, leads to message representations within the same event not being sufficiently tight and blurring the boundaries between different event clusters, thereby affecting the model's performance.
}

\textbf{Similarity Threshold ($\alpha$).} 
In studying the similarity threshold, we test four thresholds: 0.1, 0.3, 0.6, and 0.9. 
Figure~\ref{fig:Hyperparameter}(d) shows that model performance improves with increasing $\alpha$ values. 
This phenomenon can be attributed to a higher $\alpha$ value implying stricter requirements for message pair similarity. 
Consequently, the selected candidate representations are more likely to be extracted from message pairs within the same cluster. 
This leads to more stable and robust aggregated message representations. 
In addition, the choice of a reasonable similarity threshold is also crucial. 
Setting $\alpha$ too high may result in message representations that lack comprehensiveness.

\textbf{Number of Heads in Structured Attention Mechanism ($o$).}
The experiment examines settings of 1, 2, 3, and 4 attention heads.
As illustrated in Figure~\ref{fig:Hyperparameter}(e), when there is only one attention head, the model is limited in capturing key message features, resulting in limited message representations and consequently affecting model performance. 
Conversely, too many attention heads may introduce noise, leading to a decline in message representation quality and impacting model performance.

\textbf{Parameter Update Weight during Model Maintenance ($\zeta$).} We select 0.2, 0.4, 0.6, and 0.8 as candidate values for the parameter update weight. 
As shown in Figure~\ref{fig:Hyperparameter}(f), the model performs best when $\zeta$ is set to 0.4. When $\zeta$ is set to 0.2, there is a slight decrease in model performance, possibly due to the model retaining too much irrelevant historical information or information unrelated to current events. 
Conversely, when $\zeta$ values are too high (such as 0.6 and 0.8), the model retains too little historical knowledge, potentially making it difficult for the model to recognize and understand the connection between current and historical events, leading to a performance decline.

\revised{\textbf{Number of message sampled ($y$).}
We investigate the impact of varying $y$ on model performance, including four different values: 20, 40, 60, and 80.
As shown in Figure~\ref{fig:Hyperparameter}(g), both excessively large and small $y$ values decrease model performance.
For events with a large number of messages, sampling too few may result in insufficiently comprehensive learned message representations. 
Conversely, social media messages have the long-tail distribution characteristic.
When the sampling size is excessively large, the model may overfit the head events with a large number of messages, which neglects the representation learning of messages from long-tail events.
}

\revised{\textbf{Window Size ($w$).}
The window size determines the extent of message blocks included during continuous detection and the message block numbers used for model maintenance. 
We analyze the impact of four different window sizes: 1, 2, 3, and 4.
As observed in Figure~\ref{fig:Hyperparameter}(h), both excessively small (with w set to 1 or 2) and large window sizes (with w set to 4) result in varying degrees of decline in model performance.
Smaller windows are likely to contain more newly emerging or less frequently discussed events.
These events' relevant message counts are relatively less, making the contextual information insufficiently clear.
Consequently, during the model maintenance phase, the learned message representations of these events may not be comprehensive.
In subsequent window message blocks, when the message numbers describing these events increase, the model struggles to recognize them effectively due to insufficient learning of these events, thereby affecting overall performance.
Moreover, the smaller window requires more frequent model maintenance. 
This may accelerate the forgetting of historical knowledge and lead to the overfitting of current knowledge.
Conversely, when the window size is excessively large, some events may have already dissipated in subsequent message blocks.
Continuing to fit messages related to these events during model maintenance is unnecessary and may introduce noise, thereby compromising the model's performance.
}

\subsection{Case Study}
\label{sec: case study}
To investigate the impact of message structural information on GNN-based methods and proposed $\mathrm{RPLM}_{SED}$, we select $M_{19}$ from Events2012 for comparison and analysis with the optimal baseline method FinEvent.
The message graph of $M_{19}$ contains more noisy edges and isolated nodes, and the structural information is also relatively scarce.
This graph comprises 1,399 message nodes, including 92 isolated nodes, across 28 events, and possesses 21,374 edges based on entity relations (with 8,862 edges connecting messages from different events), 83 edges based on user relations (with 63 edges connecting messages from different events), and 623 edges based on hashtag relations (with 196 edges connecting messages from different events).

\begin{figure}[h]
    \centering
    \includegraphics[width=1\linewidth, trim={0.5cm 0 0.5cm 0}, clip]{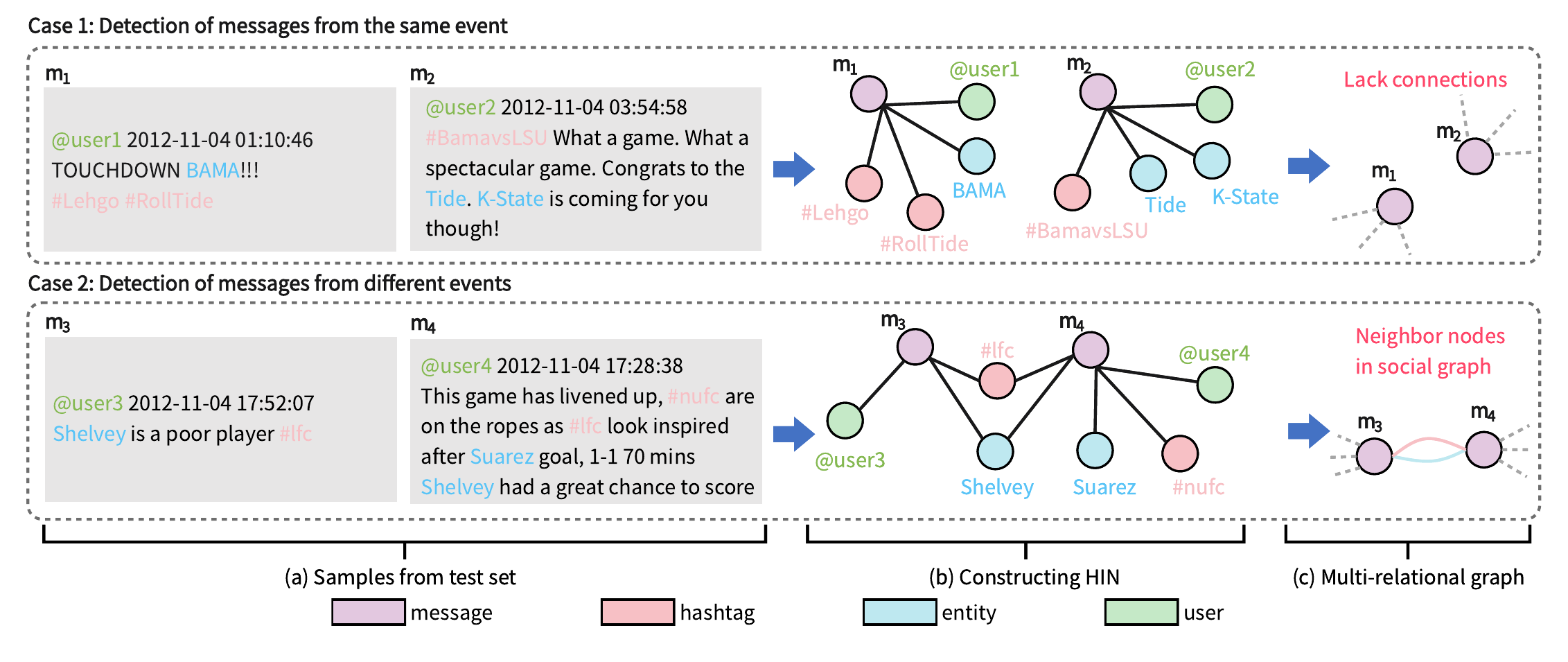}
    \caption{Raw Messages of $M_{19}$ within Events2012.}
    \label{fig:cases}
\end{figure}

Firstly, we extract two pairs of messages from $M_{19}$ where FinEvent predicts incorrectly while RPLM predicts accurately.
For the `case1' part in Figure~\ref{fig:cases}, $m_1$ and $m_2$, posted by different users, pertain to positive reactions towards the University of Alabama's performance in a specific game. 
\revised{The first message is focused on celebrating a particular scoring moment, while the second provides a broader perspective on the game and mentions a forthcoming challenge.}
Although $m_1$ and $m_2$ describe the same event, they share no common attributes other than belonging to the same time segment.
Unfortunately, FinEvent utilizes time by transforming it into temporal features combined with message content, resulting in no connections between these two messages in the constructed multi-relational message graph. 
This results in FinEvent's inability to aggregate and learn effective information about both from the explicit message graph directly, thereby failing to group them correctly.
Moreover, in the `Case 2' part from Figure~\ref{fig:cases}, $m_3$ articulates an opinion regarding a player, whereas $m_4$ delineates a football match.
These two messages from different events share the same hashtag `\#lfc' and entity `Shelvey', which, in turn, leads to their erroneous classification together by FinEvent.
In contrast,  $\mathrm{RPLM}_{SED}$ does not rely on the explicit structural relations between messages to learn representations.
It transforms the structural relations between messages into multi-relational prompt embeddings to complement message content embeddings, thereby enabling the joint utilization of message content and structural information.
During the learning process, by consistently encoding the structural relations and semantic content between messages, it can better learn and understand the complex structural relations between messages and the interaction between structural relations and semantic content. 
In the absence of common attributes or the presence of misleading information, $\mathrm{RPLM}_{SED}$ can still accurately group them by capturing the semantic and contextual relations between the messages.

\begin{figure}[h]
    \centering
    \includegraphics[width=1\linewidth]{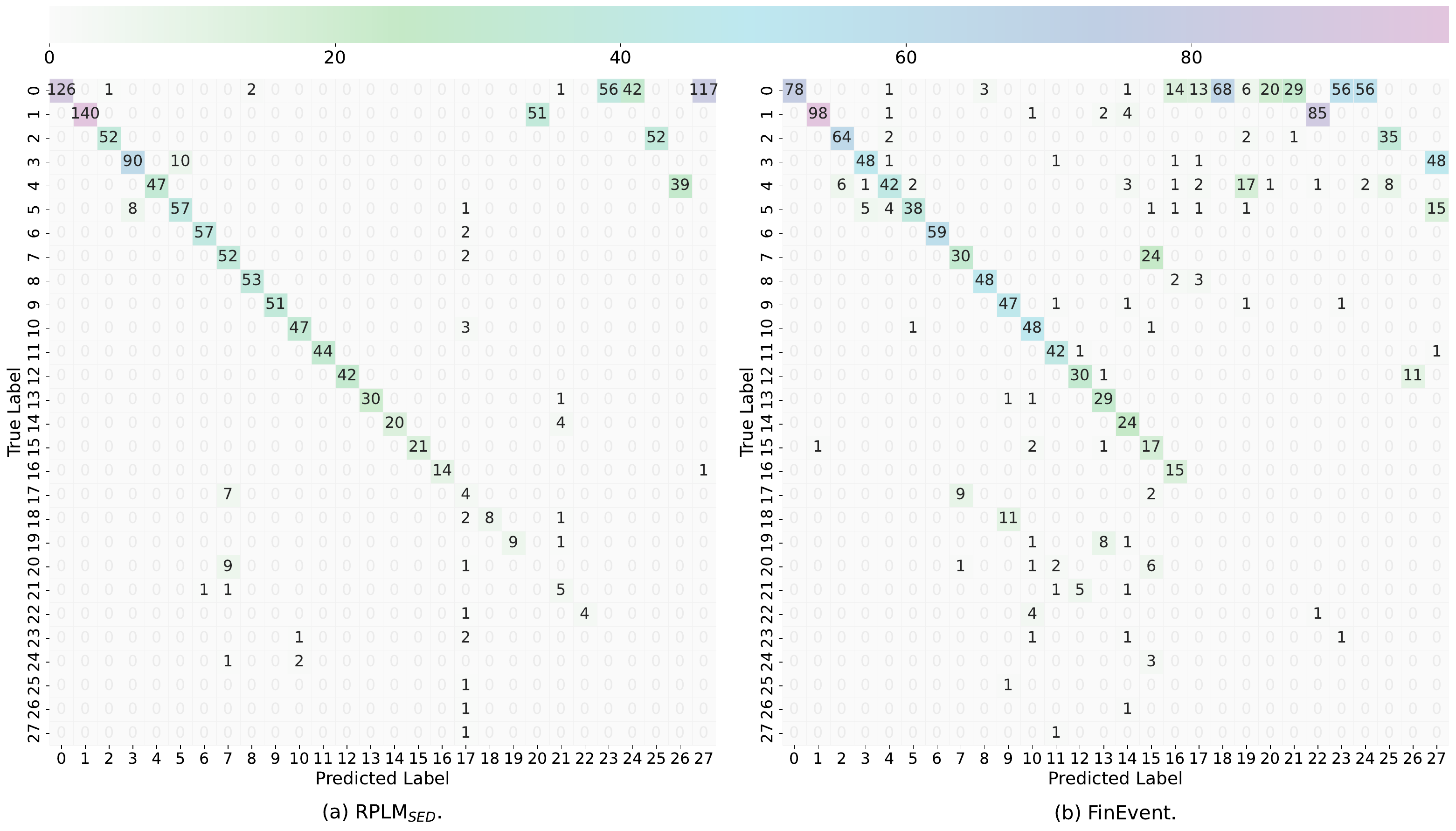}
    \caption{Confusion matrices of $M_{19}$ within Events2012.}
    \label{fig:Confusion matrices}
\end{figure}

Additionally, we construct confusion matrices for both RPLM and FinEvent, comparing their predicted labels with the true event labels, thereby observing the accuracy of event detection for each.
Given the discrepancy between the number of events obtained from density-based clustering methods and the actual number of events, we uniformly adopt the K-Means method to obtain predicted labels.
As shown in Figure~\ref{fig:Confusion matrices}, we observe that the clustering results of $\mathrm{RPLM}_{SED}$ (Figure~\ref{fig:Confusion matrices}(a)) generally have higher values on the diagonal compared to FinEvent (Figure~\ref{fig:Confusion matrices}(b)), and its prediction results are more concentrated. 
This indicates that the message representations obtained by $\mathrm{RPLM}_{SED}$ have higher accuracy, whereas FinEvent is more disturbed by noise during the feature aggregation process, leading to less clear feature boundaries. 
It is noteworthy that FinEvent is almost unable to identify events with very few messages (events numbered 17 to 27).

\revised{
\subsection{Clustering Algorithm Study}
\label{sec: Clustering Algorithm Study}
This section investigates the performance of three density-based clustering algorithms, \textbf{OPTICS}~\cite{ankerst1999optics}, \textbf{DBSCAN}~\cite{ester1996density}, and \textbf{HDBSCAN}~\cite{campello2013density}, on message representation obtained by $\mathrm{RPLM}_{SED}$. 
These algorithms are widely used in real-world applications due to their ability to cluster data without requiring a predefined number of event clusters.
Specifically, the core of OPTICS  is to generate an ordered list of points by analyzing the density and mutual relationships of data points, thus identifying clustering structures at various scales. 
OPTICS is particularly sensitive to the parameters of \textit{Minimum Points} and \textit{Cluster Reachability}.
The parameter of \textit{Minimum Points} determines the minimum number of points in a cluster. 
Setting this parameter too low may lead to the formation of numerous small clusters, whereas setting it too high may cause small clusters to be merged.
\textit{Cluster Reachability} evaluates the density changes at cluster boundaries to decide whether to split regions into separate clusters. 
Setting this parameter too low may blur cluster boundaries, causing regions with small density differences to be merged; setting it too high could result in over-segmentation of regions with minor density variations into multiple clusters.}

\revised{
DBSCAN forms clusters by defining core, border, and noise points.
It is highly sensitive to the parameters of \textit{Epsilon Neighbourhood} and \textit{Minimum Points}.
A point is considered a core point if the number of points in its neighborhood is greater than or equal to \textit{Minimum Points}; otherwise, it is classified as a border or noise point.
The algorithm starts from an unvisited core point and identifies all points within its neighborhood to form a new cluster. it recursively applies the same process to the core points within the neighborhood until the cluster can no longer be expanded.
When the \textit{Epsilon Neighbourhood} is set too small, a single cluster may be divided into multiple sub-clusters, instead, set too large, different clusters may be merged.
Setting the \textit{Minimum Points} too low may result in too many core points and ignoring noise points; 
setting it too high may result in only high-density regions being recognized as clusters, leading to more points being classified as noise.}

\revised{Compared to DBSCAN, HDBSCAN automatically adjusts the density threshold to recognize clustering structures at varying densities.
It constructs a density hierarchy tree to represent data structures at different densities and eventually prunes the tree to obtain the optimal clustering result.
The key parameters for HDBSCAN are the \textit{Minimum Cluster Size} and the \textit{Minimum Sample Number}. 
The \textit{Minimum Cluster Size} specifies the minimum number of points required for a cluster; 
setting it too low increases the number of noise points, while setting it too high may ignore small clusters. 
The minimum number of samples determines core points and influences cluster shapes. 
Typically, the \textit{Minimum Cluster Size} and the \textit{Minimum Sample Number} are set to be equal. }

\begin{figure}[h]
    \centering
    \subfigure{
        \includegraphics[width=0.7\linewidth, trim={0 2cm 0 2cm}, clip]{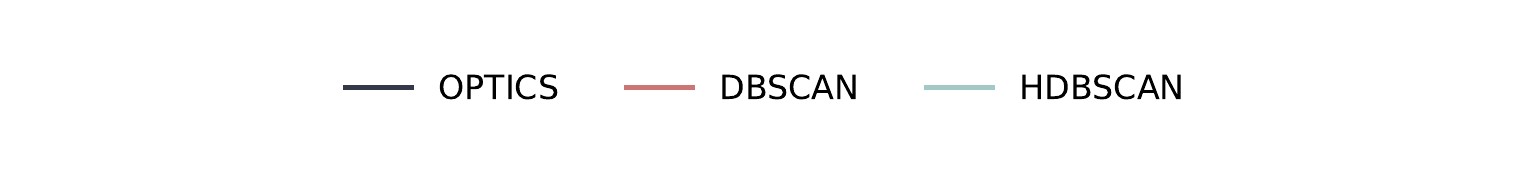}
    }
    \vspace{-10pt}
    \setcounter{subfigure}{0}
    \subfigure[NMI value in each message block of different algorithms.]{
        \includegraphics[height=5.6cm,trim={0.3cm 0.2cm 0 0}, clip]{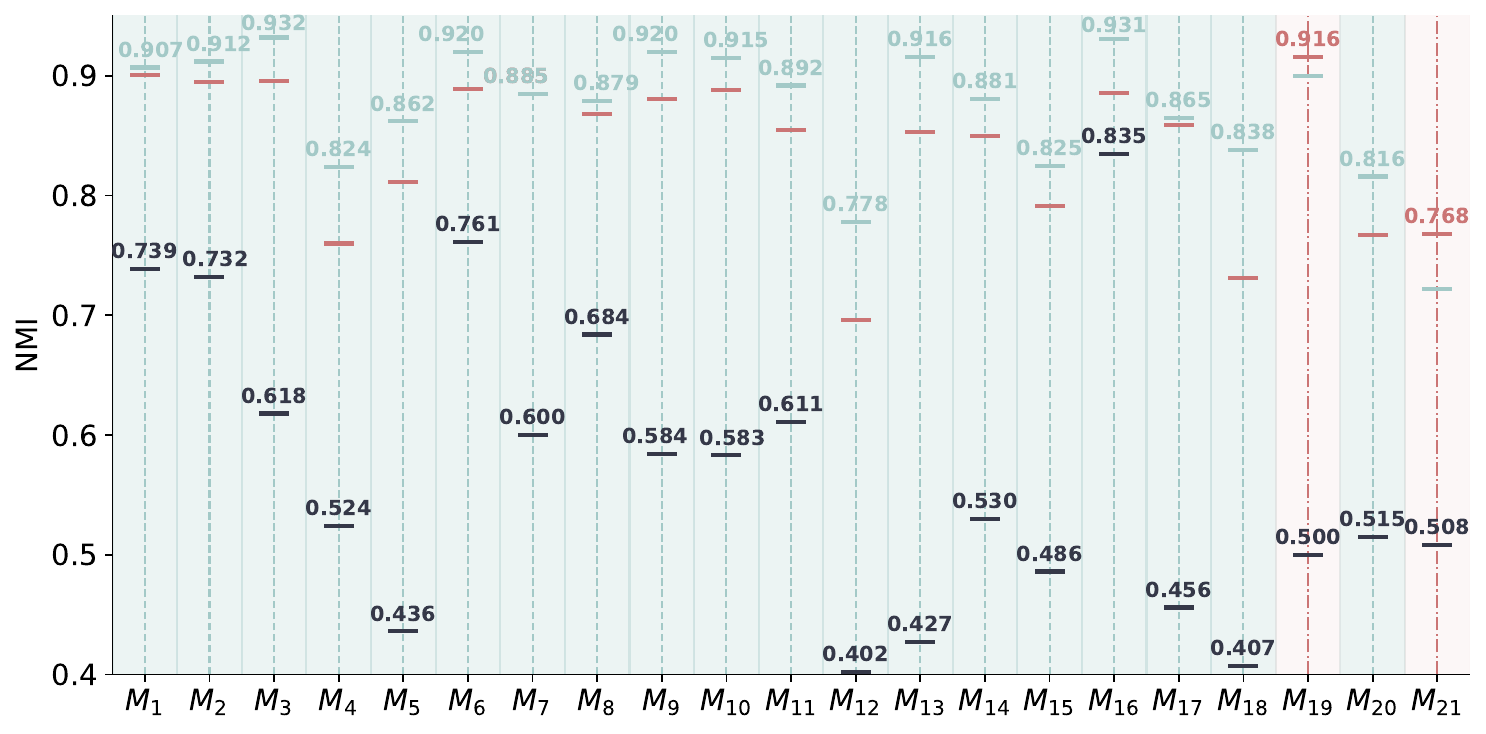}
    }
    \hspace{-10pt} 
    \subfigure[Average metrics]{
        \includegraphics[height=5.6cm,trim={0 0.2cm 0.3cm 0}, clip]{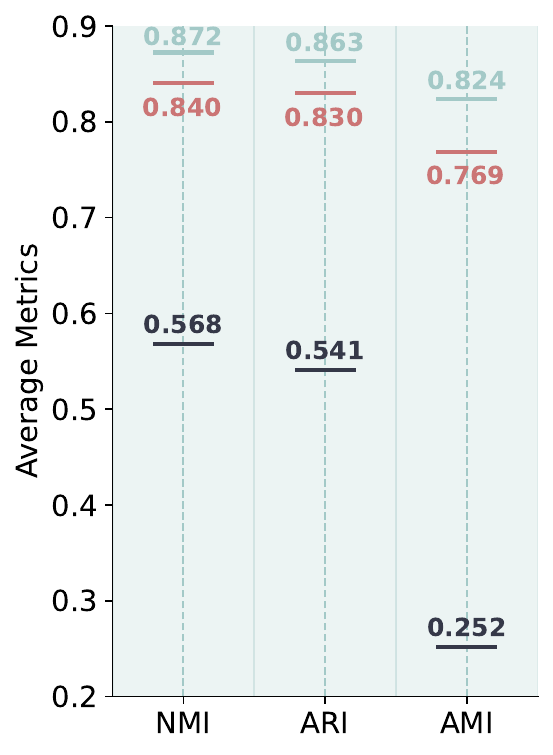}
    }
    \vspace{10pt}
    \caption{\revised{Performance of different clustering algorithms on Events2012.
    (a) The horizontal axis represents the message blocks, while the vertical axis denotes each message block’s corresponding NMI value.
    (b) The horizontal axis represents the metrics, and the vertical axis denotes each metric’s corresponding average value.
    Each message block (metric)’s background color corresponds to the algorithm’s color with the highest score with that block (metric).}}
    \label{fig:Performance of different clustering algorithms}
\end{figure}

\revised{
Experiments are conducted on the Events2012 dataset in the online detection scenario, in which metrics (NMI, AMI, ARI) for each message block under different algorithms are calculated.
The \textit{Minimum Points} and \textit{Cluster Reachability} parameters for OPTICS are set to 8 and 0.2, respectively. 
For DBSCAN, the \textit{Epsilon Neighbourhood} and \textit{Minimum Points} are set to 5 and 8, respectively. 
Additionally, the \textit{Minimum Cluster Size} for HDBSCAN is set to 8.
Figure~\ref{fig:Performance of different clustering algorithms} demonstrates that HDBSCAN is most suitable for the message representations learned by $\mathrm{RPLM}_{SED}$.
This is due to HDBSCAN requiring adjustment of only one parameter, the \textit{Minimum Cluster Size}.
In contrast, OPTICS and DBSCAN are more sensitive to multiple parameters. 
The distribution of data points across different message blocks may vary. 
Therefore, setting the same \textit{Cluster Reachability} (OPTICS) or \textit{Epsilon Neighborhood} (DBSCAN) parameters is not applicable to all message blocks.
}

\section{Related Work}
\label{sec: Related Work}

This section provides an overview of the existing literature and methodologies relevant to the proposed $\mathrm{RPLM}_{SED}$. 
The discussion is structured into three categories: Social Event Detection, Pre-trained Language Models, and Prompt Learning.
\subsection{Social Event Detection}

Social event detection represents a pivotal research domain, extensively categorized based on its objectives and application scenarios. 
In terms of goal-oriented classification, \revised{social event detection methods} are primarily divided into document-pivot (DP)~\cite{cao2021knowledge,peng2019fine, ren2022known,peng2022reinforced,liu2020story, aggarwal2012event, hu2017adaptive,zhou2014event, zhang2007new} and feature-pivot (FP)~\cite{fedoryszak2019real, 
fung2005parameter} approaches. 
DP methods focus on clustering social messages by analyzing their interrelations, emphasizing the grouping of relevant messages for a deeper understanding of the development and evolution of societal events.
These methods are particularly crucial in analyzing specific domain events such as disasters, stock trading, major sports events, and political activities, as they aid in organizing and comprehending the historical information of these events. 
The proposed $\mathrm{RPLM}_{SED}$ method in this paper is an exemplar of this category. 
In contrast, FP methods cluster messages based on the distribution characteristics of social message elements, such as vocabulary and named entities. 
They focus on the co-occurrence and distribution patterns of message elements to reveal inter-element correlations. 
Furthermore, \revised{social event detection methods} can be categorized based on their application scenarios into offline ~\cite{peng2019fine} and online methods~\cite{peng2021streaming,cao2021knowledge,ren2022known,peng2022reinforced, liu2020story,  hu2017adaptive,fedoryszak2019real,zhang2017triovecevent,goswami2016survey}. 
Offline methods are primarily used for analyzing historical events, whereas online methods continuously monitor social data streams, tracking the real-time development of events.
Online methods are crucial for rapid response and real-time event monitoring, especially in situations requiring swift action. 
Additionally, based on the employed techniques and mechanisms, these methods can be further subdivided into incremental clustering methods~\cite{aggarwal2012event, hu2017adaptive, zhang2007new,ozdikis2017incremental}, community detection methods~\cite{liu2020story,fedoryszak2019real, liu2018event, yu2017ring,liu2020deep,liu2020event,xing2022comprehensive}, and topic modeling methods~\cite{zhou2014event,becker2011beyond,cheng2014btm,cordeiro2012twitter,peng2018large,blei2003latent}

Recently, methods based on Graph Neural Networks (GNNs)~\cite{peng2021streaming,cao2021knowledge,cui2021mvgan,peng2019fine,peng2022reinforced,ren2022known,ren2021transferring,ren2022evidential,ren2021transferring,sun2023social} have demonstrated potential in the domain of social event detection.
By modeling social messages as graphs, these approaches effectively integrate messages' semantic and structural information, overcoming the limitations of insufficient information utilization within traditional methods.
For instance, KPGNN~\cite{cao2021knowledge} converts messages with common attributes into homogeneous message graphs, which achieves incremental social event detection by leveraging the inductive learning capabilities of GNNs.
FinEvent~\cite{peng2022reinforced} reduces the impact of noisy edges between message nodes by constructing weighted multi-relational graphs and utilizes reinforcement learning to determine optimal relation weights.
Additionally, QSGNN~\cite{ren2022known} employs utilizing the finest known samples and dependable knowledge transfer to extend knowledge from the known to the unknown, markedly diminishing the requirement for manual annotation in model training.
However, GNN-based methods learn representations from explicit message graphs.
These graphs may contain noisy edges, and messages from the same event without common attributes lack connections. 
This leads to GNNs being impacted when learning message representations, an effect that is exacerbated in the presence of sparse structural information.
Moreover, these methods initialize message embeddings statically before training, preventing the model from dynamically obtaining better message embeddings during the training phases.
In recent research, \citet{cao2024Hierarchical} propose HISEvent, which explores the neighborhood of a message graph by minimizing 1-dimensional Structural Entropy (1D SE) and supplementing edges in the existing message graph between semantically close messages. 
Subsequently, by minimizing 2-dimensional structural entropy (2D SE) for partitioning the message graph, the approach eliminates the cost of manual annotation, achieving unsupervised social event detection.
It is imperative to acknowledge that although HISEvent employs a hierarchical approach to mitigate the computational complexity of conventional vanilla greedy 2D SE minimization algorithm~\cite{li2016Three-Dimensional} from $O(N^3)$ to $O(n^3)$ (where $N$ and $n$ respectively denote the number of nodes within a message graph and that within a manually selected sub-graph), methods based on GNNs or PLMs exhibit a computational complexity of merely $O(N)$. 
Given that HISEvent is an unsupervised approach, it has not been compared in this paper.
In addition, Our proposed `pairwise' approach significantly differs from that introduced in PPGCN~\cite{peng2019fine}.
In PPGCN, after obtaining representations, the pairwise sampling method selects a positive and a negative sample for each message, addressing the challenge of numerous event types with few messages per event.
Unlike PCGNN, our pairwise message modeling strategy is implemented during the data preprocessing stage to preserve structural relations rather than only focusing on inter or intra-event relations.

\subsection{Pre-trained Language Models}
In recent years, Pre-trained Language Models (PLMs)~\cite{liu2019roberta,zhang2022twhin,loureiro2023tweet,devlin-etal-2019-bert,radford2018improving,lewis2020bart} have made significant strides in the field of Natural Language Processing (NLP). 
These models have learned rich language representations by undergoing pre-training on large-scale text corpora. 
When applied to downstream tasks, fine-tuned PLMs can adapt to a variety of application scenarios, including, but not limited to, text classification~\cite{devlin-etal-2019-bert,sun2019fine}, question answering~\cite{shu2022tiara,wei2023menatqa}, language generation~\cite{radford2018improving}, and text summarization~\cite{lewis2020bart}. 
In processing social data, PLMs have demonstrated high adaptability to informal texts and multiple languages~\cite{zhang2022twhin,loureiro2023tweet}, effectively handling texts containing internet-specific abbreviations, emojis, or slang. 
This has extended the application of PLMs to social event detection tasks.
However, previous social event detection approaches often treated PLMs merely as a tool for word embedding, extracting message embeddings through PLMs and then applying clustering algorithms for event detection. 
This overlooks the capabilities of PLMs to learn, retain, and expand knowledge, leading to suboptimal performance in the complex environment of social media data streams.
In contrast to prior studies, we leverage PLM as the backbone architecture and fine-tune it during the initial training and maintenance phases, thereby achieving high-quality detection of social events.

\subsection{Prompt Learning}
Prompt learning~\cite{lester2021power,liu2023pre}, as an emerging technology in NLP, has garnered widespread attention and has been applied to a variety of tasks, including recommendation systems~\cite{ZhangW23}, aspect sentiment triplet extraction~\cite{Kun2023Promp}, text generation~\cite{LiTNWZ22}, question
answering~\cite{PetroniRRLBWM19}, etc.
The core of prompt learning lies in the design of prompt templates, which are typically categorized into discrete templates, continuous templates, and hybrid templates. 
Discrete templates employ explicit natural language terms, making them direct, easy to understand, and implement, but they also rely on human experience.
For instance, \citet{PetroniRRLBWM19} design specific sentence structures as prompts to query the PLMs about their knowledge of certain facts.
Contrary to discrete templates, continuous templates do not directly carry semantic information but consist of a set of learnable embedding vectors~\cite{GuHLH22, DingHZCLZS22}. 
By continuously optimizing these prompts throughout the training process, the model is better adapted to the task.
In our proposed $\mathrm{RPLM}_{SED}$, the multi-relational prompt embeddings are a type of continuous prompt template.
Hybrid templates combine discrete and continuous templates, aiming to leverage human expertise and model auto-optimization to achieve superior performance and broader adaptability~\cite{ZhangW23}.
Unlike traditional prompt learning, to ensure that $\mathrm{RPLM}_{SED}$ can adapt to dynamic social streams, we utilize multi-relation prompts and fine-tune the model's parameters during the training or maintenance phases.

\section{Conclusion}
\label{sec: conclusion} 

This paper studies \textit{$\mathrm{RPLM}_{SED}$} relational prompt-based pre-trained language models for Social Event Detection.
A pairwise message modeling strategy is proposed to address the issues of missing and noisy edges in social message graphs. 
A multi-relational prompt-based pairwise message learning mechanism is presented to simultaneously leverage the content and structural information of messages, learning more robust and stable message representations.
A clustering constraint is designed to optimize the training process, enhancing message representations' distinguishability. 
The conducted experiments on three real-world social media datasets demonstrate that $\mathrm{RPLM}_{SED}$ achieves SOTA performance in offline, online, low-resource, and long-tail distribution social event detection tasks.
In the future, we aim to explore how supervised $\mathrm{RPLM}_{SED}$ can be extended to semi-supervised or unsupervised social event detection tasks while also investigating ways to optimize the model architecture further to achieve more efficient detection. 

\section*{Acknowledgments}
This work is supported by the NSFC through grants 62266028, 62322202, U21B2027, and 62432006, Yunnan Provincial Major Science and Technology Special Plan Projects through grants 202402AD080002, 202302AD080003 and 202202AD080003, Key Projects of Basic Research in Yunnan Province through grants 202301AS070047, 202301AT070471.

\bibliography{reference}

\end{document}